\documentclass[a4paper,fleqn]{cas-sc}

\usepackage[authoryear]{natbib}
\def\tsc#1{\csdef{#1}{\textsc{\lowercase{#1}}\xspace}}
\tsc{WGM}
\tsc{QE}
\usepackage{graphicx}
\usepackage{caption}
\usepackage{subcaption}
\usepackage{booktabs}
\usepackage{multirow}
\usepackage{placeins}
\usepackage{lineno}
\usepackage{float}
\floatstyle{plaintop}
\restylefloat{table}
\makeatletter
\def\ps@appfirstpage{
  \let\@oddhead\@empty    
  \let\@evenhead\@empty      
}

\begin{document}
\let\WriteBookmarks\relax
\def\floatpagepagefraction{1}
\def\textpagefraction{.001}

\shorttitle{LC-SLab}

\shortauthors{Leonhardt et al.}

\title [mode = title]{LC-SLab -- An Object-based Deep Learning Framework for Large-scale Land Cover Classification from Satellite Imagery and Sparse In-situ Labels} 


\author[1]{Johannes Leonhardt}[orcid=0000-0002-4505-5086]
\cormark[1] % Corresponding author indication
\ead{jleonhardt@uni-bonn.de}
\credit{Conceptualization, Methodology, Software, Validation, Formal Analysis, Writing - Original Draft, Visualization}

\author[1,2]{Juergen Gall}
\credit{Conceptualization, Writing - Review \& Editing, Supervision, Funding acquisition}

\author[1,2]{Ribana Roscher}
\credit{Conceptualization, Writing - Review \& Editing, Supervision, Funding acquisition}

\affiliation[1]{organization={University of Bonn},
    city={Bonn},
%          citysep={}, % Uncomment if no comma needed between city and postcode
    country={Germany}
}

\affiliation[2]{organization={Lamarr Institute for Machine Learning and Artificial Intelligence},
    city={St. Augustin},
%          citysep={}, % Uncomment if no comma needed between city and postcode
    country={Germany}
}

\begin{abstract}
Large-scale land cover maps generated using deep learning play a critical role in data-driven analysis and decision-making across a wide range of Earth science applications. Open in-situ datasets from principled land cover surveys offer a scalable alternative to manual annotation for training such models. However, their sparse spatial coverage leads to fragmented and noisy predictions when used with existing deep learning-based land cover mapping approaches. A promising direction to address this issue is object-based classification, which assigns labels to semantically coherent image regions rather than individual pixels, thereby imposing a minimum mapping unit that controls spatial fragmentation. Despite this potential, object-based methods remain underexplored in deep learning-based land cover mapping pipelines, especially in the context of medium-resolution imagery and sparse supervision.

To address this gap, we propose LC-SLab, the first deep learning framework for systematically exploring object-based deep learning methods for large-scale land cover classification under sparse supervision. LC-SLab supports both input-level aggregation via graph neural networks, and output-level aggregation by postprocessing results from established semantic segmentation models. Additionally, we incorporate features from a large pre-trained network to improve performance on small datasets.

We evaluate the framework on annual Sentinel-2 composites with sparse LUCAS labels, focusing on the tradeoff between accuracy and fragmentation, as well as sensitivity to dataset size. Our results show that object-based methods can match or exceed the accuracy of common pixel-wise models while producing substantially more coherent maps. Input-level aggregation proves more robust on smaller datasets, whereas output-level aggregation performs best with more data. Several configurations of LC-SLab also outperform existing land cover products, highlighting the framework's practical utility.
\end{abstract}


% % Research highlights
% \begin{highlights}
% \item LC-SLab framework for land cover classification with sparse labels is introduced
% \item Different object-based deep learning methods are compared
% \item Methods provide more coherent maps while matching the accuracy of pixel-wise methods
% \item Ideal method depends on dataset size and prescribed minumum mapping unit
% \item Produced maps are more accurate than existing large-scale products
% \end{highlights}

% Keywords

\maketitle

% \linenumbers

% Main text
\section{Introduction}
\label{sec:introduction}

Land cover data, i.e., spatial information on the Earth surface's biophysical properties, is essential for many applications, such as climate science, biodiversity conservation, regional planning, and disaster management. To obtain continental or global scale land cover maps, researchers and practitioners have for a long time relied on automatic interpretation of satellite images, relating the spectral information captured by the sensor to specific land cover classes \citep{townshend1992land,wulder2018land,radeloff2024need}. As a result, the recent advances in satellite technology on the one hand and machine learning methodology on the other, have massively improved the level of detail of existing land cover products \citep{zhu2017deep,ma2019deep,kotaridis2021remote,zang2021land}.

To train classification models and validate the results, researchers typically rely on training datasets comprising pairs of satellite images and corresponding land cover labels. The most common way of obtaining such datasets is annotation by photointerpretation. While the datasets generated in this manner have been instrumental for both methodological development and practical land cover mapping, they have two significant problems which inhibit their potential for large-scale application: First, the acquisition is time- and cost-intensive, as it requires paid annotators to manually label large volumes of imagery. This poses a significant bottleneck especially in the context of land cover mapping on broad geographic scales, where representation of geographic diversity in the training dataset is important. Second, the resulting labels are susceptible to human error and interpretation. This results in label uncertainty which directly limits the accuracy of the resulting land cover maps \citep{elmes2020accounting,koller2022uncertainty}.

Because of this, using labels derived from open in-situ datasets collected in surveys such as Eurostat's Land Use/Cover Area frame Survey (LUCAS) \citep{d2020harmonised} or in the context of citizen science initiatives like Geo-Wiki \citep{fritz2012geo,fritz2017global} is an appealing idea for land cover mapping on large scales. In previous work, LUCAS data and Landsat-7 imagery have been used for crop type mapping across Germany \citep{conrad2010mapping,mack2017semi}. Additionally, LUCAS was harmonized with Sentinel-2 to automatize the generation of general land use and land cover maps \citep{weigand2020spatial}. On pan-European scales, LUCAS-based labels have also been used to generate both general land cover and land use maps \citep{pflugmacher2019mapping,mirmazloumi2022elulc}, as well as crop type maps \citep{d2021parcel,ghassemi2024european}. However, the above described works mostly use comparably simple workflows based on pixel or object-wise hand-crafted features in combination with rule-based classification methods like random forest or simple multi-layer perceptrons (MLP) for classification, limiting their potential in terms of accuracy.

Notably, while the methodological developments in the field of deep learning-based semantic segmentation with convolutional nerual networks (CNNs) \citep{lecun1989handwritten} and vision transformers (ViTs) \citep{dosovitskiy2020image} have already been applied successfully for large-scale land cover classification \citep{kussul2017deep,karra2021global,brown2022dynamic}, they have only sporadically been applied in the context of sparse labels: As such, \citet{wang2020weakly} showed that even in a sparsely labeled regime, deep learning approaches outperform standard pixel-wise classifiers. \citet{galatola2023land} trained different CNN- and ViT-based semantic segmentation models on a dataset consisting of monotemporal Sentinel-2 images, sparse LUCAS labels, and scribble annotations spanning southern Europe. A multi-task learning setup for crop cover mapping from Chinese survey data was used by \citet{cai2024costeffective}, while \citet{leonhardt2024sparsely} developed an architecture incorporating oversegmentations to improve CNN-classification accuracy based on only Sentinel-2 composites and LUCAS labels. In concurrent work, \citet{sharma2024sen4map} presented an open source benchmark dataset consisting of Sentinel-2 time series and LUCAS labels, specifically highlighting the potential of deep learning also in the context of sparse in-situ data.

One of the key challenges holding back the application of deep learning-based semantic segmentation approaches is the spatially sparse distribution of the in-situ labels. This is because most models for semantic segmentation are trained on datasets with dense labels, i.e., every pixel is assigned a label, limiting their transferability to the sparsely labeled setting. In this context, several works have described that models trained directly on such sparse annotations tend to produce under‑constrained, spatially inconsistent segmentations, with missing or poorly localized object regions and degraded boundary quality \citep{bearman2016whats,alonso2019coralseg,hua2021semantic,chan2025sparsea}. However, this effect is typically characterized only qualitatively through visual comparisons or descriptive discussion and a systematic analysis between spatial fragmentation and classification accuracy is missing. 

In the context of land cover classification, the presence of spatially inconsistent land cover units severely limits the resulting maps' usefulness for downstream applications like field delineation or building detection where consistent object-level classification is essential \citep{saura2002effects,garcia2019sensitivity}. To tackle this problem, a common approach in land cover mapping is to enforce spatial smoothness by defining a minimum mapping unit (MMU) which is the area of the smallest connected unit that is assigned a distinct land cover label. This practice reduces noise and improves the semantic consistency of the map by suppressing spurious small patches that may arise from classification errors or local variation in input features.

Object-based classification methods \citep{blaschke2010object,blaschke2014geographic}, in particular when combined with deep learning \citep{ma2024deep}, offer a powerful, yet scalable way to enforce the MMU by assigning labels to semantically coherent image regions, i.e., objects, rather than individual pixels. In this context, the object information can be incorporated at the input level by segmenting the image prior to classification, represent the data as a set of objects, and using these as units of analysis \citep{mirmazloumi2022elulc}. Recent works have identified the use of graph neural networks (GNNs) as a particularly promising methodological direction in this context \citep{diao2022superpixel,zhao2022contextual,kavran2023graph}. Alternatively, the object definitions can be used at the output level, where pixel-level predictions are aggregated into region-wise decisions via postprocessing \citep{liu2019integration,mi2020superpixel,timilsina2020mapping}. As a middle course between these two paradigms, recent studies have begun to explore intermediate feature-level aggregation, where pixel-level features are pooled within object boundaries to create region descriptors used for classification. While some recent works have considered object-based land cover classification from medium-resolution imagery \citep{ienco2019combining}, most existing work has focused on very high resolution imagery which is not suitable for large-scale land cover mapping due to computation and data availability constraints. On the other hand, no work so far has systematically explored different object-based deep learning strategies in the context of large-scale land cover mapping.

To fill this research gap, we present the first framework for object-based deep learning in the context of large-scale land cover classification from sparse labels (LC-SLab). LC-SLab comprises both input-level and output-level object aggregation methods, as well as the capability to incorporate features from a pre-trained model which has also not been previously considered in this particular setting. For the first time, LC-SLab allows for the systematic analysis of the different approaches' capabilities and synergies. Our evaluation specifically targets the analysis of the tradeoff between classification accuracy and MMU, as well as the methods' robustness with respect to dataset size, which is an important consideration in the context of in-situ data. 

Our most important findings are summarized as follows:
\begin{itemize}
    \item We demonstrate that training deep learning models on sparse in-situ labels without additional regularization leads to overly fragmented land cover maps. Object-based deep learning methods can match or even surpass the accuracy of semantic segmentation approaches while at the same time enforcing a MMU, leading to significantly less fragmented results.
    \item Output-level object aggregation approaches perform best when many sparse labels are available, while input-level aggregation is preferable for small training datasets.
    \item The incorporation of features from a pre-trained model consistently leads to improvements in terms of both accuracy and label efficiency and enables training with smaller models.
    \item Several configurations of LC-SLab outperform established third-party products.
\end{itemize}
Overall, these findings position LC-SLab as a useful foundation for the production of interpretable and actionable large-scale land cover maps under real-world data constraints.

\section{Methods and data}
\label{sec:methods_and_data}

% Topical subheadings are allowed. Authors must ensure that their Methods section includes adequate experimental and characterization data necessary for others in the field to reproduce their work.

In this section, we describe the methodology and materials used in this study: First, we present the components of the proposed LC-SLab framework in Section \ref{sec:the_lc-slab_framework}. Subsequently, we will provide details on the dataset used in this study, including the underlying data sources and the compilation in Section \ref{sec:dataset}. Finally, we outline our evaluation strategy in Section \ref{sec:evaluation}, as well as the experimental setup in Section \ref{sec:implementation_details}.

\subsection{The LC-SLab framework}
\label{sec:the_lc-slab_framework}

Formally, land cover classification is a standard semantic segmentation task: The input is a satellite image $\mathbf{X} \in \mathbb{R}^{(H \times W \times C)}$, with $H$, $W$, and $C$ being the number of rows, columns, and channels, respectively. The output is a predicted map of pixel-wise land cover cover class probabilities $\widehat{\mathbf{Y}} \in \mathbb{R}^{(H \times W \times K)}$, where $K$ is the number of classes.

\begin{figure}[]
    \centering
    \includegraphics[scale=1.0]{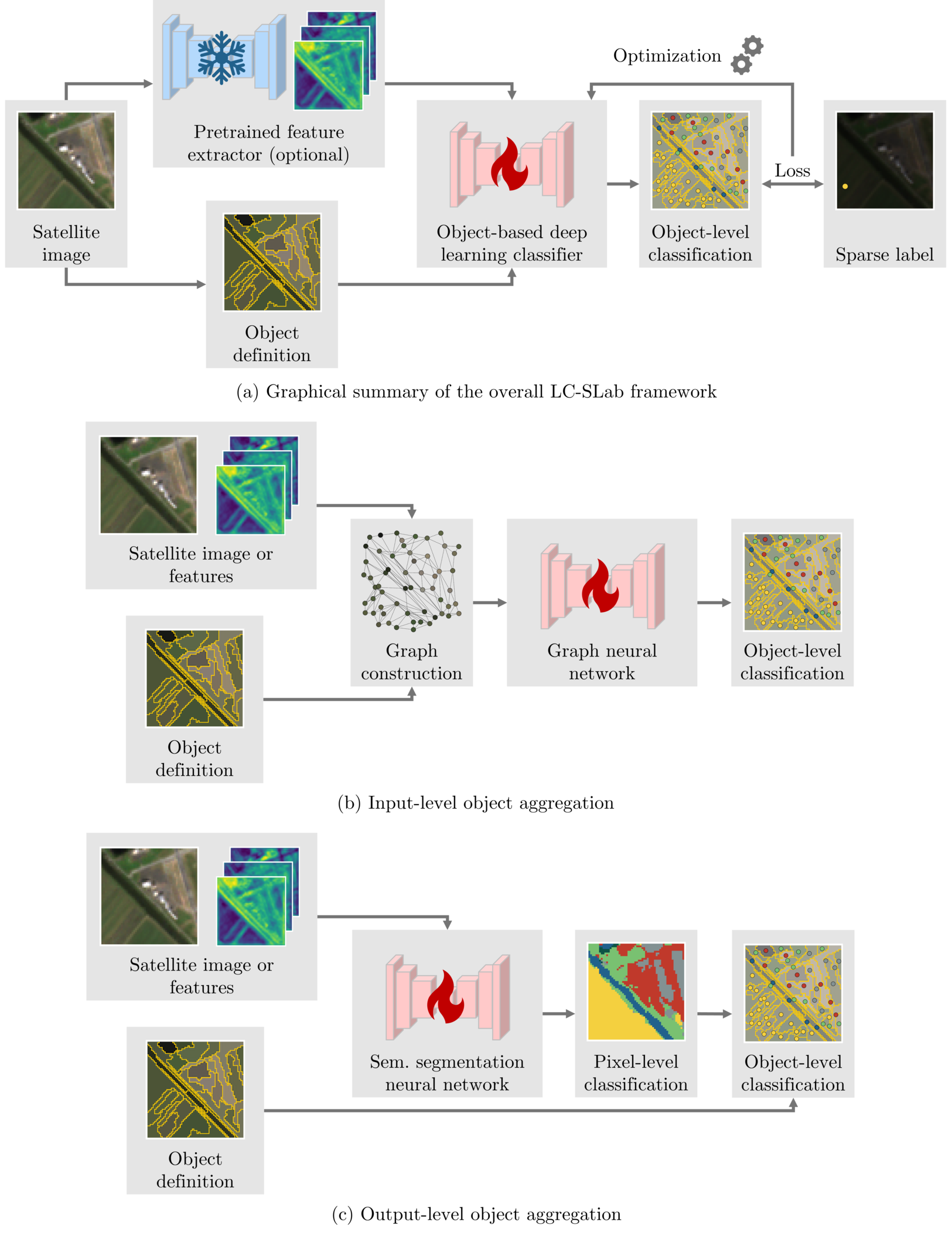}
     \caption{Outline of the LC-SLab framework. The object-based deep learning classifier from (a) can use either input-level object aggregation which is outlined in (b) or output-level object aggregation which is outlined in (c). Best viewed zoomed in.}
    \label{fig:methods}
\end{figure}

In the context of the proposed LC-SLab framework, the prediction is based on a series of different operations, which are described in detail below. In particular, we make use of object-based deep learning methods, which conduct the classification on the basis of groups of pixels called objects, rather than individual pixels. As outlined in Figure \ref{fig:methods}, the framework comprises an object definition module, an optional feature extraction module, as well as the object-based deep learning classifier. For the latter, we consider methods which aggregate the objects on an input level, and ones which aggregate object at the output level. The classifiers are optimized with respect to the sparse labels.

Below, we describe these components in more detail. To this end, we focus on conveying the underlying ideas of each technique and their role in the LC-SLab framework, rather than presenting detailed technical descriptions. For formal definitions and implementation details, we refer the reader to the original publications cited throughout the section.

\subsubsection{Object definition}
\label{sec:object_definition}

First, the objects in a satellite image are defined using an unsupervised oversegmentation algorithm $\mathbf{S} = \mathcal{S}(\mathbf{X}) \in \mathbb{N}^{(H \times W)}$. Such algorithms partition an image into segments by mapping each pixel to an object index $1 ... S$. For the subsequent object-based classification, it is assumed that all pixels assigned to the same object belong to the same land cover class. This assumption is more likely to be violated in case of larger objects which is why the granularity of the object definition generally enables a tradeoff between accuracy and complexity.

In our implementation, we mainly use the established Felzenszwalb-Huttenlocher (FH) algorithm for image segmentation \citep{felzenszwalb2004efficient} which constructs a graph over the image where each pixel is a node connected to its neighbors by weighted edges based on color similarity. It then iteratively merges regions in a way that favors consistency within a segment and prominent boundaries. One of the FH algorithm's key parameters is the minimum segment size $A_{min}$, which prevents small, spurious object definitions by enforcing a lower bound on the number of pixels per segment. The MMU of the resulting land cover map is thus defined a-priori as the product of $A_{min}$ and the spatial resolution.

The results when using FH are compared to that of Simple Linear Iterative Clustering (SLIC) which works by performing a localized adaptation of the $k$-means algorithm to cluster pixels in a combined space defined by spectral values and image coordinates. To enforce a MMU in SLIC, the number of clusters $k$ is set to $(64^2 / A_{min})$, rounded to the nearest lower integer, and the minimum size of any segment is specified as $A_{min}$, effectively suppressing small clusters which would lead to a violation of the prescribed MMU. For the hyperparameter which controls the weighting between spatial and spectral distance during the clustering, we choose a value of $0.1$ which we found to provide meaningful segments for all specified MMUs.

Throughout our experiment, we use five different object definitions: First, we use the FH and SLIC algorithm with the MMU set to $5$, $10$, $20$, and $40$ pixels, which in the case of Sentinel-2, as used in this study, corresponds to the MMUs $500m^2$, $1000m^2$, $2000m^2$, and $4000m^2$, respectively. For simplicity, we will use pixels ($px$) as the unit for the MMU below. 

As an additional baseline, we also consider a object definition, where each pixel is assigned an individual object, resulting in a MMU of $1px$. Using such a naive object definition, the object-wise classifiers, as investigated in this study, can be generalized to the pixel-wise classification case usually employed for semantic segmentation and compared to established models also in this setting.

\subsubsection{Feature extraction with a pre-trained model}
\label{sec:feature_extraction_with_a_pre-trained_model}

In parallel to the object definitions, a pixel-wise feature map $\mathbf{F} \in \mathbb{R}^{(H \times W \times F)}$, where $F$ is the number of extracted features, may be extracted from the input image using a pre-trained deep learning model $\mathcal{F}$: $\mathbf{F} = \mathcal{F}(\mathbf{X})$. These extracted features are subsequently used as the input to the object-based classifier. If a pre-trained model is not specified, the original image is used, i.e., $\mathbf{F} = \mathbf{X}$ and $F=C$.

The pre-training approach bridges the gap between methodological developments achieved on large benchmark dataset and applied land cover mapping, as models pre-trained on large datasets or in a self-supervised manner can be used in the sparsely labeled setting considered in this work. Generally, transfer learning based on pre-trained models has been an established practice in deep learning \citep{zhuang2020comprehensive}. To our knowledge, however, this is the first study to systematically assess such an approach in the setting of large-scale land cover classification with sparse labels. In this context, the use of pre-trained models is an especially promising direction, as it offers a scalable solution to the label sparsity and availability problems.

For our experiments, we use a UPerNet \citep{xiao2018unified} with a ResNet-152 backbone pre-trained on dense pseudo-labels from an existing land cover product as the feature extractor. The features are taken from the last layer before the final prediction. Even though the land cover product used for pre-training may not be sufficiently accurate, the backbone will have already been exposed to the task of land cover classification and thus have learned meaningful features in the context of this task. Additional details on the pre-training procedure are provided in Section \ref{sec:implementation_details}. 

\subsubsection{Input-level object aggregation}
\label{sec:input-level_object_aggregation}

For the object-based deep learning classifier, we differ between input-level object aggregarion approaches and output-level object aggregation approaches. In the case of input-level object aggregation, as depicted in Figure \ref{fig:methods}(b), the image or the derived feature map $\mathbf{F}$ is represented as an object graph $\mathcal{G}_{\mathbf{F}} = (\mathbf{V}, \mathbf{E})$. Each node $\mathbf{v}_s \in \mathbf{V}$ corresponds to a segment index $s$, and the edges $[s, s'] \in \mathbf{E}$ represent the neighborhood between segments. This graph is subsequently fed into a graph neural network which predicts a land cover class for each node. Since every node represents a segment, the node-wise result can easily be remapped to the original image resolution based on the object definition  $\mathbf{S}$.

\paragraph{Graph construction}

The node features $\mathbf{v}_s$ are designed to capture the corresponding segment's spectral and geometric properties. They are computed from the set of pixel-wise features assigned to the object with index $s$.

First, the mean intensity features $\mu_s \in \mathbb{R}^F$ describe the channel-wise mean intensity. To characterize spectral variability, we additionally compute segment-level variability features, including the channel-wise minimum $\min_s \in \mathbb{R}^F$ and maximum $\max_s \in \mathbb{R}^F$, as well as the standard deviation, $\mathbf{\sigma}_s \in \mathbb{R}^F$. 

Beyond spectral properties, we also include geometric features to describe the shape and spatial distribution of the segments, as is commonly done in object-based image analysis \citep{blaschke2010object,blaschke2014geographic,ma2017evaluation,cosma2023geometric}. We incorporate both size and shape-based descriptors, which are invariant to image orientation, an important property in remote sensing applications. The segment size is given by its pixel count, normalized by the number of pixels in the image $n_s \in \mathbb{R}$. Shape is captured by the mean radial distance $\delta_s \in \mathbb{R}$ and the mean radial dispersion $\varphi_s \in \mathbb{R}$, which are the mean and the standard deviation of the distance of a segment's pixels from its centroid, respectively. 

Note that the features describing spectral variability ($\min_s$, $\max_s$, $\sigma_s$) as well as the geometric features ($n_s$, $\delta_s$, and $\varphi_s$), are only meaningful if a MMU is specified. We also do not compute these features if learned feature maps are used for the prediction in place of the original image, as it produces significant computational overhead and because variability around a pixel can already be sufficiently captured by a pre-trained feature extractor.

Edges in the graph are defined using a region adjacency graph (RAG), where two nodes are connected if their corresponding segments share a boundary in the image. For implementation and notation purposes, we also include self-loops for all nodes.

\paragraph{Graph neural networks}

The constructed graph is passed into a GNN, which conducts a node-wise classification. In our experiments, we consider different kinds of GNNs, both in terms of architecture and the graph convolutional operator: While the architecture defines the way a GNN's different layers are arranged, the graph convolutional operator defines the way information is propagated throughout the graph within a single graph convolution layer.

As GNN architectures, we first use a simple GNN (BaseGNN). In BaseGNN, the inputs are first mapped to a $256$-dimensional feature vector using a single linear layer. This is followed by three graph convolution layers which maintain the feature dimensionality of $256$. All layers are complemented by batch normalization and ReLU activations. In the final layer, the learned features are mapped to the unnormalized pixel-wise outputs, i.e., the logits, using an unactivated linear layer.

Additionally, we consider a variant of Graph UNet (GUNet) \citep{gao2019graph,gao2022graph} which mimics the CNN-based UNet by introducing graph pooling layers and skip connections. In particular, our implementation first maps the input to a $64$-dimensional feature map which is subsequently passed through an encoder consisting of three blocks of double graph convolutions and graph pooling layers which double the number of feature maps. The structure is mirrored in the decoder where the blocks consist of graph upsampling and double convolution layers. As in BaseGNN, all convolution layers are accompanied by batch normalization and ReLU activations.

In the original implementation of GUNet, attention-based node selection is used for pooling which leads to isolated nodes in deeper layers for the rather sparsely connected RAGs. Therefore, we instead use pooling based on the greedy Graclus graph clustering algorithm \citep{dhillon2007weighted,fagginger2012gpu}. Here, a randomly chosen node is merged with its most similar neighbor by averaging the two nodes' intermediate features. This process is repeated until all nodes have either been merged or have no unmerged neighbors. In addition to the graph neural networks as outlined above, we also consider a node-wise MLP (BaseMLP) which is similar to BaseGNN in terms of architecture, but uses linear layers instead of graph convolutions.

The described GNN architectures can be combined with a variety of operators, such as the ones from graph convolutional networks (GCN) \citep{kipf2017semi}, GraphSAGE \citep{hamilton2017inductive}, or graph attention networks (GAT) \citep{velivckovic2018graph}. These differ in the way information is propagated through the graph: GCN applies a learnable weight matrix to each node's features and then averages the results across neighborhoods. Beyond that, GraphSAGE introduces separate weight matrices for the central node and the neighbor nodes, while GAT uses an attention mechanism to model the importance of specific neighborhoods based on the node features.

Based on our experiments, we propose the use of the graph transformer convolution operator (GT) \citep{yun2019graph,shi2021masked}. In essence, GT combines the ideas of GraphSAGE and GAT by using both separate weight matrices, as well as attention-based weighting.

\subsubsection{Output-level object aggregation}
\label{sec:output-level_object_aggregation}

Conversely, output-level aggregation, as depicted in Figure \ref{fig:methods}(c) first produces a pixel-wise classification map based on a neural network for semantic segmentation. The object definition $\mathbf{S}$ is then used to postprocess the result in order to achieve an object-wise result. Particularly, the logits which are output by the last learnable layer are averaged across each segment.

As the underlying semantic segmentation model, established network architectures with different backbones from computer vision are used. In our experiments we consider different well-established CNN- and ViT-based architectures: As a simple baseline, we use a 3-layer CNN (BaseCNN) which is similar to BaseMLP and BaseGNN, as previously described, but uses standard convolution layers instead of graph convolutions or linear layers, repectively. For the more elaborate CNNs, we use UNet with a ResNet-18 backbone \citep{ronneberger2015u} and DeepLabV3 \citep{chen2017deeplab,chen2017rethinking} with a ResNet-34 backbone which are established models in the context of land cover classification and have been used extensively in related work \citep{wang2020weakly,solorzano2021land,hua2021semantic}. In addition, UNet++ \citep{zhou2018unet++} with a ResNet-18 backbone is tested to assess the effect of more complex CNN-based models. To provide a comparison with more recent ViT-based models, we experiment with Segformer in its B3 variant which relies on mix transformer (MiT) \citep{xie2021segformer}. 

The choices for the respective backbone models are justified experimentally by demonstrating that choosing higher capacity backbones does not guarantee improved performance in our given setting. Similarly, while more elaborate and recent transformer-based semantic segmentation methods and larger backbones may be considered, we generally limit ourselves to rather lightweight models in this study. This is because dataset sizes are typically comparably small in the context of sparsely labeled segmentation with in-situ data, making the problem data- rather than model parameter-constrained. As such, models do not necessarily benefit from higher parameter counts.

\subsubsection{Optimization}
\label{sec:optimization}

Due to the presence of sparse labels in our setting, the commonly used cross entropy loss function for classification is adapted accordingly. Especially, we use a partial cross entropy \citep{cicek20163d} where the loss is evaluated only at the labeled pixels. While other loss functions for the sparse supervision case exist \citep{tang2018normalized,bokhovkin2019boundary,hua2021semantic,liang2022tree}, they do not trivially apply to cases with prescribed MMUs, as in our setting which is why we view them outside the scope of this work.

We train all models for 20 epochs using the Adam optimizer with an initial learning rate of $1 \times 10^{-4}$. The accuracy on the validation set is used to select the best performing model. Additionally, the learning rate is halved each time validation accuracy does not improve for two epochs. Each model is trained three times and we compute the mean and standard deviation for each metric.

\subsection{Dataset}
\label{sec:dataset}

For training and evaluation of the above-described framework, we develop a machine learning-ready dataset specifically targeted at the task of training large-scale land cover classifiers using sparse in-situ labels. We note that earlier iterations of this dataset have been used in previous work \citep{leonhardt2024sparsely,leonhardt2025climsat}.

\subsubsection{LUCAS}
\label{sec:lucas}

Eurostat LUCAS \citep{d2020harmonised} is a program with the aim of providing coherent and harmonized land use and land cover statistics throughout the European Union. It has been running since 2006 with surveys taking place every three to four years. In each survey, land cover data, as well as other data on, e.g., land use and irrigation, are recorded on a pre-specified and constant over time $2km$ by $2km$ grid, although not all grid points contain labels for each year. For our dataset, we use LUCAS data from the 2018 survey as our sparse labels. We also restrict ourselves to the highest level of land cover semantics in LUCAS, which differentiates between the eight land cover classes \textit{artificial land}, \textit{cropland}, \textit{woodland}, \textit{grassland}, \textit{shrubland}, \textit{bare land and lichens/moss}, \textit{water}, and \textit{wetlands}.

\subsubsection{Sentinel-2}
\label{sec:sentinel-2}

Sentinel-2 is a multispectral imaging satellite of the European Space Agency's Copernicus program. We preprocess the imagery broadly following the procedure of \citet{corbane2020global} for producing analysis-ready large-scale composites. Of Sentinel-2's 12 overall spectral bands, we use the red, green, blue and near infrared bands, all of which are at a $10m$ resolution. we create annual composites for the year 2018 and for each EU-28 country. To this end, we first collect all Sentinel-2 images, processed to level 2A, which were recorded within the specified year, and whose footprint intersects with the specified area. We then mask all pixels which depict clouds, cloud shadows, or defective pixels, as indicated by the images' scene classification (SCL) band. After unifying the coordinate reference system of all images, we then perform reduction over time by selecting the 25-th percentile of all unmasked pixels for a specific location. Finally, the pixel values are normalized by dividing the pixel values by 3000 for the red, green, and blue channel, and 7000 for the near infrared channel. The result is finally clipped to the range $[0,1]$.

\subsubsection{Third-party land cover products}
\label{sec:third-party_land_cover_products}

For additional analysis, we also collect data from two high resolution annual land cover products, ESRI Land Cover 2018 (ESRI-LC) \citep{karra2021global} and ESA WorldCover 2020 (ESA-WC) \citep{zanaga2021esa}. Besides being two of the most popular land cover products, the two datasets are especially interesting to consider due to the large difference regarding the machine learning frameworks used to produce them: While ESRI-LC is generated using a conventional CNN, trained on photointerpretation-based labels, ESA-WC uses pixel-wise classification with labels from the crowdsourced GeoWiki dataset \citep{fritz2012geo,fritz2017global}. We include ESA-WC in our analysis despite of the temporal mismatch, as it has previously been demonstrated to have good agreement with LUCAS 2018 data \citep{venter2022global,xu2024comparative}. Note that in our experiments, we never use these data products to directly train the land cover classifiers. They are used solely for evaluation and for training the feature extractor. For the purposes of this study, the class definitions of both ESA-WC and ESRI-LC are adapted to match the LUCAS classes, as described in Section \ref{sec:lucas}.

\subsubsection{Dataset generation}
\label{sec:dataset_generation}

Sentinel-2 and LUCAS data are harmonized in a way that allows for training, validating, and testing deep learning models for land cover classification. To this end, we extract a $64 \times 64$ pixel Sentinel-2 patch around each LUCAS point such that the land cover label can be related to exactly one random pixel position within the patch. To guarantee sufficient spatial context, we ensure that a label is not within a $5px$ boundary of the patch's border. The same processing pipeline is also used for the third-party land cover products, ensuring an exact spatial alignment between ESA-WC, ESRI-LC, and Sentinel-2.

We find that such a random arrangement of the labels within the image patches is crucial for training land cover classification models. Especially, when using a fixed pixel position for the label, we found that models overfit to this position and are not able to reliably classify land cover in other image locations. Notably, this limitation is present in the recently proposed dataset of \citet{sharma2024sen4map} which always places the label in the center of the image patch. 

\begin{figure}[]
    \centering
    \includegraphics[scale=1.0]{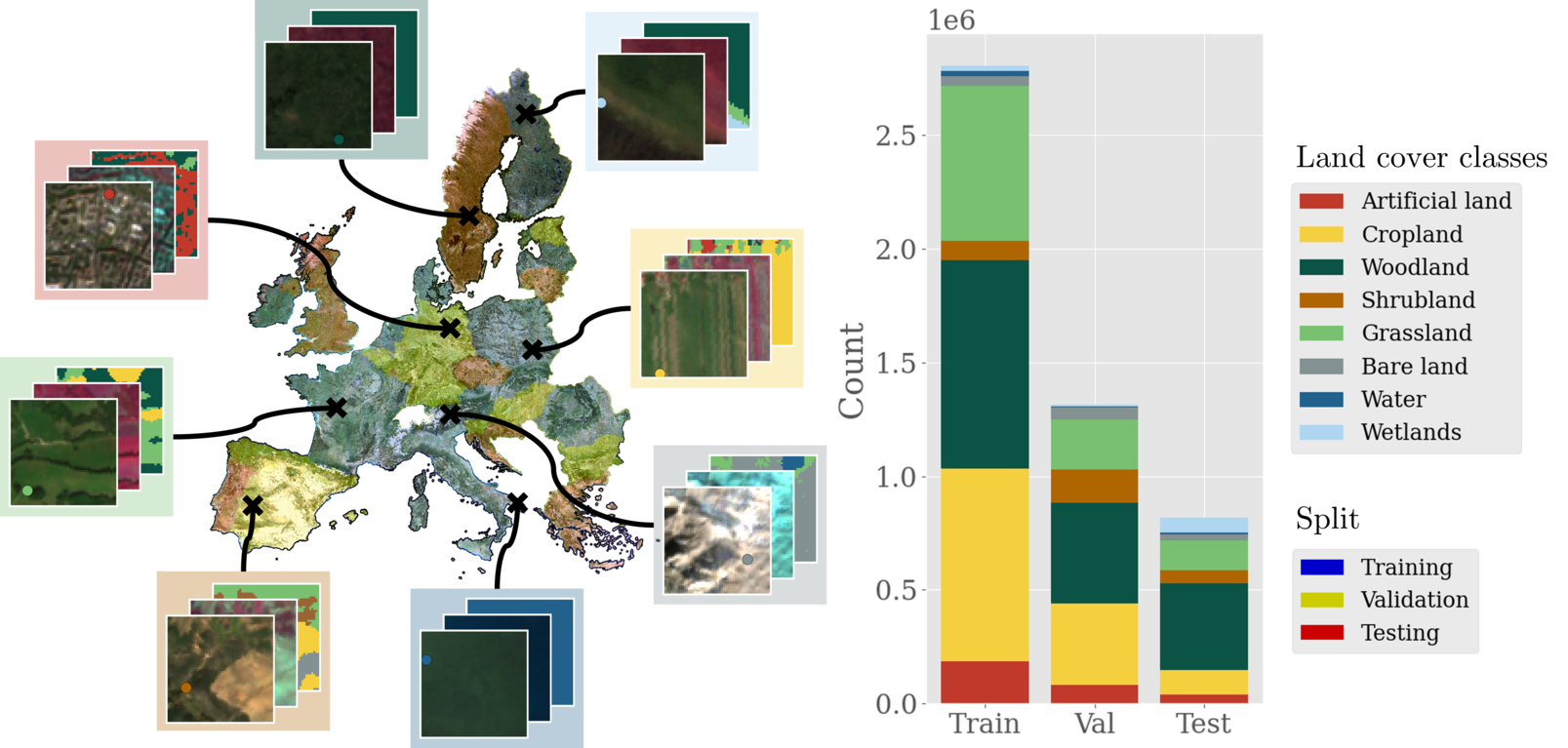}
     \caption{Geographical coverage, data split, class distribution, and one sample for each land cover class from the presented dataset for land cover classification with sparse labels. Best viewed zoomed in.}
    \label{fig:data}
\end{figure}

In total we obtain 342330 samples, which are geographically divided into training ($48.5\%$ of samples), validation ($28.5\%$ of samples) and test ($23.0\%$ of samples) split, ensuring that samples from the same country remain in the same split. Examples for the data samples, as well as the split are depicted in Figure \ref{fig:data}.

\subsubsection{Variation of dataset size}
\label{sec:dataset_variants}

In addition to the full dataset, as described above, we also consider subsets which are reduced in terms of sample size in our experiments. This way, we simulate scenarios where less data is available for training, which further limits the applicability of deep learning models. For example, accuracies of African land cover products still lack behind those of Europe and the USA due to a lack of data \citep{kerner2024accurate}. It is thus an important methodological challenge to use the available data as efficiently as possible. 

In our experiments, we use randomly sampled subsets of the full dataset which correspond to $1/2$, $1/4$, $1/8$, and $1/16$ of the full dataset in terms of size. We ensure that the datasets are hierarchical, meaning that the $1/16$ dataset is a subset of the $1/8$ dataset, and so on. Importantly, all models are still evaluated on the full testing set regardless of which dataset variant it was trained on to ensure comparability of the results.

\subsection{Evaluation metrics}
\label{sec:evaluation}

Regarding the evaluation of the land cover classification, we report accuracy metrics, which assess how well the result matches the provided sparse labels. However, since we also want to assess the effect of prescribing a MMU in terms of fragmentation of the resulting land cover maps, we also compute metrics to quantify this effect.

\subsubsection{Accuracy metrics}

Accuracy-based metrics assess how well the predicted land cover classes match the corresponding LUCAS labels. Due to the label sparsity, accuracy can only be evaluated at individual pixel locations, as described in Section \ref{sec:dataset_generation}. 

We report both overall accuracy, which provides a general measure of correctness, as well as the F1 score, which balances precision and recall to provide a more robust metric in the presence of class imbalance, as is the case in our dataset. 

In addition to strict pixel-wise evaluation, we also use a relaxed version of both metrics that allows for a tolerance $t$ of one pixel: In this case, the predicted class is considered correct if it matches the LUCAS label at the pixel or any of its immediate neighbors. This accounts for minor boundary misalignments, which are particularly relevant to take into account when evaluation on sparse labels because of possible georeferencing inaccuracies \citep{weigand2020spatial,xu2024comparative}.

\subsubsection{Fragmentation metrics}

To assess the tradeoff between accuracy and fragmentation for differently specified MMUs, complexity-based metrics are used to evaluate the structural characteristics of the classification output. In particular, these metrics are largely borrowed from the field of landscape ecology and are commonly used to quantify a landscape's configuration \citep{turner2015landscape}.

First, patch density describes the number of connected label regions per image patch with lower values indicating less fragmented landscapes and more cohererent classification. Edge density captures the proportion of pixels lying on the boundary between different classes, reflecting how smooth or jagged the segment transitions are. Finally, entropy quantifies the uncertainty or heterogeneity of label distributions across the image; high entropy suggests more diverse or fragmented segmentations, while low entropy implies stronger spatial coherence. These complexity metrics complement accuracy by characterizing the visual quality and structural compactness of the resulting land cover map.

\subsection{Implementation details}
\label{sec:implementation_details}

All considered models (BaseMLP, BaseGNN, GUNet, UNet, UNet++, DeepLabV3, and Segformer) are trained on all variants of our dataset (full, $1/2$, $1/4$, $1/8$, and $1/16$) using the different object definitions (MMU = $1px$, $5px$, $10px$, $20px$, and $40px$). This allows us to systematically assess how the MMU on the one hand and the dataset size on the other affect the classification performance for different models.

An important detail of using established implementations of semantic segmentation models for output-level object aggregation is the need to increase the input image size to achieve satisfactory classification performance. This requirement likely stems from the initial downsampling in the CNN-based ResNet backbones and the fixed patch-based processing in ViTs, both of which reduce spatial resolution early in the network. Without resizing, this can lead to overly coarse feature representations and degraded classification accuracy. To address this, we upscale the input patches from $64 \times 64$ to $224 \times 224$ pixels using bilinear interpolation before feeding them into the network and restore the initial shape prior to the final prediction layer.

The pre-trained model for feature extraction is trained using the pixel-wise ESA-WC data as pseudo-labels. Besides an extended training time of 50 epochs, we use the same training strategy as for the model trained on the sparse labels, as described above.

The trained models are evaluated based on the accuracy and fragmentation metrics, as described above. To this end, we do not consider the $5px$ boundary of the patch border. This is because, as mentioned in Section \ref{sec:dataset_generation}, no LUCAS labels were placed there during training resulting in incoherent classification for these areas of the image patch.

The implementation of LC-SLab relies on Pytorch \citep{paszke2019pytorch} and Pytorch lightning as the overall deep learning frameworks, pytorch-geometric \citep{fey2019fast} for GNNs, segmentation-models-pytorch for CNNs, huggingface transformers \citep{wolf2020transformers} for ViTs, scikit-image \citep{van2014scikit} for object definition, and torchmetrics \citep{detlefsen2022torchmetrics} for evaluation. 

The Sentinel-2 imagery, as well as the third-party products were accessed using Google Earth Engine \citep{gorelick2017google}. The LUCAS data were accessed through the Eurostat website\footnote{https://ec.europa.eu/eurostat/web/lucas/data/primary-data/}.

\section{Results}
\label{sec:results}

In this section, we first present the results for the proposed configurations of LC-SLab for different MMUs in Section \ref{sec:effect_of_mmu}. Afterwards, we study the effect of dataset size, as well as the impact of incorporating features from the pre-trained model in Sections \ref{sec:effect_of_dataset_size} and \ref{sec:effect_of_features_from_pre-trained_models}, respectively. In Section \ref{sec:comparison_to_third-party_land_cover_products}, we compare our framework's results to existing land cover products. Finally, our design choices for the object definition module, as well as the input-level and output-level object aggregation models are justified in the ablation studies in Section \ref{sec:ablation_studies}.

In the descriptions of our results, we primarily focus on the metrics overall accuracy without tolerance and patch density. This is because we found these metrics to be most robust across different models. However, the presented results and our conclusions are generally consistent and representative of the models' behavior across different accuracy and fragmentation metrics. The detailed results for all metrics are provided as tables the appendix.

\subsection{Effect of MMU}
\label{sec:effect_of_mmu}

To provide an impression of the effect of different MMU specifications, an overall qualitative comparison between LC-SLab models for differently specified MMUs, as well as third-party products is provided in Figure \ref{fig:qualitative}.

\begin{figure}[]
    \centering
    \includegraphics[scale=1.0]{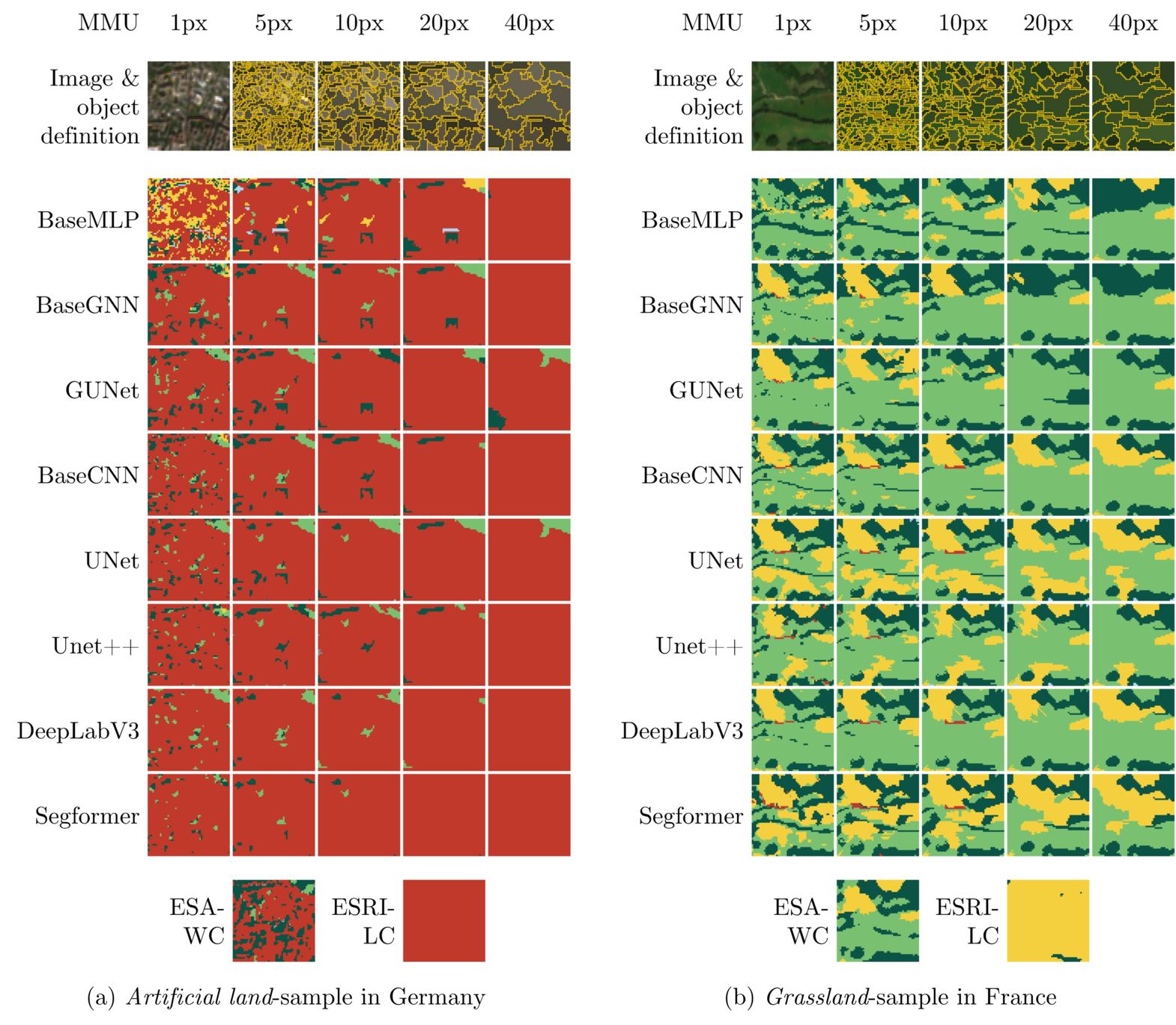}
    \caption{Qualitative comparison of results from different object-based deep learning classifiers for selected samples. Best viewed zoomed in.}
    \label{fig:qualitative}
\end{figure}
\begin{figure}[]
    \ContinuedFloat
    \centering
    \includegraphics[scale=1.0]{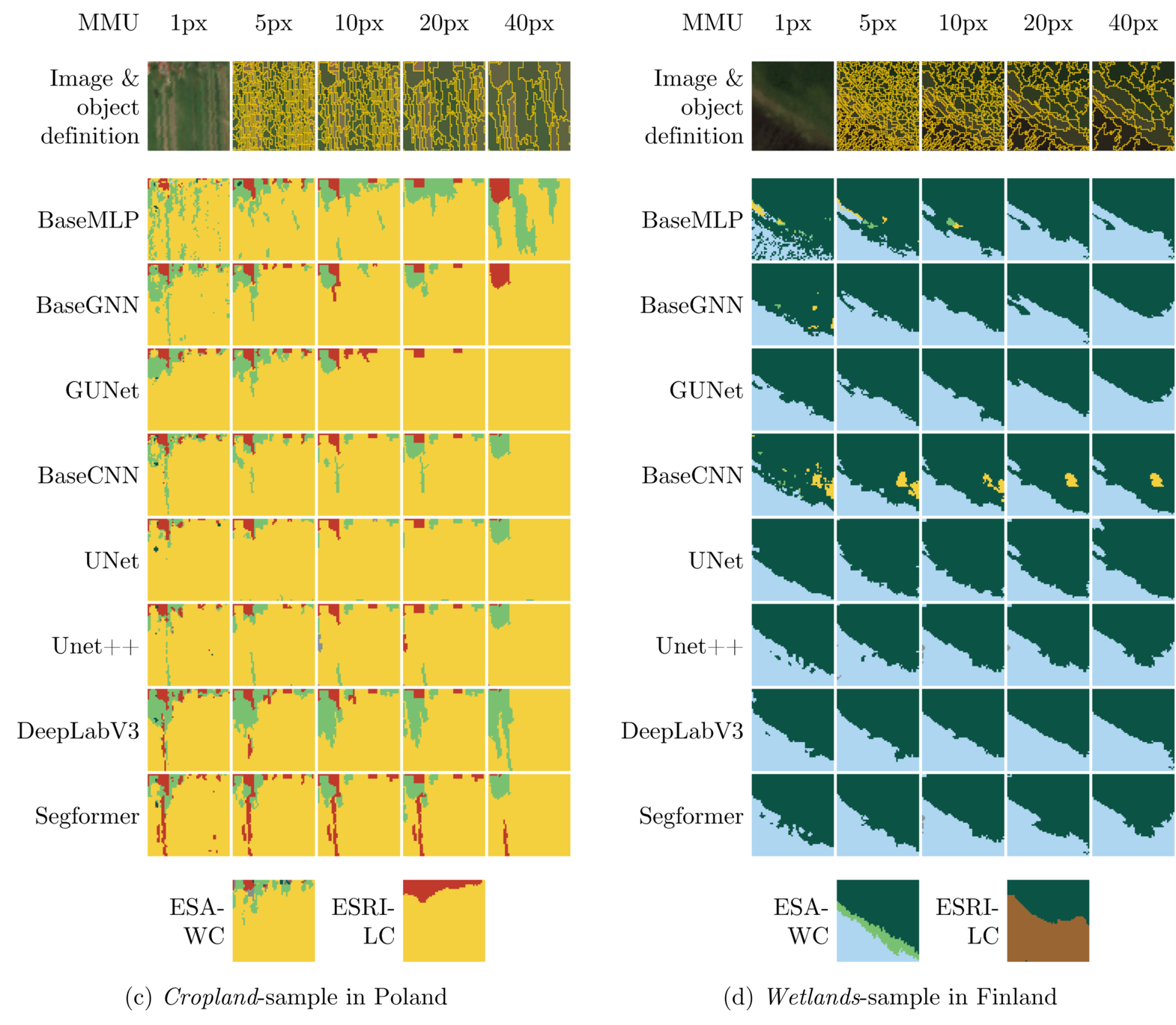}
    \caption{Continued}
\end{figure}

\begin{figure}[]
    \centering
    \includegraphics[scale=1.0]{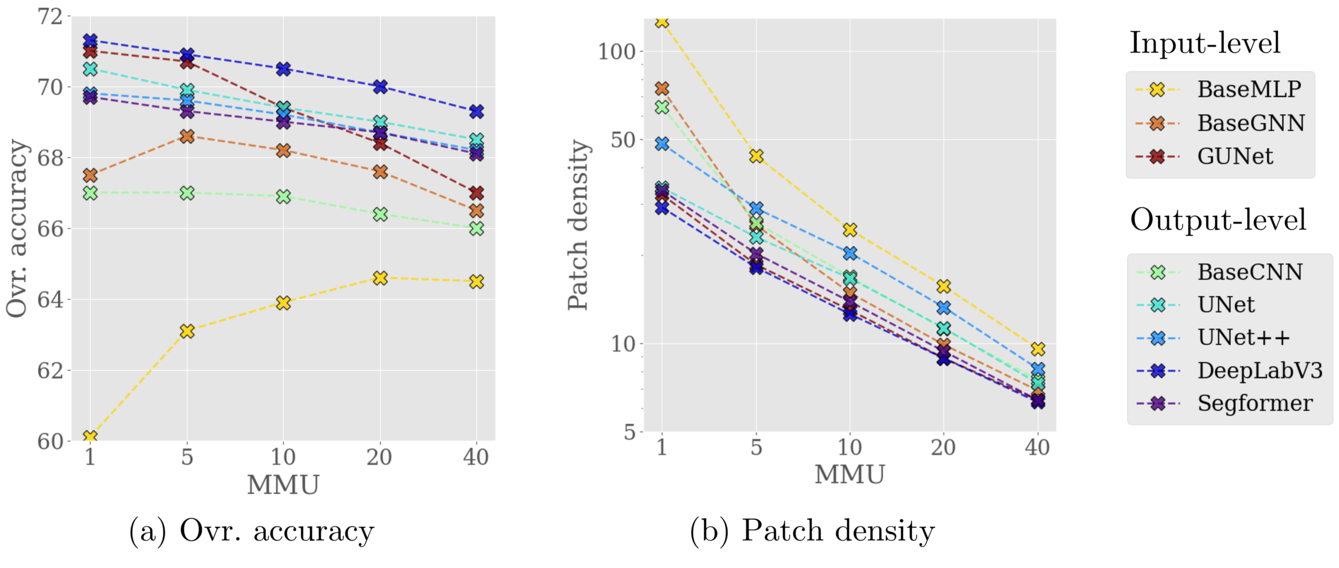}
    \caption{Overall accuracy and patch density achieved by different methods with different MMU specifications. Best viewed zoomed in.}
    \label{fig:effect_of_mmu}
\end{figure}

\begin{figure}[]
    \centering
    \includegraphics[scale=1.0]{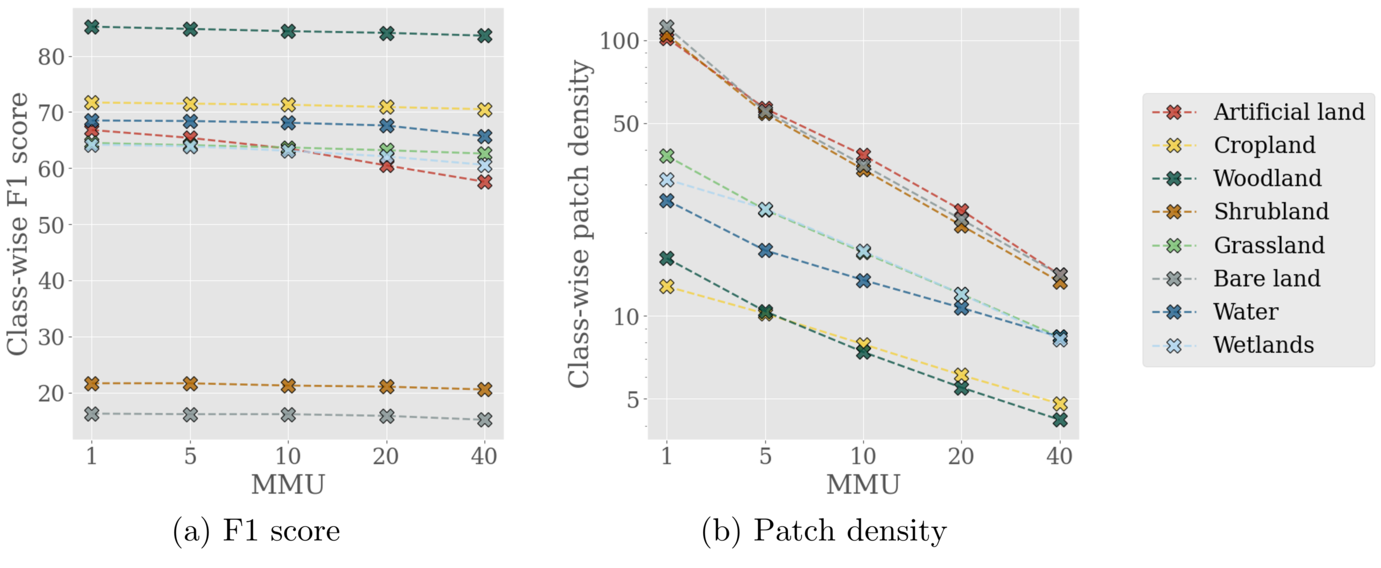}
    \caption{Class-wise F1 scores and patch densities achieved by the DeepLabV3 model with different MMU specifications. Best viewed zoomed in.}
    \label{fig:effect_of_mmu_class-wise}
\end{figure}

In addition, we compare the models' overall accuracies and the results' patch densities for the different MMUs. The summarized results are displayed in Figure \ref{fig:effect_of_mmu}. Additionally, we report the class-wise F1 scores and patch densities for the DeepLabV3 model in Figure \ref{fig:effect_of_mmu_class-wise}. For more detailed results, we refer to Appendix A.

For most models, we observe that both overall accuracy and patch density decrease for higher MMUs. Especially, this is true for the models which perform well when no MMU is specified, such as DeepLabV3 and GUNet, which are the best overall performers. For these models, it is notable that patch density is already decreased by about half for most models even when prescribing a relatively modest MMU of $5px$ while the accuracy suffers by less than $1\%$. On the other hand, the small base models generally benefit from the specification of a MMU also in terms of accuracy.

These results are confirmed by the qualitative comparison in Figure \ref{fig:qualitative}. Especially, the results when a MMU is enforced are more semantically consistent, better adhere to object boundaries, and avoid isolated stray pixels. In the setting with a MMU of $40px$, however, we start to see that some objects such as patches of woodland in Figure \ref{fig:qualitative}(a) are no longer captured by the object definition.

Moreover, both accuracy and fragmentation greatly vary across land cover classes for the best-performing DeepLabV3 model. In particular, the highest F1 score is achieved for \textit{woodland} at over $80\%$ while the classes \textit{shrubland} and \textit{bare land} achieve F1 scores of only about 20\%. Regarding patch density, artificial land stands out as being the most fragmented land cover class while woodland and cropland exhibit the lowest patch densities.

\subsection{Effect of dataset size}
\label{sec:effect_of_dataset_size}

We next study the effect of the dataset size on the different models' performances. For different subsets of the dataset, the overall accuracy is depicted in Figure \ref{fig:effect_of_ds_size}. The detailed results for all metrics are reported in Appendix B.

\begin{figure}[]
    \centering
    \includegraphics[scale=1.0]{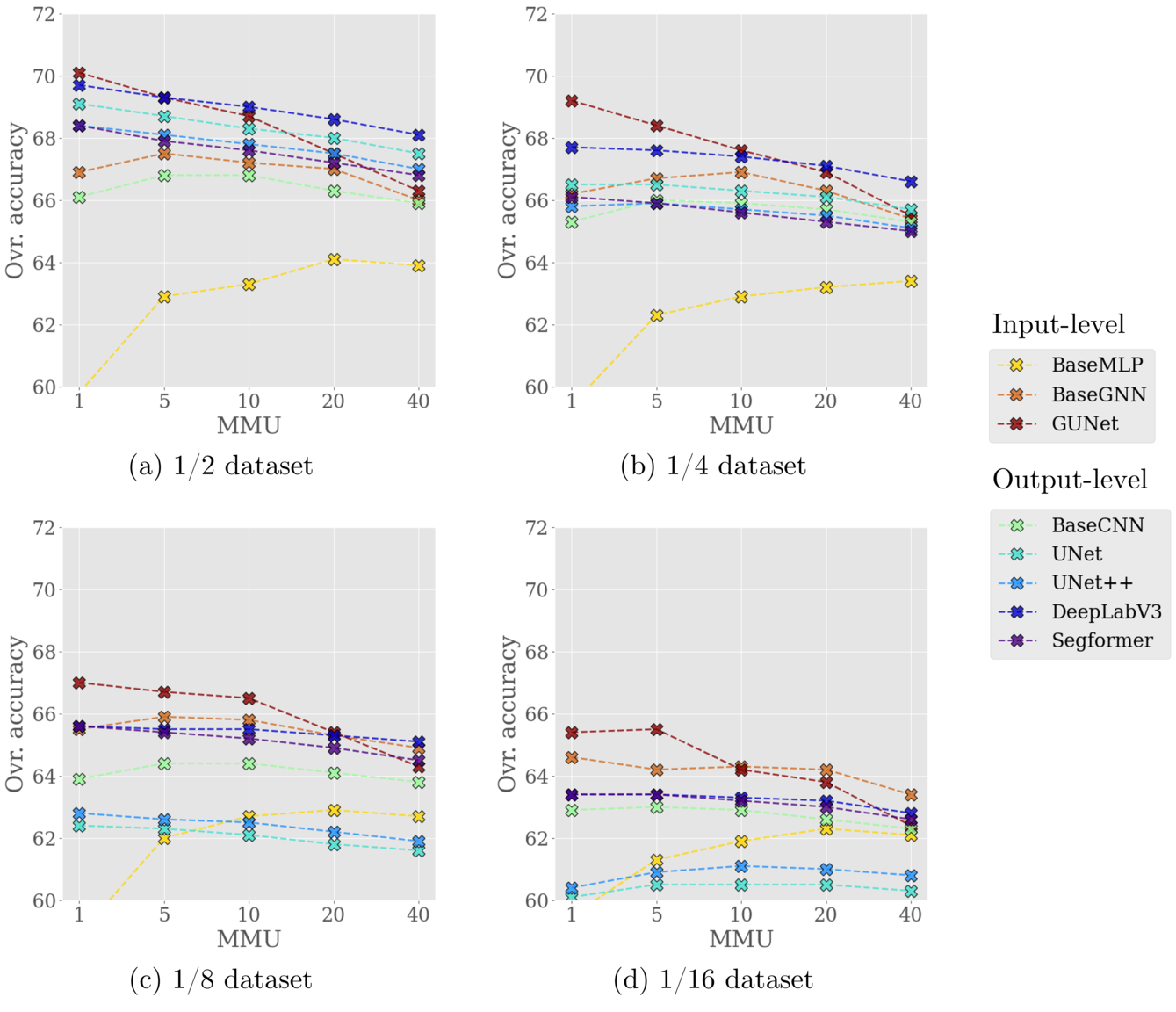}
    \caption{Overall accuracy of the different methods plotted against MMU for reduced dataset sizes. Best viewed zoomed in.}
    \label{fig:effect_of_ds_size}
\end{figure}

Generally, we expectedly observe that accuracy suffers from smaller training datasets. However, input-level object aggregation methods generally seem to be less affected by this effect that output-level aggregation methods. Whereas DeepLabV3 outperforms the other methods when the full dataset is used for training, GUNet and BaseGNN achieve the highest accuracies for the smallest variant of the dataset. In particular, DeepLabV3's accuracy is decreased by about $8\%$ when training on the $1/16$ dataset, while GUNet only suffers an average accuracy loss of about $5\%$. 

Similarly, small models demonstrate a stronger robustness with respect to the dataset size. For example, while BaseMLP performs the worst out of all models for the larger dataset variants, it performs better than UNet and UNet++ when trained on the smallest considered dataset.

\subsection{Effect of features from pre-trained models}
\label{sec:effect_of_features_from_pre-trained_models}

We next assess the effect of using features from the model pre-trained on the ESA-WC pseudo-labels as the inputs to the models. In initial testing, we found that in this setting, using comparably large models such as GUNet for input-level object aggregation or DeepLabV3 for output-level object aggregation did not provide an advantage over the smaller models. Hence, we only present results for BaseMLP, BaseGNN, and BaseCNN. The summarized results are presented in Figure \ref{fig:effect_of_features_from_pre-trained_models}. The detailed results can be found in Appendix C.

\begin{figure}[]
    \centering
    \includegraphics[scale=1.0]{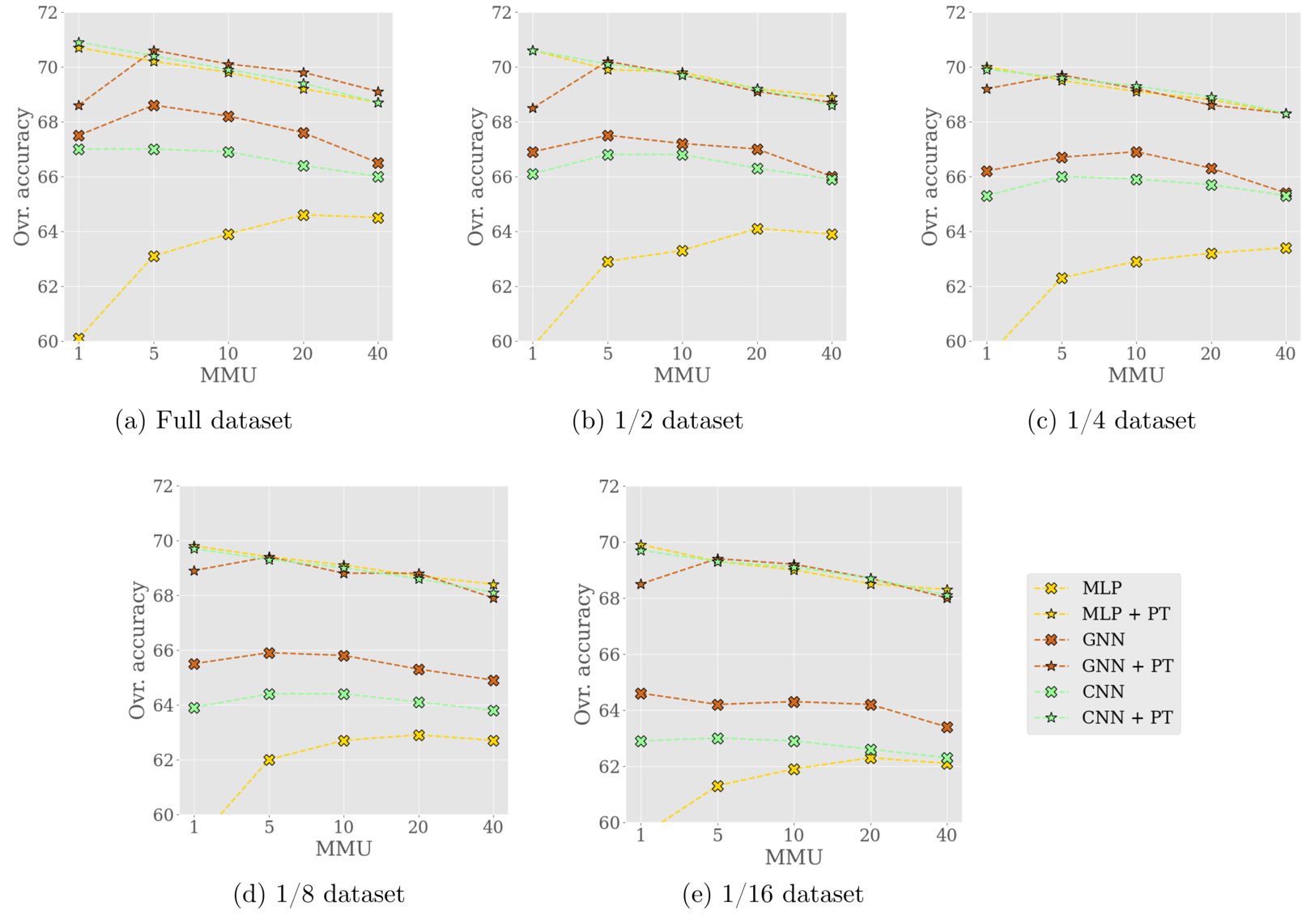}
    \caption{Overall accuracies achieved when using features from a pre-trained model (+PT) as input to the object-based deep learning classifiers. Best viewed zoomed in.}
    \label{fig:effect_of_features_from_pre-trained_models}
\end{figure}

The small models perform significantly better than on the raw intensity features. Notably, the differences between the different classifiers become insignificant when training on features from the pre-trained model. 

Furthermore, the models' performances do not suffer strongly from decreasing the dataset size: For BaseGNN for example, the difference in terms of overall accuracy between the full and the $1/16$ dataset is about $5\%$ when the raw intensity features are used. On the other hand, only $1\%$ is lost when using pre-trained features when comparing the models trained on the full dataset and the ones trained on the $1/16$ dataset.

\subsection{Comparison to third-party land cover products}
\label{sec:comparison_to_third-party_land_cover_products}

To evaluate the advantages of the proposed LC-SLab framework compared to existing products, we compare our models' results to the well-established ESA-WC and ESRI-LC products, as described in Section \ref{sec:third-party_land_cover_products}. These products do not enforce a MMU by design. For a meaningful comparison we therefore compare them to our models by plotting the accuracy in relation to the fragmentation as measured by the patch density. In particular, we compare to GUNet and DeepLabV3 using the raw intensity features which were the best-performing input- and output-level feature aggregation models when training on the full dataset, respectively. The results are shown in Figure \ref{fig:comparison_to_third-party_land_cover_products}. Details are provided in Appendix D.

\begin{figure}[]
    \centering
    \includegraphics[scale=1.0]{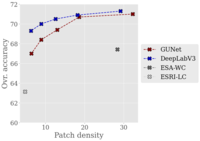}
    \caption{Overall accuracies achieved at different patch densities for the best configurations of LC-SLab and third-party products. Best viewed zoomed in.}
    \label{fig:comparison_to_third-party_land_cover_products}
\end{figure}

We observe that both the best input-level object aggregation approach and the best output-level object aggregation approach provide better accuracies at their respective levels of fragmentation than the established third-party products. In this context, ESA-WC has a high patch density, indicating a high degree of fragmentation. On the other hand, ESRI-LC, has a low patch density, but is also significantly less accurate.

This impression is again confirmed by the qualitative results from Figure \ref{fig:qualitative}. Here, we see that ESA-WC is most similar to our results for no or small MMUs while ESRI-LC is more similar to results for high MMUs. However, ESRI-LC in contrast to LC-SLab does not provide guarantees in terms of MMU. Because of this, stray pixels are still present in the resulting land cover maps despite the low degree of fragmentation, as can be seen for example in Figure \ref{fig:qualitative}(b).

\subsection{Ablation studies}
\label{sec:ablation_studies}

Finally, we conduct several ablation studies to justify certain algorithmic choices of the LC-SLab framework.

\subsubsection{Object definition}
\label{sec:ablation_object_definition}

We compare the overall accuracies when using the FH algorithm to the setting where SLIC is used for the object definition with the same MMUs. The results are summarized in Figure \ref{fig:ablation_object_definition}. The detailed results are listed in Appendix E.

\begin{figure}[]
    \centering
    \includegraphics[scale=1.0]{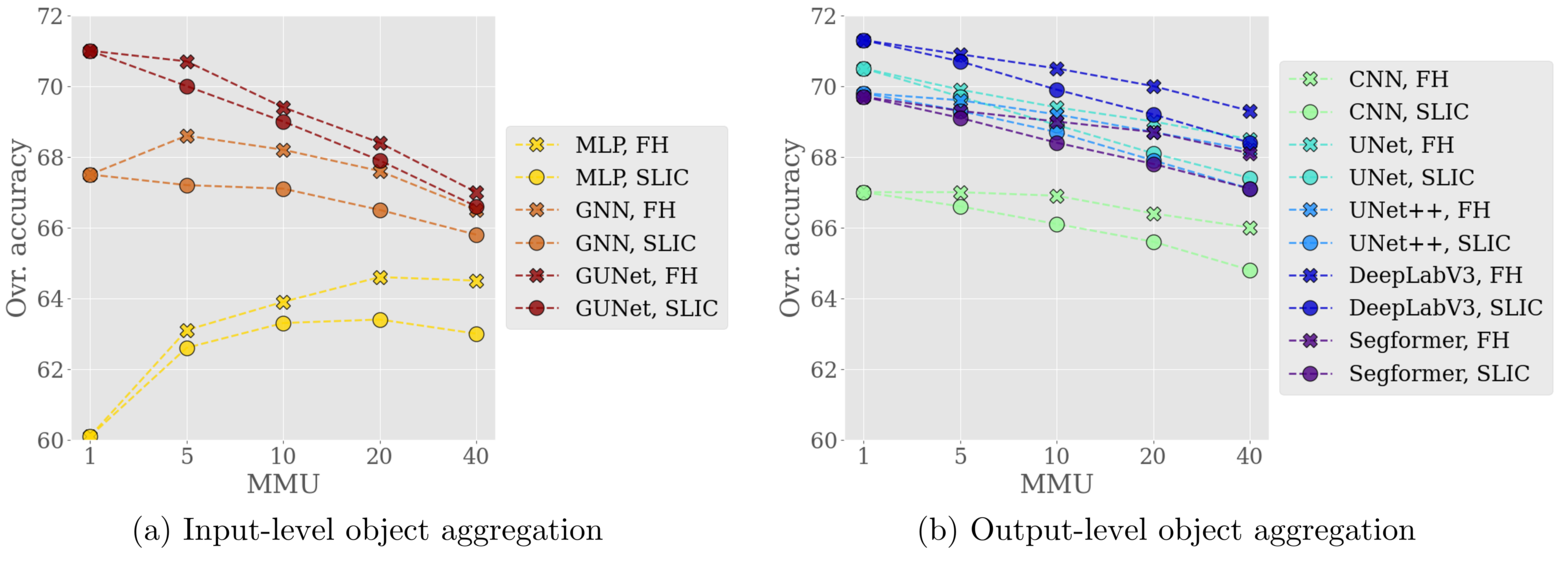}
    \caption{Overall accuracies achieved by different models when using the FH and the SLIC algorithm for object definition and specification of the MMU. Best viewed zoomed in.}
    \label{fig:ablation_object_definition}
\end{figure}

We observe that for all models the use of the FH algorithm results in higher accuracies compared to when SLIC is used. The difference generally becomes more significant for higher MMUs.

\subsubsection{Graph convolution operators in input-level object aggregation}
\label{sec:graph_convolution_operators_in_input-level_object_aggregation}

\begin{figure}[]
    \centering
    \includegraphics[scale=1.0]{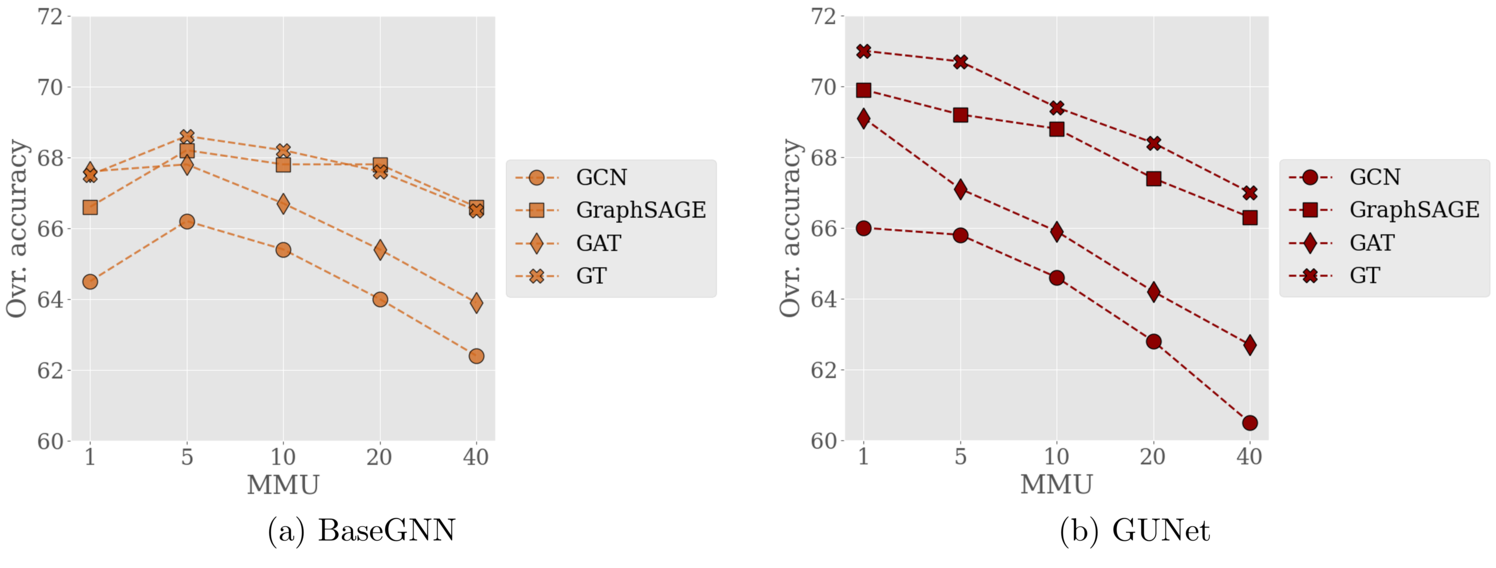}
    \caption{Overall accuracies achieved by different input-level object aggregation settings using different graph convolution operators on the full dataset. Best viewed zoomed in.}
    \label{fig:ablation_graph_convolution_operators}
\end{figure}

For input-level aggregation, the achievable accuracy significantly differs depending on which graph convolutional operator is selected. The detailed results from Appendix F are summarized in Figure \ref{fig:ablation_graph_convolution_operators}. 

Overall, the suggested GT operator outperforms the more commonly used GCN, GraphSAGE and GAT operators. GraphSAGE is the next best option while GCN performs the worst. This observation is consistent across both BaseGNN and GUNet.

\subsubsection{Features in input-level object aggregation}
\label{sec:features_in_input-level_object_aggregation}

\begin{figure}[]
    \centering
    \includegraphics[scale=1.0]{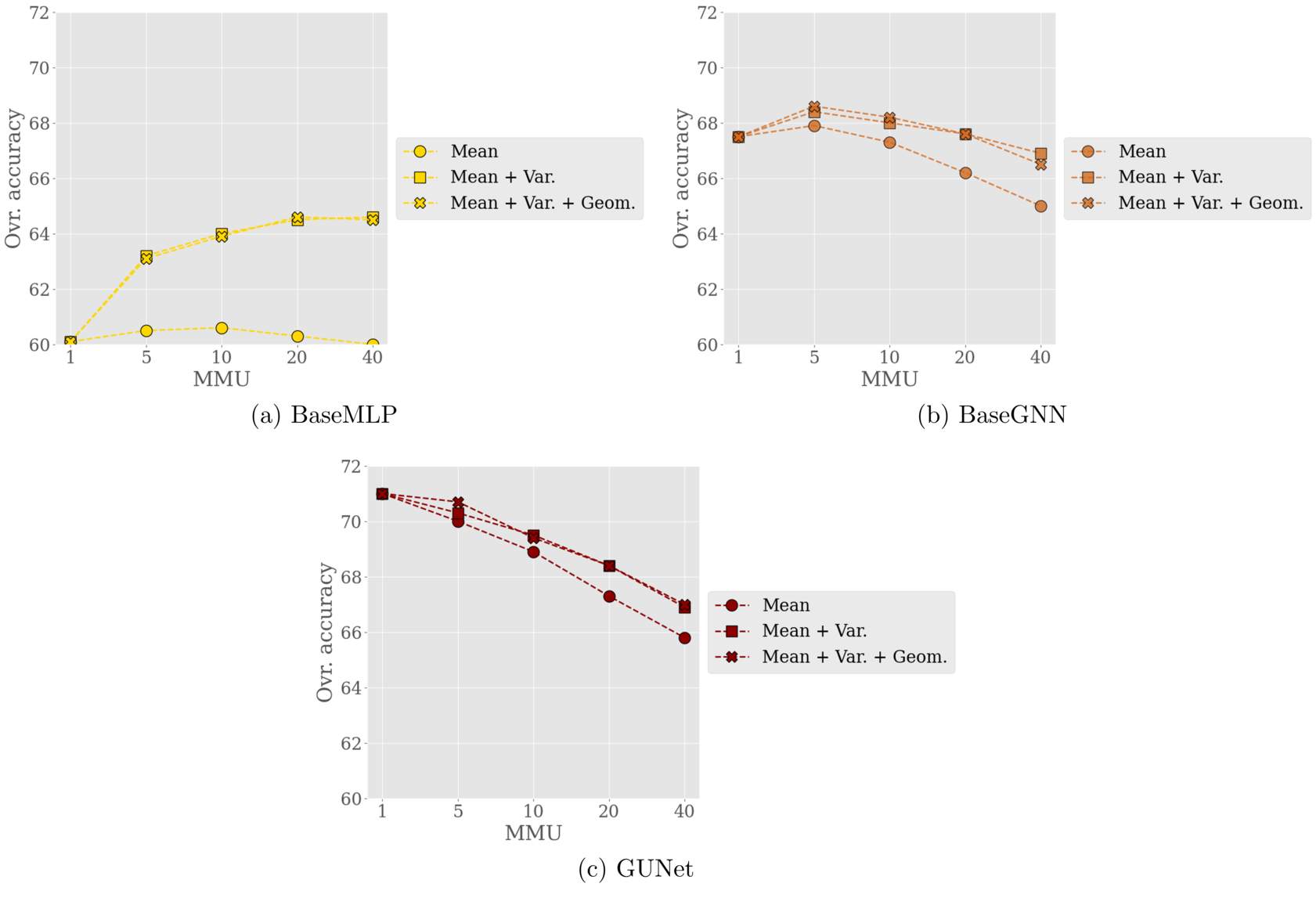}
    \caption{Overall accuracies achieved with different node features derived for input-level object aggregation methods on the full dataset. Best viewed zoomed in.}
    \label{fig:ablation_features}
\end{figure}

We conduct an additional ablation study to quantify the contribution of the different features used in input-level object aggregation. Models were trained and evaluated using three incremental feature sets. These sets consist of spectral intensity means only, spectral means with intensity variability, and the full feature set including geometric descriptors, respectively. Figure \ref{fig:ablation_features} displays the resulting overall accuracies for the different settings for different MMUs. Details are provided in Appendix G.

The results show that the incorporation of variability features leads to significant improvements with regard to the achievable overall accuracy. On the other hand, we find that the geometric features have a negligible effect.

\subsubsection{Segmentation backbones in output-level object aggregation}
\label{sec:segmentation_backbones_in_output-level_object_aggregation}

For output-level object aggregation, we justify the choice of backbone for each considered semantic segmentation model. The results are shown in Figure \ref{fig:ablation_studies_output-level} and listed in Appendix H.

\begin{figure}[]
    \centering
    \includegraphics[scale=1.0]{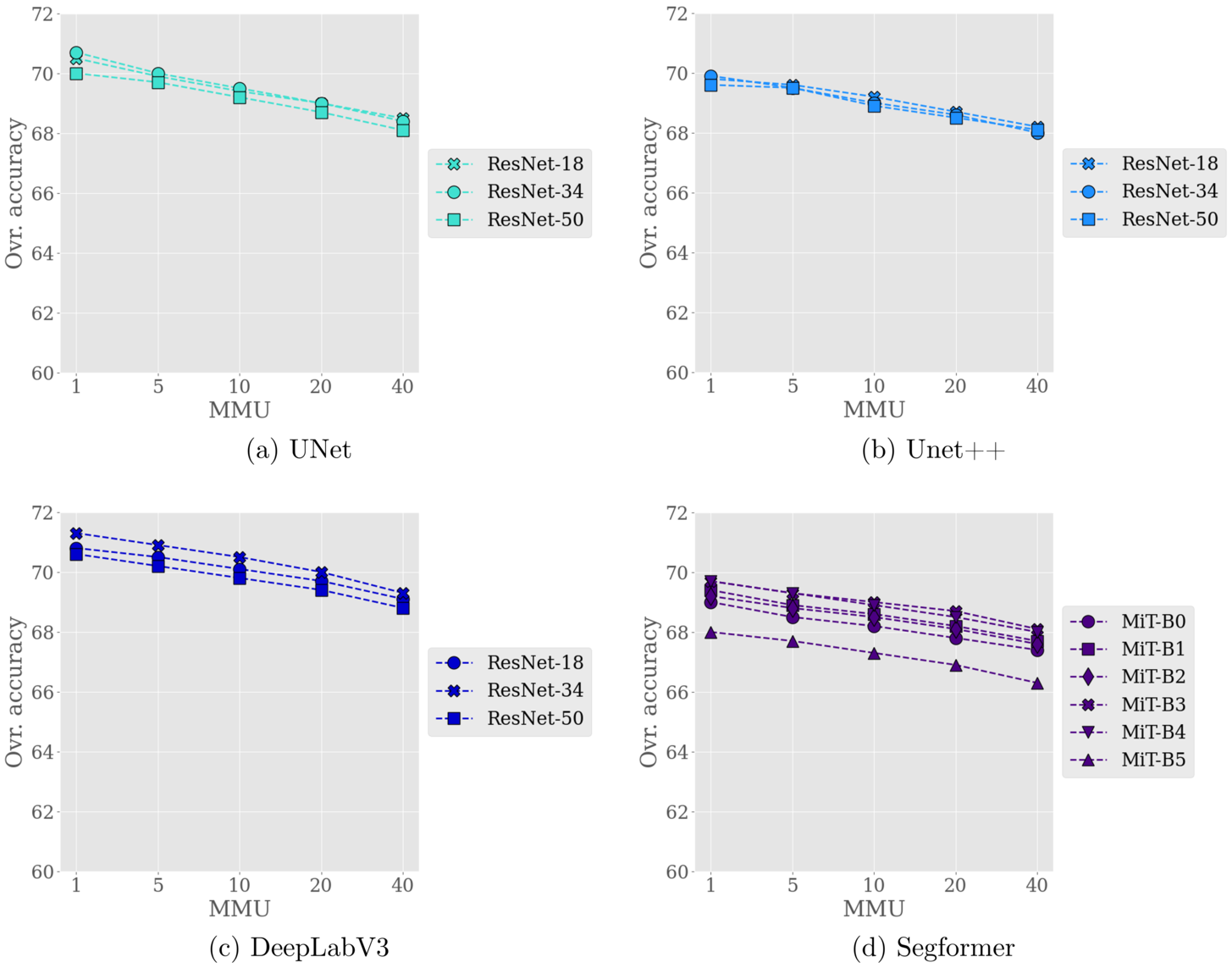}
    \caption{Overall accuracies achieved with different segmentation backbones for output-level object aggregation methods on the full dataset. Best viewed zoomed in.}
    \label{fig:ablation_studies_output-level}
\end{figure}

For UNet and UNet++, there is no significant difference between using a ResNet-18 and ResNet-34 backbone. Hence, the more lightweight option was chosen. For DeepLabV3, ResNet-34 performs the best. Using the larger ResNet-50 backbone on the other hand lead to worse results across all CNN-based models. For Segformer, we observe the largest variation between different backbones with the B3 backbone performing best and the largest B5 backbone performing worst.

\subsubsection{Rescaling in output-level object aggregation}
\label{sec:rescaling_in_output-level_object_aggregation}

\begin{figure}[]
    \centering
    \includegraphics[scale=1.0]{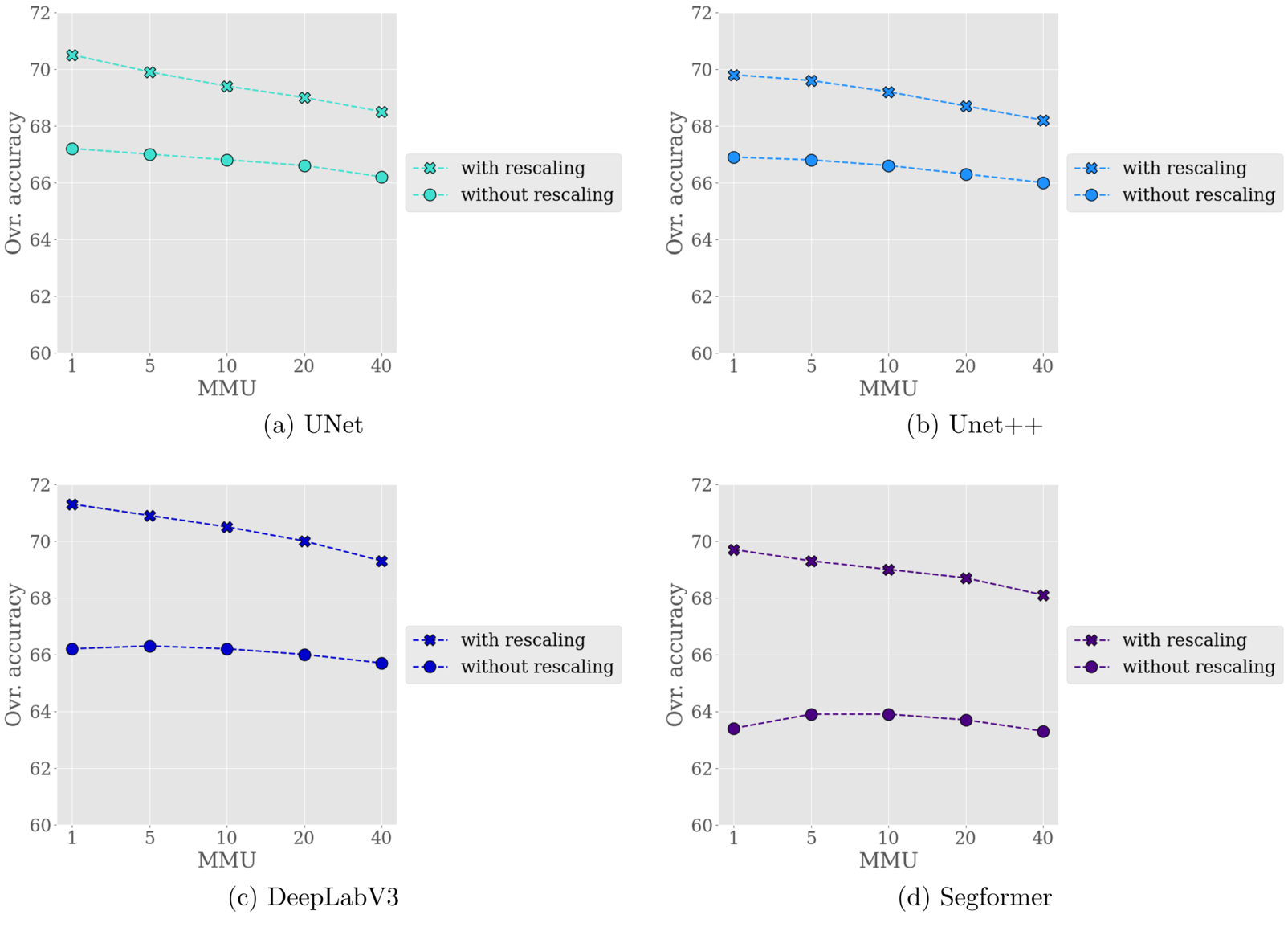}
    \caption{Effect of rescaling in output-level object aggregation on overall accuracy. Best viewed zoomed in.}
    \label{fig:ablation_rescaling}
\end{figure}

Lastly, the effect of the rescaling from $64 \times 64$ to $244 \times 244$ pixels applied to the images before being fed into the segmentation models used in output-level object aggregation is studied. The results in Figure \ref{fig:ablation_rescaling} and Appendix I demonstrate that this simple preprocessing operation leads to very significant improvements regarding overall accuracy.

\section{Discussion}
\label{sec:discussion}

The results of the presented LC-SLab framework from Section \ref{sec:results} offer several novel and meaningful insights into the potentials of object-based deep learning for large-scale land cover classification with sparse in-situ labels which we will discuss in this section.

\subsection{Object-based deep learning for spatial regularization}
\label{sec:object-based_deep_learning_for_spatial_regularization}

When considering the results where no MMU is specified, we find that while accuracy is not necessarily negatively affected, the models struggle with boundary delineation. This is indicated by the high fragmentation metrics, especially the patch density, for this setting. We also note that fragmentation in this setting is often highly variable across the three model runs, as indicated by the corresponding standard deviations provided, e.g., in Appendix A. This indicates that the models struggle to achieve consistent optimization minima during training without further regularization. This interpretation is consistent with other works who report that sparse labeling leads to poor object delineation and spatial irregularities \citep{bearman2016whats,alonso2019coralseg,hua2021semantic,chan2025sparsea}. This work therefore underlines that additional regularization is recommended when training on sparse labels.

At the same time, the results suggest that the specification of a MMU combined with object-based deep learning approaches helps to mitigate the fragmentation issue effectively.

Overall, we show that across different methods and variants of the presented dataset, object-based methods provide an effective tradeoff between accuracy and fragmentation by enforcing a MMU as determined by the object definition. In all tested settings, fragmentation is significantly reduced at the cost of no or a small loss in terms of accuracy. 

As mentioned above, the tradeoff arises from the fact that larger MMUs lead to a violation of the assumption that all pixels belonging to the same object share a land cover class which in turn negatively effects accuracy. 

For example, in the best-performing DeepLabV3 model, the specification of a small MMU of $5px$ lead to a drop of accuracy by less than half of a percentage point, while fragmentation as measured by patch density was reduced by about $40\%$. For a MMU of $40px$, overall accuracy dropped by $2$ percentage points accompanied by a $80\%$ drop in fragmentation. In some other cases, accuracy even improved when a MMU was enforced. Especially, this was the case for models which otherwise tended towards very fragmented results. For these models, the additional spatial context seems to positively affect the accuracy despite the additional restriction of the MMU.

Finally, the results in Section \ref{sec:effect_of_mmu} show that the different land cover classes considered in this study are affected differently by fragmentation: While classes made up of small spatial structures such as \textit{artificial land} are more severely affected, spatially homogeneous land cover classes like \textit{cropland} or \textit{water} are less fragmented by a factor of almost 10. Considering the different land cover classes, we find again that this effect is especially pronounced in the context of land cover classes like \textit{artificial land} where such a violation is more likely due to the small scale structures.

\subsection{Input-level vs. output-level object aggregation}
\label{sec:input-level_vs_output-level_object_aggregation}

When comparing the two different paradigms for object-based deep learning methods, we find the output-level object aggregation methods which rely on established semantic segmentation models to slightly outperform the input-level object aggregation when training on the full dataset, as shown in Section \ref{sec:effect_of_mmu}. We attribute this observation to the comparably weak inductive biases of the semantic segmentation models adapted from computer vision which allow them to benefit from abundant training data more effectively.

However, when training on the reduced datasets, as highlighted in Section \ref{sec:effect_of_dataset_size}, input-level object aggregation provided better accuracies across differently specified MMUs. For small MMUs, this effect is already seen when dataset size is reduced by $1/2$. For the smallest considered subdataset, input-level object aggregation performs best across all specified MMUs.

This observation suggests that the selection of the ideal strategy heavily depends on the size of the training dataset at hand. One of the reasons for their robustness with respect to dataset size is likely that input-level object aggregation methods generally introduce stronger inductive biases than output-level object aggregation, e.g., by using hand-crafted features per node and imposing the graph structure. This is an important consideration for land cover mapping in regions of the world where data from principled land cover surveys is less plentiful or unavailable.

\subsection{Other drivers of performance}
\label{sec:drivers_of_performance}

Beyond the described differences between input-level and output-level object aggregation, our results demonstrate that several other factors play a crucial role with respect to the achievable accuracy of object-based deep learning methods.

First, we find that the object definition is a crucial element of the object-based deep learning pipeline: As can be seen in the ablation study in Section \ref{sec:ablation_object_definition}, the FH algorithm produces segments which are better suited for object-based land cover classification than SLIC, leading to higher classification accuracies. This can be explained by the fact that the specification of a MMU in FH is a straightforward extension whereas it requires surpression of smaller segments in SLIC, as described in Section \ref{sec:object_definition}. This leads to more objects which violate the assumption that a single land cover class describes all pixels within the segment.

For input-level object aggregation, we find that the overall design of the deep learning model strongly affects the achievable accuracy. This is demonstrated by the fact that GUNet performs significantly better than BaseGNN, as shown in Section \ref{sec:effect_of_mmu} and that there is a large difference between the different graph convolution operators, as shown in Section \ref{sec:graph_convolution_operators_in_input-level_object_aggregation}. Both graph neural networks, however, outperform BaseMLP which shows that the spatial context provided by the graph structure is strongly beneficial in input-level object aggregation. Moreover, in Section \ref{sec:features_in_input-level_object_aggregation}, we observe a significant improvement when variability features were introduced into the classification. On the other hand, geometric features, as commonly used in object-based image analysis literature \citep{blaschke2010object,blaschke2014geographic,ma2017evaluation,cosma2023geometric} did not lead to significant improvements. Possible reasons for this may be that the shape features are redundant with the spectral features or that the coupling between geometric features and land cover classes is weak due to their diversity across geographies.

In the context of output-level object aggregation, an important finding is that choosing a good architecture for a model is more important than scaling the parameter count in the considered setting with sparse labels. For instance, the results in Section \ref{sec:segmentation_backbones_in_output-level_object_aggregation} show that there is a comparably large difference between the performances of DeepLabV3 and UNet, while the differences when using ResNet-18 or ResNet-50 is barely significant in any of the two. Indeed, none of the tested semantic segmentation methods reliably delivered improved performances when combined with increasingly powerful backbones. Again, this indicates that the inductive biases provided by a model are an important consideration under data sparsity and data availability constraints. Additionally, the results in Section \ref{sec:rescaling_in_output-level_object_aggregation} suggest that rescaling is a necessary step in output-level object aggregation. The likely reason for this phenomenon is that the backbones used by these models conduct spatial aggregation already in the first layers of the networks such that the high spatial granularity cannot be recovered later. The sparse label setting, which further introduces instability likely exacerbates this issue.

Finally, we tested the capabilities of models trained on features from a pre-trained model, rather than the raw intensity features. We observe that this leads to significant improvements in terms of accuracy when using small models. The improvement was especially noticeable for smaller training datasets. Moreover, the incorporation of pre-trained features lead to a equalization of the tested models in terms of their performance. This suggests that the classification in this setting is not constrained by the classifiers' capacity, but rather the expressiveness of the learned features.

\subsection{Implications for Earth observation practice}
\label{sec:implications_for_earth_observation_practice}

The presented results and conclusion therefrom highlight some more general implications in the context of practical Earth observation.

Most importantly, we show that several configurations of our framework outperform established third-party products, namely ESA-WC and ESRI-LC, in terms of accuracy in Section \ref{sec:comparison_to_third-party_land_cover_products}. In this context, our results in terms of accuracy broadly agree with those determined in other studies who compare established products with LUCAS data \citep{gao2020consistency,venter2022global,xu2024comparative}. The small differences can be explained by the fact that we only consider a subset of all LUCAS data, i.e., the test set, for our evaluation. Our results therefore demonstrate that land cover classification based on sparse in-situ labels using object-based deep learning approaches is a scalable alternative to the expensive practice of manual annotation without having to sacrifice accuracy.

Secondly, we point out that our combined results suggest that model inductive biases are a more important consideration than model complexity in the context of sparse labels. This leads us to the conclusion that the setting is constrained by data more than by parametrization. In general, smaller dataset sizes should be dealt with by heavier inductive biases, e.g. input-level object aggregation or pre-training on existing land cover maps, whereas the more flexible semantic segmentation models have the advantage in settings where training data is more abundant.

\subsection{Limitations and future work}
\label{sec:limitations_and_future_work}

While the presented LC-SLab framework has proven effective for analyzing the tradeoff between accuracy and fragmentation in object-based, there are still clear limitations which should be addressed in future work.

First, while it was demonstrated that the resulting land cover maps are competitive with established products, the use of annually aggregated satellite imagery still represents an obvious limitation of the presented approach. By compressing the image time series into a single composite, we effectively discard intra‑annual dynamics that are crucial for distinguishing phenologically similar classes, such as \textit{cropland} vs. \textit{grassland}. Thus, we plan to extend the LC-SLab framework to multi-temporal object-based deep learning approaches and use an enhanced dataset including monthly or bi-monthly satellite image time series, similar to \citet{sharma2024sen4map}. This may also open up the possibility to distinguish between finer-grained land cover classes, as present in the LUCAS class nomenclature.

The compression of temporal information also partly explains the great variation across class-wise F1 scores, as described in Section \ref{sec:effect_of_mmu}, as it leads to a lack of separability of certain classes in feature space. Other likely reasons for this discrepancy are the underrepresentation of certain classes in the training dataset as showcased in Figure \ref{fig:data}. However, since our results are in agreement with previous work on LUCAS-based classification \citep{sharma2024sen4map}, we view this as a general problem rather than one specific to our framework. Nonetheless, the variation of achievable accuracy across land cover classes also implies that accuracy varies based on the general land cover scenario and geography. In particular, the model may underperform in regions and where hard to classify classes like \textit{shrubland} and \textit{bare land} are especially present, such as mountainous or coastal landscapes. When large MMUs are specified, accuracy also suffers for the \textit{artificial land} class. This is explained by the high fragmentation of urban landscapes where smaller MMUs would generally be preferrable.

In addition, the focus of this study is on annual land cover classification, as it has been the most common practical approach in other recent investigations \citep{karra2021global,zanaga2021esa,brown2022dynamic,mirmazloumi2022elulc,sharma2024sen4map}. As such, natural phenomena which occur on smaller temporal timescales such as crop rotation, temporary flooding, or deforestation cannot be represented in the data. As a consequence, the current framework is better suited for temporally stable land cover conditions than for highly dynamic or management‑driven landscapes. Some misclassifications may be structurally unavoidable in the present setup.

Despite these limitations, LC‑SLab is designed to be extensible. On the input side, the framework can incorporate additional data sources that help address current data gaps, such as SAR or lower resolution optical data. Beyond land cover mapping, the framework's sparse-label object-based design is also readily transferable to other Earth observation tasks where in-situ annotations are costly or scarce, such as crop type mapping, wetland delineation, or biodiversity monitoring from field observations.

On the modeling side, our formulation of sparse supervision and object‑based processing is compatible with a wide range of backbone architectures, This also includes foundation models, the integration of which we view as an promising research direction. Especially, recently proposed remote sensing-specific foundation models such as Prithvi \citep{jakubik2023foundation}, TerraMind \citep{jakubik2025terramind}, or AlphaEarth \citep{brown2025alphaearth} which provide highly complex representations of multi-modal and multi-temporal geospatial data could be used in place of the pre-trained feature extractor, as suggested in this work. While these models have achieved great performances on several benchmark datasets, their practical capabilities for large-scale land cover mapping are not yet extensively explored \citep{ma2026harvesting}. In particular, foundation models should be compared to models trained from scratch or using more targeted pre-training approaches, as suggested in this study. Additionally, the Segment Anything Model (SAM) \citep{kirillov2023segment}, a foundation model for image segmentation, could be used for object definition. This would eliminate the reliance of our approaches on traditional oversegmentation approaches.

\section{Conclusion}
\label{sec:conclusion}

This study introduces LC-SLab, an object-based deep learning framework targeted and large-scale land cover classification from sparse in-situ label data. It is demonstrated that object-based deep learning methods can effectively mitigate the spatial fragmentation associated with training land cover classification on sparse labels, without sacrificing classification accuracy. 

By systematically evaluating the effect of multiple integration strategies, dataset sizes, and other design choices within a unified framework, we show that the choice of model components and the level at which object information is introduced both play a critical role under limited label availability. Most notably, we differentiate between input-level and output-level object aggregation. In input-level object aggregation, the image is represented as a graph and classification works best for small dataset sizes, whereas output-level object aggregation combines established semantic segmentation models with an object-based preprocessing step which was most successful for larger datasets. Additionally, it was shown that pretraining leads to greater label efficiency and accuracy, particularly for simple classifiers. Finally, our results beat established land cover products in terms of accuracy showcasing that despite the additional effort, sparse labels are a viable alternative when training neural networks for Earth observation.

While these findings are grounded in a dataset of  Sentinel-2 imagery and LUCAS survey data, our insights provide actionable guidance for designing label-efficient pipelines across diverse geospatial scenarios. As such, LC-SLab provides not only a practical tool for land cover classification, but also a structured basis for analyzing tradeoffs between accuracy and spatial coherence in other settings. The framework's modular design positions it as a foundation for future extensions, including multi-temporal analysis and the incorporation of geospatial foundation models, thus serving as a starting point for continued methodological research in the context of sparsely labeled Earth observation problems.

\section*{Acknowledgements}

Funded by the Deutsche Forschungsgemeinschaft (DFG, German Research Foundation) – SFB 1502/1-2022 – Projektnummer: 450058266. 

\section*{Data availability} 

All code and data are publicly available at \url{https://github.com/johannes-leonhardt/lc-slab}.

\printcredits

\bibliographystyle{cas-model2-names}

\bibliography{references}

\newpage

% Appendix
{\LARGE \noindent Supplementary material for ``LC-SLab -- An Object-based Deep Learning Framework for Large-scale Land Cover Classification from Satellite Imagery and Sparse In-situ Labels"\par}
\thispagestyle{appfirstpage}

\appendix

\section{Supplementary material for Section 3.1}
\label{app:supplementary_material_for_section_3_1}
\counterwithin{table}{section}

In this section, we provide the detailed results for all accuracy and fragmentation metrics from the experiments regarding the effect of the MMU in Section 3.1.

\begin{table}[H]
\caption{Metrics for different models and different specified MMUs. Accompanies Figure 4.}
\begin{subtable}{\textwidth}
\centering
\scriptsize
\subcaption{MMU=$1px$}
\begin{tabular}{lccccccc}
\toprule
 \multirow{2}{*}{Method} & \multicolumn{2}{c}{Overall accuracy} & \multicolumn{2}{c}{F1 score} & \multirow{2}{*}{Patch density} & \multirow{2}{*}{Edge density} & \multirow{2}{*}{Entropy} \\
 & t=0 & t=1 & t=0 & t=1 &  &  &  \\ \midrule
 BaseMLP & 60.1 $\pm$ 0.0 & 65.4 $\pm$ 0.0 & 43.1 $\pm$ 0.1 & 47.7 $\pm$ 0.1 & 126.7 $\pm$ 0.5 & 907 $\pm$ 2 & 1.166 $\pm$ 0.001 \\
 BaseGNN & 67.5 $\pm$ 0.0 & 71.8 $\pm$ 0.1 & 52.1 $\pm$ 0.1 & 56.4 $\pm$ 0.1 & 74.4 $\pm$ 4.5 & 665 $\pm$ 20 & 1.124 $\pm$ 0.006 \\
 GUNet & 71.0 $\pm$ 0.1 & 73.5 $\pm$ 0.1 & 56.9 $\pm$ 0.1 & 59.9 $\pm$ 0.1 & 32.3 $\pm$ 0.5 & 400 $\pm$ 1 & 0.997 $\pm$ 0.003 \\ \midrule
 BaseCNN & 67.0 $\pm$ 0.3 & 70.7 $\pm$ 0.1 & 50.9 $\pm$ 0.1 & 54.8 $\pm$ 0.4 & 64.5 $\pm$ 9.9 & 629 $\pm$ 39 & 1.123 $\pm$ 0.011 \\
 UNet & 70.5 $\pm$ 0.1 & 72.9 $\pm$ 0.2 & 54.8 $\pm$ 0.2 & 57.9 $\pm$ 0.3 & 34.1 $\pm$ 2.7 & 411 $\pm$ 17 & 0.919 $\pm$ 0.011 \\
 UNet++ & 69.8 $\pm$ 0.1 & 73.1 $\pm$ 0.2 & 55.6 $\pm$ 0.1 & 59.7 $\pm$ 0.3 & 48.2 $\pm$ 4.0 & 523 $\pm$ 15 & 1.039 $\pm$ 0.003 \\
 DeepLabV3 & \textbf{71.3 $\pm$ 0.4} & \textbf{73.8 $\pm$ 0.4} & \textbf{57.4 $\pm$ 0.2} & \textbf{60.5 $\pm$ 0.3} & \textbf{29.2 $\pm$ 0.7} & \textbf{396 $\pm$ 4} & \textbf{0.917 $\pm$ 0.003} \\
 Segformer & 69.7 $\pm$ 0.0 & 72.3 $\pm$ 0.0 & 55.7 $\pm$ 0.3 & 58.8 $\pm$ 0.4 & 33.4 $\pm$ 0.3 & 419 $\pm$ 2 & 0.920 $\pm$ 0.004 \\ \bottomrule
\end{tabular}
\end{subtable}
\end{table}
% \end{minipage}

\begin{table}[H]
\ContinuedFloat
\begin{subtable}{\textwidth}
\centering
\scriptsize
\subcaption{MMU=$5px$}
\begin{tabular}{lccccccc}
\toprule
 \multirow{2}{*}{Method} & \multicolumn{2}{c}{Overall accuracy} & \multicolumn{2}{c}{F1 score} & \multirow{2}{*}{Patch density} & \multirow{2}{*}{Edge density} & \multirow{2}{*}{Entropy} \\
 & t=0 & t=1 & t=0 & t=1 &  &  &  \\ \midrule
 BaseMLP & 63.1 $\pm$ 0.3 & 67.1 $\pm$ 0.3 & 46.8 $\pm$ 0.5 & 50.5 $\pm$ 0.4 & 43.9 $\pm$ 1.9 & 599 $\pm$ 15 & 1.150 $\pm$ 0.018 \\
 BaseGNN & 68.6 $\pm$ 0.1 & 71.4 $\pm$ 0.1 & 54.6 $\pm$ 0.1 & 57.6 $\pm$ 0.2 & 25.4 $\pm$ 2.2 & 403 $\pm$ 21 & 1.010 $\pm$ 0.020 \\
 GUNet & 70.7 $\pm$ 0.1 & \textbf{73.1 $\pm$ 0.1} & 56.9 $\pm$ 0.3 & \textbf{59.6 $\pm$ 0.4} & 18.6 $\pm$ 0.1 & 324 $\pm$ 2 & 0.874 $\pm$ 0.006 \\ \midrule
 BaseCNN & 67.0 $\pm$ 0.2 & 69.8 $\pm$ 0.2 & 50.5 $\pm$ 0.2 & 53.4 $\pm$ 0.3 & 25.9 $\pm$ 0.4 & 420 $\pm$ 1 & 1.028 $\pm$ 0.003 \\
 UNet & 69.9 $\pm$ 0.1 & 72.2 $\pm$ 0.2 & 54.1 $\pm$ 0.2 & 56.7 $\pm$ 0.3 & 23.1 $\pm$ 1.8 & 340 $\pm$ 12 & 0.881 $\pm$ 0.015 \\
 UNet++ & 69.6 $\pm$ 0.3 & 72.4 $\pm$ 0.3 & 54.7 $\pm$ 0.5 & 58.0 $\pm$ 0.5 & 29.0 $\pm$ 1.0 & 398 $\pm$ 1 & 0.975 $\pm$ 0.004 \\
 DeepLabV3 & \textbf{70.9 $\pm$ 0.4} & \textbf{73.1 $\pm$ 0.4} & \textbf{57.0 $\pm$ 0.3} & 59.5 $\pm$ 0.4 & \textbf{18.2 $\pm$ 0.2} & \textbf{319 $\pm$ 2} & \textbf{0.861 $\pm$ 0.002} \\
 Segformer & 69.3 $\pm$ 0.0 & 71.6 $\pm$ 0.0 & 55.3 $\pm$ 0.3 & 57.9 $\pm$ 0.3 & 20.3 $\pm$ 0.1 & 336 $\pm$ 1 & \textbf{0.861 $\pm$ 0.002} \\ \bottomrule
\end{tabular}
\end{subtable}
\end{table}

\begin{table}[H]
\ContinuedFloat
\begin{subtable}{\textwidth}
\centering
\scriptsize
\subcaption{MMU=$10px$}
\begin{tabular}{lccccccc}
\toprule
 \multirow{2}{*}{Method} & \multicolumn{2}{c}{Overall accuracy} & \multicolumn{2}{c}{F1 score} & \multirow{2}{*}{Patch density} & \multirow{2}{*}{Edge density} & \multirow{2}{*}{Entropy} \\
 & t=0 & t=1 & t=0 & t=1 &  &  &  \\ \midrule
 BaseMLP & 63.9 $\pm$ 0.0 & 67.2 $\pm$ 0.0 & 47.0 $\pm$ 0.0 & 50.0 $\pm$ 0.1 & 24.5 $\pm$ 0.0 & 453 $\pm$ 0 & 1.086 $\pm$ 0.001 \\
 BaseGNN & 68.2 $\pm$ 0.2 & 70.5 $\pm$ 0.2 & 54.0 $\pm$ 0.1 & 56.6 $\pm$ 0.1 & 14.9 $\pm$ 0.2 & 308 $\pm$ 3 & 0.930 $\pm$ 0.003 \\
 GUNet & 69.4 $\pm$ 0.2 & 71.4 $\pm$ 0.2 & 56.0 $\pm$ 0.3 & 58.3 $\pm$ 0.3 & 13.0 $\pm$ 0.2 & 277 $\pm$ 3 & 0.852 $\pm$ 0.005 \\ \midrule
 BaseCNN & 66.9 $\pm$ 0.2 & 69.2 $\pm$ 0.2 & 50.0 $\pm$ 0.3 & 52.4 $\pm$ 0.3 & 16.9 $\pm$ 0.1 & 342 $\pm$ 0 & 0.977 $\pm$ 0.003 \\
 UNet & 69.4 $\pm$ 0.1 & 71.5 $\pm$ 0.2 & 53.3 $\pm$ 0.3 & 55.8 $\pm$ 0.3 & 16.7 $\pm$ 1.5 & 291 $\pm$ 10 & 0.859 $\pm$ 0.018 \\
 UNet++ &  69.2 $\pm$ 0.3 & 71.7 $\pm$ 0.3 & 53.6 $\pm$ 0.9 & 56.7 $\pm$ 0.7 & 20.4 $\pm$ 0.8 & 333 $\pm$ 3 & 0.946 $\pm$ 0.006 \\
 DeepLabV3 & \textbf{70.5 $\pm$ 0.4} & \textbf{72.5 $\pm$ 0.4} & \textbf{56.4 $\pm$ 0.2} & \textbf{58.7 $\pm$ 0.3} & \textbf{12.6 $\pm$ 0.1} & \textbf{269 $\pm$ 1} & 0.828 $\pm$ 0.001 \\
 Segformer & 69.0 $\pm$ 0.0 & 71.1 $\pm$ 0.1 & 54.9 $\pm$ 0.4 & 57.2 $\pm$ 0.4 & 13.9 $\pm$ 0.1 & 280 $\pm$ 0 & \textbf{0.825 $\pm$ 0.001} \\ \bottomrule
\end{tabular}
\end{subtable}
\end{table}

\begin{table}[H]
\ContinuedFloat
\begin{subtable}{\textwidth}
\centering
\scriptsize
\subcaption{MMU=$20px$}
\begin{tabular}{lccccccc}
\toprule
 \multirow{2}{*}{Method} & \multicolumn{2}{c}{Overall accuracy} & \multicolumn{2}{c}{F1 score} & \multirow{2}{*}{Patch density} & \multirow{2}{*}{Edge density} & \multirow{2}{*}{Entropy} \\
 & t=0 & t=1 & t=0 & t=1 &  &  &  \\ \midrule
 BaseMLP & 64.6 $\pm$ 0.0 & 67.2 $\pm$ 0.0 & 48.5 $\pm$ 0.2 & 51.1 $\pm$ 0.1 & 15.7 $\pm$ 0.0 & 355 $\pm$ 0 & 1.056 $\pm$ 0.001 \\
 BaseGNN & 67.6 $\pm$ 0.2 & 69.5 $\pm$ 0.2 & 53.2 $\pm$ 0.6 & 55.2 $\pm$ 0.6 & 9.9 $\pm$ 0.4 & 244 $\pm$ 8 & 0.867 $\pm$ 0.015 \\
 GUNet & 68.4 $\pm$ 0.1 & 70.1 $\pm$ 0.1 & 54.7 $\pm$ 0.5 & 56.6 $\pm$ 0.6 & \textbf{8.9 $\pm$ 0.0} & 225 $\pm$ 1 & 0.799 $\pm$ 0.005 \\ \midrule
 BaseCNN & 66.4 $\pm$ 0.2 & 68.4 $\pm$ 0.2 & 49.0 $\pm$ 0.3 & 51.0 $\pm$ 0.3 & 11.2 $\pm$ 0.0 & 273 $\pm$ 0 & 0.913 $\pm$ 0.003 \\
 UNet & 69.0 $\pm$ 0.2 & 70.9 $\pm$ 0.2 & 52.8 $\pm$ 0.3 & 54.8 $\pm$ 0.3 & 11.3 $\pm$ 0.9 & 239 $\pm$ 8 & 0.819 $\pm$ 0.017 \\
 UNet++ & 68.7 $\pm$ 0.4 & 70.8 $\pm$ 0.4 & 52.7 $\pm$ 0.9 & 55.3 $\pm$ 0.8 & 13.3 $\pm$ 0.5 & 267 $\pm$ 2 & 0.894 $\pm$ 0.006 \\
 DeepLabV3 & \textbf{70.0 $\pm$ 0.4} & \textbf{71.7 $\pm$ 0.4} & \textbf{55.7 $\pm$ 0.3} & \textbf{57.5 $\pm$ 0.3} & \textbf{8.9 $\pm$ 0.0} & \textbf{222 $\pm$ 1} & 0.785 $\pm$ 0.002 \\
 Segformer & 68.7 $\pm$ 0.1 & 70.4 $\pm$ 0.1 & 54.3 $\pm$ 0.4 & 56.2 $\pm$ 0.4 & 9.4 $\pm$ 0.1 & 226 $\pm$ 0 & \textbf{0.778 $\pm$ 0.001} \\ \bottomrule
\end{tabular}
\end{subtable}
\end{table}

\begin{table}[H]
\ContinuedFloat
\begin{subtable}{\textwidth}
\centering
\scriptsize
\subcaption{MMU=$40px$}
\begin{tabular}{lccccccc}
\toprule
 \multirow{2}{*}{Method} & \multicolumn{2}{c}{Overall accuracy} & \multicolumn{2}{c}{F1 score} & \multirow{2}{*}{Patch density} & \multirow{2}{*}{Edge density} & \multirow{2}{*}{Entropy} \\
 & t=0 & t=1 & t=0 & t=1 &  &  &  \\ \midrule
 BaseMLP & 64.5 $\pm$ 0.1 & 66.4 $\pm$ 0.1 & 48.3 $\pm$ 0.2 & 50.3 $\pm$ 0.2 & 9.6 $\pm$ 0.1 & 260 $\pm$ 2 & 0.970 $\pm$ 0.005 \\
 BaseGNN & 66.5 $\pm$ 0.3 & 68.0 $\pm$ 0.3 & 51.5 $\pm$ 0.9 & 53.2 $\pm$ 1.0 & 6.9 $\pm$ 0.4 & 194 $\pm$ 9 & 0.802 $\pm$ 0.030 \\
 GUNet & 67.0 $\pm$ 0.1 & 68.4 $\pm$ 0.0 & 52.5 $\pm$ 0.1 & 54.0 $\pm$ 0.1 & 6.4 $\pm$ 0.1 & 181 $\pm$ 4 & 0.745 $\pm$ 0.011 \\ \midrule
 BaseCNN & 66.0 $\pm$ 0.2 & 67.6 $\pm$ 0.2 & 48.1 $\pm$ 0.4 & 49.6 $\pm$ 0.4 & 7.5 $\pm$ 0.0 & 210 $\pm$ 0 & 0.835 $\pm$ 0.002 \\
 UNet & 68.5 $\pm$ 0.1 & 70.0 $\pm$ 0.2 & 52.0 $\pm$ 0.3 & 53.5 $\pm$ 0.4 & 7.3 $\pm$ 0.4 & 187 $\pm$ 5 & 0.757 $\pm$ 0.014 \\
 UNet++ & 68.2 $\pm$ 0.4 & 69.8 $\pm$ 0.3 & 52.1 $\pm$ 0.7 & 54.0 $\pm$ 0.6 & 8.2 $\pm$ 0.2 & 205 $\pm$ 1 & 0.815 $\pm$ 0.005 \\
 DeepLabV3 & \textbf{69.3 $\pm$ 0.4} & \textbf{70.7 $\pm$ 0.4} & \textbf{54.5 $\pm$ 0.3} & \textbf{56.1 $\pm$ 0.3} & \textbf{6.3 $\pm$ 0.0} & \textbf{177 $\pm$ 0} & 0.728 $\pm$ 0.002 \\
 Segformer & 68.1 $\pm$ 0.0 & 69.5 $\pm$ 0.0 & 53.4 $\pm$ 0.4 & 54.9 $\pm$ 0.4 & 6.4 $\pm$ 0.1 & \textbf{177 $\pm$ 0} & \textbf{0.718 $\pm$ 0.001} \\ \bottomrule
\end{tabular}
\end{subtable}
\end{table}

% Class-wise metrics

\begin{table}[H]
\caption{Class-wise metrics for DeepLabV3 under different MMUs. Accompanies Figure 5.}
\begin{subtable}{\textwidth}
\centering
\scriptsize
\subcaption{MMU=$1px$}
\begin{tabular}{lcccc}
\toprule
 \multirow{2}{*}{Land cover class} & \multicolumn{2}{c}{F1 score} & \multirow{2}{*}{Patch density} & \multirow{2}{*}{Edge density} \\
 & t=0 & t=1 &  &  \\ \midrule
 Artificial Land & 66.8 $\pm$ 0.2 & 71.5 $\pm$ 0.0 & 102.1 $\pm$ 2.1 & 2131 $\pm$ 39 \\
 Cropland & 71.7 $\pm$ 0.3 & 72.7 $\pm$ 0.4 & \textbf{12.8 $\pm$ 0.8} & \textbf{474 $\pm$ 12} \\
 Woodland & \textbf{85.2 $\pm$ 0.2} & \textbf{87.3 $\pm$ 0.2} & 16.2 $\pm$ 0.4 & 513 $\pm$ 5 \\
 Shrubland & 21.7 $\pm$ 0.1 & 26.5 $\pm$ 0.4 & 105.0 $\pm$ 1.7 & 2022 $\pm$ 31 \\
 Grassland & 64.5 $\pm$ 0.3 & 67.8 $\pm$ 0.4 & 38.1 $\pm$ 2.2 & 1060 $\pm$ 24 \\
 Bare land & 16.3 $\pm$ 0.5 & 18.3 $\pm$ 0.2 & 112.1 $\pm$ 1.7 & 1977 $\pm$ 38 \\
 Water & 68.5 $\pm$ 0.2 & 71.8 $\pm$ 0.6 & 26.3 $\pm$ 2.7 & 698 $\pm$ 19 \\
 Wetlands & 64.2 $\pm$ 1.0 & 68.0 $\pm$ 1.4 & 31.3 $\pm$ 4.8 & 894 $\pm$ 80 \\ \bottomrule
\end{tabular}
\end{subtable}
\end{table}

\begin{table}[H]
 
\ContinuedFloat
\begin{subtable}{\textwidth}
\centering
\scriptsize
\subcaption{MMU=$5px$}
\begin{tabular}{lcccc}
\toprule
 \multirow{2}{*}{Land cover class} & \multicolumn{2}{c}{F1 score} & \multirow{2}{*}{Patch density} & \multirow{2}{*}{Edge density} \\
 & t=0 & t=1 &  &  \\ \midrule
 Artificial Land & 65.4 $\pm$ 0.1 & 68.9 $\pm$ 0.1 & 56.6 $\pm$ 0.1 & 1624 $\pm$ 20 \\
 Cropland & 71.5 $\pm$ 0.3 & 72.5 $\pm$ 0.4 & \textbf{10.2 $\pm$ 0.4} & 424 $\pm$ 13 \\
 Woodland & \textbf{84.8 $\pm$ 0.2} & \textbf{86.8 $\pm$ 0.2} & 10.4 $\pm$ 0.2 & \textbf{416 $\pm$ 6} \\
 Shrubland & 21.7 $\pm$ 0.4 & 25.2 $\pm$ 0.8 & 54.3 $\pm$ 1.8 & 1513 $\pm$ 28 \\
 Grassland & 64.1 $\pm$ 0.3 & 66.8 $\pm$ 0.4 & 24.3 $\pm$ 0.6 & 847 $\pm$ 14 \\
 Bare land & 16.2 $\pm$ 0.0 & 17.4 $\pm$ 0.4 & 55.4 $\pm$ 2.5 & 1432 $\pm$ 24 \\
 Water & 68.4 $\pm$ 0.3 & 71.2 $\pm$ 0.4 & 17.3 $\pm$ 0.9 & 641 $\pm$ 7 \\
 Wetlands & 63.9 $\pm$ 1.2 & 67.2 $\pm$ 1.4 & 24.5 $\pm$ 3.4 & 794 $\pm$ 67 \\ \bottomrule
\end{tabular}
\end{subtable}
\end{table}

\begin{table}[H]
 
\ContinuedFloat
\begin{subtable}{\textwidth}
\centering
\scriptsize
\subcaption{MMU=$10px$}
\begin{tabular}{lcccc}
\toprule
 \multirow{2}{*}{Land cover class} & \multicolumn{2}{c}{F1 score} & \multirow{2}{*}{Patch density} & \multirow{2}{*}{Edge density} \\
 & t=0 & t=1 &  &  \\ \midrule
 Artificial Land & 63.5 $\pm$ 0.1 & 66.5 $\pm$ 0.1 & 38.3 $\pm$ 0.3 & 1333 $\pm$ 16 \\
 Cropland & 71.3 $\pm$ 0.3 & 72.2 $\pm$ 0.3 & 7.9 $\pm$ 0.2 & 378 $\pm$ 11 \\
 Woodland & \textbf{84.4 $\pm$ 0.3} & \textbf{86.3 $\pm$ 0.2} & \textbf{7.4 $\pm$ 0.1} & \textbf{357 $\pm$ 5} \\
 Shrubland & 21.3 $\pm$ 0.3 & 24.1 $\pm$ 0.6 & 34.0 $\pm$ 1.8 & 1221 $\pm$ 27 \\
 Grassland & 63.7 $\pm$ 0.3 & 66.1 $\pm$ 0.3 & 17.0 $\pm$ 0.4 & 709 $\pm$ 14 \\
 Bare land & 16.2 $\pm$ 0.1 & 17.1 $\pm$ 0.3 & 35.3 $\pm$ 2.1 & 1163 $\pm$ 21 \\
 Water & 68.1 $\pm$ 0.5 & 70.7 $\pm$ 0.5 & 13.5 $\pm$ 0.4 & 597 $\pm$ 4 \\
 Wetlands & 63.1 $\pm$ 1.2 & 66.3 $\pm$ 1.5 & 17.2 $\pm$ 2.4 & 677 $\pm$ 57 \\ \bottomrule
\end{tabular}
\end{subtable}
\end{table}

\begin{table}[H]
 
\ContinuedFloat
\begin{subtable}{\textwidth}
\centering
\scriptsize
\subcaption{MMU=$20px$}
\begin{tabular}{lcccc}
\toprule
 \multirow{2}{*}{Land cover class} & \multicolumn{2}{c}{F1 score} & \multirow{2}{*}{Patch density} & \multirow{2}{*}{Edge density} \\
 & t=0 & t=1 &  &  \\ \midrule
 Artificial Land & 60.5 $\pm$ 0.1 & 63.0 $\pm$ 0.2 & 24.2 $\pm$ 0.4 & 1023 $\pm$ 14 \\
 Cropland & 70.9 $\pm$ 0.3 & 71.8 $\pm$ 0.3 & 6.1 $\pm$ 0.1 & 329 $\pm$ 9 \\
 Woodland & \textbf{84.1 $\pm$ 0.2} & \textbf{85.7 $\pm$ 0.2} & \textbf{5.5 $\pm$ 0.1} & \textbf{303 $\pm$ 5} \\
 Shrubland & 21.1 $\pm$ 0.8 & 23.2 $\pm$ 0.9 & 21.3 $\pm$ 1.5 & 951 $\pm$ 25 \\
 Grassland & 63.2 $\pm$ 0.3 & 65.3 $\pm$ 0.3 & 12.0 $\pm$ 0.3 & 581 $\pm$ 13 \\
 Bare land & 15.9 $\pm$ 0.4 & 16.4 $\pm$ 0.3 & 22.4 $\pm$ 1.3 & 921 $\pm$ 19 \\
 Water & 67.6 $\pm$ 0.2 & 70.3 $\pm$ 0.2 & 10.7 $\pm$ 0.1 & 548 $\pm$ 3 \\
 Wetlands & 62.1 $\pm$ 1.1 & 64.7 $\pm$ 1.3 & 12.0 $\pm$ 1.7 & 561 $\pm$ 49 \\ \bottomrule
\end{tabular}
\end{subtable}
\end{table}

\begin{table}[H]
 
\ContinuedFloat
\begin{subtable}{\textwidth}
\centering
\scriptsize
\subcaption{MMU=$40px$}
\begin{tabular}{lcccc}
\toprule
 \multirow{2}{*}{Land cover class} & \multicolumn{2}{c}{F1 score} & \multirow{2}{*}{Patch density} & \multirow{2}{*}{Edge density} \\
 & t=0 & t=1 &  &  \\ \midrule
 Artificial Land & 57.6 $\pm$ 0.2 & 59.3 $\pm$ 0.3 & 14.1 $\pm$ 0.3 & 725 $\pm$ 11 \\
 Cropland & 70.5 $\pm$ 0.3 & 71.1 $\pm$ 0.3 & 4.8 $\pm$ 0.1 & 280 $\pm$ 7 \\
 Woodland & \textbf{83.6 $\pm$ 0.3} & \textbf{85.0 $\pm$ 0.2} & \textbf{4.2 $\pm$ 0.1} & \textbf{250 $\pm$ 5} \\
 Shrubland & 20.6 $\pm$ 0.6 & 22.4 $\pm$ 0.6 & 13.3 $\pm$ 1.0 & 710 $\pm$ 24 \\
 Grassland & 62.6 $\pm$ 0.3 & 64.3 $\pm$ 0.3 & 8.4 $\pm$ 0.2 & 462 $\pm$ 11 \\
 Bare land & 15.2 $\pm$ 0.1 & 15.7 $\pm$ 0.1 & 14.1 $\pm$ 0.8 & 700 $\pm$ 24 \\
 Water & 65.7 $\pm$ 0.0 & 68.1 $\pm$ 0.2 & 8.4 $\pm$ 0.0 & 494 $\pm$ 3 \\
 Wetlands & 60.6 $\pm$ 1.2 & 62.9 $\pm$ 1.2 & 8.2 $\pm$ 1.0 & 448 $\pm$ 40 \\ \bottomrule
\end{tabular}
\end{subtable}
\end{table}

\section{Supplementary material for Section 3.2}

In this section, we provide the detailed results for all accuracy and fragmentation metrics from the experiments regarding the effect of the dataset size in Section 3.2.

\begin{table}[H]
\caption{Metrics for the $1/2$ dataset, accompanies Figure 6(a)}
\begin{subtable}{\textwidth}
\centering
\scriptsize
\subcaption{MMU=$1px$}
\begin{tabular}{lccccccc}
\toprule
 \multirow{2}{*}{Method} & \multicolumn{2}{c}{Overall accuracy} & \multicolumn{2}{c}{F1 score} & \multirow{2}{*}{Patch density} & \multirow{2}{*}{Edge density} & \multirow{2}{*}{Entropy} \\
 & t=0 & t=1 & t=0 & t=1 &  &  &  \\ \midrule
 BaseMLP & 59.8 $\pm$ 0.1 & 65.0 $\pm$ 0.1 & 42.8 $\pm$ 0.1 & 47.4 $\pm$ 0.1 & 118.9 $\pm$ 1.7 & 882 $\pm$ 7 & 1.157 $\pm$ 0.005 \\
 BaseGNN & 66.9 $\pm$ 0.0 & 70.7 $\pm$ 0.1 & 50.4 $\pm$ 0.1 & 54.3 $\pm$ 0.2 & 63.5 $\pm$ 2.2 & 611 $\pm$ 12 & 1.107 $\pm$ 0.006 \\
 GUNet & \textbf{70.1 $\pm$ 0.0} & \textbf{72.7 $\pm$ 0.1} & \textbf{54.9 $\pm$ 0.4} & 58.1 $\pm$ 0.5 & 35.4 $\pm$ 2.0 & 419 $\pm$ 14 & 1.014 $\pm$ 0.014 \\ \midrule
 BaseCNN & 66.1 $\pm$ 0.4 & 70.6 $\pm$ 0.6 & 50.2 $\pm$ 0.4 & 55.0 $\pm$ 1.1 & 104.3 $\pm$ 29.7 & 782 $\pm$ 133 & 1.167 $\pm$ 0.055 \\
 UNet & 69.1 $\pm$ 0.1 & 72.0 $\pm$ 0.4 & 52.4 $\pm$ 1.2 & 55.6 $\pm$ 1.8 & 40.2 $\pm$ 4.7 & 457 $\pm$ 31 & 0.967 $\pm$ 0.024 \\
 UNet++ & 68.4 $\pm$ 0.4 & 71.3 $\pm$ 0.3 & 50.2 $\pm$ 0.5 & 53.3 $\pm$ 0.5 & 39.7 $\pm$ 0.7 & 468 $\pm$ 12 & 0.969 $\pm$ 0.009 \\
 DeepLabV3 & 69.7 $\pm$ 0.4 & 72.3 $\pm$ 0.4 & 55.1 $\pm$ 0.2 & \textbf{58.4 $\pm$ 0.0} & \textbf{30.3 $\pm$ 3.6} & \textbf{402 $\pm$ 27} & 0.922 $\pm$ 0.030 \\
 Segformer & 68.4 $\pm$ 0.3 & 71.1 $\pm$ 0.2 & 54.3 $\pm$ 0.2 & 57.4 $\pm$ 0.2 & 33.3 $\pm$ 1.1 & 420 $\pm$ 15 & \textbf{0.905 $\pm$ 0.028} \\ \bottomrule
\end{tabular}
\end{subtable}
\end{table}

\begin{table}[H]
 
\ContinuedFloat
\begin{subtable}{\textwidth}
\centering
\scriptsize
\subcaption{MMU=$5px$}
\begin{tabular}{lccccccc}
\toprule
 \multirow{2}{*}{Method} & \multicolumn{2}{c}{Overall accuracy} & \multicolumn{2}{c}{F1 score} & \multirow{2}{*}{Patch density} & \multirow{2}{*}{Edge density} & \multirow{2}{*}{Entropy} \\
 & t=0 & t=1 & t=0 & t=1 &  &  &  \\ \midrule
 BaseMLP & 62.9 $\pm$ 0.2 & 66.9 $\pm$ 0.2 & 46.4 $\pm$ 0.4 & 50.2 $\pm$ 0.4 & 46.2 $\pm$ 1.2 & 615 $\pm$ 9 & 1.168 $\pm$ 0.009 \\
 BaseGNN & 67.5 $\pm$ 0.0 & 70.4 $\pm$ 0.1 & 52.8 $\pm$ 0.2 & 56.0 $\pm$ 0.3 & 26.5 $\pm$ 2.3 & 414 $\pm$ 19 & 1.021 $\pm$ 0.017 \\
 GUNet & \textbf{69.3 $\pm$ 0.3} & \textbf{71.6 $\pm$ 0.3} & \textbf{55.1 $\pm$ 0.6} & \textbf{57.8 $\pm$ 0.6} & \textbf{18.5 $\pm$ 0.4} & \textbf{320 $\pm$ 5} & 0.881 $\pm$ 0.007 \\ \midrule
 BaseCNN & 66.8 $\pm$ 0.2 & 69.6 $\pm$ 0.4 & 50.3 $\pm$ 0.8 & 53.3 $\pm$ 1.1 & 27.6 $\pm$ 2.4 & 430 $\pm$ 26 & 1.033 $\pm$ 0.027 \\
 UNet & 68.7 $\pm$ 0.2 & 71.3 $\pm$ 0.3 & 51.5 $\pm$ 1.1 & 54.2 $\pm$ 1.4 & 24.3 $\pm$ 2.0 & 358 $\pm$ 14 & 0.913 $\pm$ 0.019 \\
 UNet++ & 68.1 $\pm$ 0.5 & 70.7 $\pm$ 0.4 & 49.8 $\pm$ 0.6 & 52.4 $\pm$ 0.5 & 24.5 $\pm$ 1.0 & 369 $\pm$ 12 & 0.914 $\pm$ 0.013 \\
 DeepLabV3 &\textbf{ 69.3 $\pm$ 0.4} & \textbf{71.6 $\pm$ 0.3} & 54.8 $\pm$ 0.0 & 57.3 $\pm$ 0.1 & 18.6 $\pm$ 1.0 & 323 $\pm$ 13 & 0.865 $\pm$ 0.025 \\
 Segformer & 67.9 $\pm$ 0.4 & 70.0 $\pm$ 0.5 & 53.2 $\pm$ 0.9 & 55.6 $\pm$ 1.0 & 19.2 $\pm$ 0.6 & 321 $\pm$ 13 & \textbf{0.812 $\pm$ 0.027} \\ \bottomrule
\end{tabular}
\end{subtable}
\end{table}

\begin{table}[H]
 
\ContinuedFloat
\begin{subtable}{\textwidth}
\centering
\scriptsize
\subcaption{MMU=$10px$}
\begin{tabular}{lccccccc}
\toprule
 \multirow{2}{*}{Method} & \multicolumn{2}{c}{Overall accuracy} & \multicolumn{2}{c}{F1 score} & \multirow{2}{*}{Patch density} & \multirow{2}{*}{Edge density} & \multirow{2}{*}{Entropy} \\
 & t=0 & t=1 & t=0 & t=1 &  &  &  \\ \midrule
 BaseMLP & 63.3 $\pm$ 0.2 & 66.6 $\pm$ 0.3 & 46.6 $\pm$ 0.2 & 49.6 $\pm$ 0.3 & 25.9 $\pm$ 0.3 & 469 $\pm$ 3 & 1.113 $\pm$ 0.004 \\
 BaseGNN & 67.2 $\pm$ 0.0 & 69.7 $\pm$ 0.1 & 52.6 $\pm$ 0.5 & 55.4 $\pm$ 0.5 & 17.2 $\pm$ 0.4 & 339 $\pm$ 6 & 0.974 $\pm$ 0.007 \\
 GUNet & 68.7 $\pm$ 0.1 & 70.7 $\pm$ 0.1 & 54.2 $\pm$ 0.5 & 56.4 $\pm$ 0.5 & \textbf{12.5 $\pm$ 0.2} & \textbf{269 $\pm$ 4} & 0.835 $\pm$ 0.007 \\ \midrule
 BaseCNN & 66.8 $\pm$ 0.3 & 69.2 $\pm$ 0.4 & 49.9 $\pm$ 0.8 & 52.4 $\pm$ 1.0 & 17.2 $\pm$ 0.9 & 343 $\pm$ 14 & 0.978 $\pm$ 0.021 \\
 UNet & 68.3 $\pm$ 0.2 & 70.6 $\pm$ 0.3 & 50.7 $\pm$ 1.2 & 53.2 $\pm$ 1.4 & 17.2 $\pm$ 1.3 & 301 $\pm$ 10 & 0.884 $\pm$ 0.019 \\
 UNet++ & 67.8 $\pm$ 0.5 & 70.1 $\pm$ 0.4 & 49.5 $\pm$ 0.6 & 51.8 $\pm$ 0.5 & 17.0 $\pm$ 0.9 & 308 $\pm$ 11 & 0.882 $\pm$ 0.015 \\
 DeepLabV3 & \textbf{69.0 $\pm$ 0.4} & \textbf{71.0 $\pm$ 0.3} & \textbf{54.4 $\pm$ 0.1} & \textbf{56.6 $\pm$ 0.2} & 12.8 $\pm$ 0.6 & 271 $\pm$ 10 & 0.831 $\pm$ 0.023 \\
 Segformer & 67.6 $\pm$ 0.4 & 69.5 $\pm$ 0.5 & 52.7 $\pm$ 0.8 & 54.8 $\pm$ 1.0 & 13.1 $\pm$ 0.4 & 268 $\pm$ 10 & \textbf{0.779 $\pm$ 0.026} \\ \bottomrule
\end{tabular}
\end{subtable}
\end{table}

\begin{table}[H]
 
\ContinuedFloat
\begin{subtable}{\textwidth}
\centering
\scriptsize
\subcaption{MMU=$20px$}
\begin{tabular}{lccccccc}
\toprule
 \multirow{2}{*}{Method} & \multicolumn{2}{c}{Overall accuracy} & \multicolumn{2}{c}{F1 score} & \multirow{2}{*}{Patch density} & \multirow{2}{*}{Edge density} & \multirow{2}{*}{Entropy} \\
 & t=0 & t=1 & t=0 & t=1 &  &  &  \\ \midrule
 BaseMLP & 64.1 $\pm$ 0.0 & 66.6 $\pm$ 0.0 & 47.8 $\pm$ 0.1 & 50.5 $\pm$ 0.1 & 15.8 $\pm$ 0.0 & 356 $\pm$ 1 & 1.064 $\pm$ 0.001 \\
 BaseGNN & 67.0 $\pm$ 0.0 & 68.9 $\pm$ 0.0 & 52.4 $\pm$ 0.1 & 54.4 $\pm$ 0.1 & 10.3 $\pm$ 0.1 & 255 $\pm$ 2 & 0.896 $\pm$ 0.004 \\
 GUNet & 67.5 $\pm$ 0.1 & 69.2 $\pm$ 0.1 & 52.3 $\pm$ 0.5 & 54.2 $\pm$ 0.4 & \textbf{9.0 $\pm$ 0.2} & 228 $\pm$ 5 & 0.805 $\pm$ 0.010 \\ \midrule
 BaseCNN & 66.3 $\pm$ 0.4 & 68.3 $\pm$ 0.5 & 49.1 $\pm$ 1.0 & 51.0 $\pm$ 1.1 & 11.2 $\pm$ 0.4 & 272 $\pm$ 8 & 0.913 $\pm$ 0.016 \\
 UNet & 68.0 $\pm$ 0.2 & 69.8 $\pm$ 0.2 & 50.0 $\pm$ 1.2 & 52.0 $\pm$ 1.4 & 11.5 $\pm$ 0.7 & 245 $\pm$ 7 & 0.837 $\pm$ 0.018 \\
 UNet++ & 67.5 $\pm$ 0.5 & 69.4 $\pm$ 0.4 & 48.9 $\pm$ 0.6 & 50.8 $\pm$ 0.4 & 11.2 $\pm$ 0.6 & 247 $\pm$ 9 & 0.832 $\pm$ 0.015 \\
 DeepLabV3 & \textbf{68.6 $\pm$ 0.4} & \textbf{70.3 $\pm$ 0.4} & \textbf{53.8 $\pm$ 0.1} & \textbf{55.6 $\pm$ 0.2} & \textbf{9.0 $\pm$ 0.3} & 222 $\pm$ 7 & 0.787 $\pm$ 0.020 \\
 Segformer & 67.2 $\pm$ 0.3 & 68.8 $\pm$ 0.4 & 52.1 $\pm$ 0.8 & 53.9 $\pm$ 0.9 & \textbf{9.0 $\pm$ 0.2} & \textbf{216 $\pm$ 9} & \textbf{0.736 $\pm$ 0.025} \\ \bottomrule
\end{tabular}
\end{subtable}
\end{table}

\begin{table}[H]
 
\ContinuedFloat
\begin{subtable}{\textwidth}
\centering
\scriptsize
\subcaption{MMU=$40px$}
\begin{tabular}{lccccccc}
\toprule
 \multirow{2}{*}{Method} & \multicolumn{2}{c}{Overall accuracy} & \multicolumn{2}{c}{F1 score} & \multirow{2}{*}{Patch density} & \multirow{2}{*}{Edge density} & \multirow{2}{*}{Entropy} \\
 & t=0 & t=1 & t=0 & t=1 &  &  &  \\ \midrule
 BaseMLP & 63.9 $\pm$ 0.1 & 65.9 $\pm$ 0.1 & 48.0 $\pm$ 0.2 & 50.0 $\pm$ 0.2 & 10.0 $\pm$ 0.1 & 268 $\pm$ 2 & 0.993 $\pm$ 0.006 \\
 BaseGNN & 66.0 $\pm$ 0.4 & 67.4 $\pm$ 0.4 & 50.6 $\pm$ 1.0 & 52.1 $\pm$ 1.1 & 7.1 $\pm$ 0.3 & 200 $\pm$ 8 & 0.821 $\pm$ 0.025 \\
 GUNet & 66.3 $\pm$ 0.1 & 67.7 $\pm$ 0.1 & 50.7 $\pm$ 0.2 & 52.2 $\pm$ 0.2 & \textbf{6.2 $\pm$ 0.1} & 177 $\pm$ 3 & 0.723 $\pm$ 0.006 \\ \midrule
 BaseCNN & 65.9 $\pm$ 0.4 & 67.5 $\pm$ 0.5 & 48.2 $\pm$ 0.9 & 49.7 $\pm$ 1.0 & 7.5 $\pm$ 0.2 & 210 $\pm$ 4 & 0.835 $\pm$ 0.013 \\
 UNet & 67.5 $\pm$ 0.1 & 68.9 $\pm$ 0.2 & 49.3 $\pm$ 1.0 & 50.7 $\pm$ 1.2 & 7.4 $\pm$ 0.3 & 190 $\pm$ 5 & 0.769 $\pm$ 0.015 \\
 UNet++ & 67.0 $\pm$ 0.5 & 68.4 $\pm$ 0.4 & 48.0 $\pm$ 0.5 & 49.4 $\pm$ 0.4 & 7.3 $\pm$ 0.3 & 191 $\pm$ 6 & 0.764 $\pm$ 0.012 \\
 DeepLabV3 & \textbf{68.1 $\pm$ 0.5} & \textbf{69.4 $\pm$ 0.5} & \textbf{52.8 $\pm$ 0.1} & \textbf{54.4 $\pm$ 0.2} & 6.3 $\pm$ 0.2 & 177 $\pm$ 4 & 0.730 $\pm$ 0.016 \\
 Segformer & 66.8 $\pm$ 0.4 & 68.1 $\pm$ 0.5 & 51.3 $\pm$ 0.9 & 52.6 $\pm$ 0.9 & \textbf{6.2 $\pm$ 0.2} & \textbf{169 $\pm$ 7} & \textbf{0.680 $\pm$ 0.025} \\ \bottomrule
\end{tabular}
\end{subtable}
\end{table}

% 1/4 dataset

\begin{table}[H]
\caption{Metrics for the $1/4$ dataset, accompanies Figure 6(b)}
\begin{subtable}{\textwidth}
\centering
\scriptsize
\subcaption{MMU=$1px$}
\begin{tabular}{lccccccc}
\toprule
 \multirow{2}{*}{Method} & \multicolumn{2}{c}{Overall accuracy} & \multicolumn{2}{c}{F1 score} & \multirow{2}{*}{Patch density} & \multirow{2}{*}{Edge density} & \multirow{2}{*}{Entropy} \\
 & t=0 & t=1 & t=0 & t=1 &  &  &  \\ \midrule
 BaseMLP & 59.4 $\pm$ 0.3 & 64.3 $\pm$ 0.6 & 42.1 $\pm$ 0.5 & 46.6 $\pm$ 0.7 & 108.2 $\pm$ 13.8 & 835 $\pm$ 57 & 1.130 $\pm$ 0.030 \\
 BaseGNN & 66.1 $\pm$ 0.1 & 70.0 $\pm$ 0.1 & 49.2 $\pm$ 0.1 & 53.0 $\pm$ 0.2 & 62.3 $\pm$ 2.0 & 605 $\pm$ 8 & 1.102 $\pm$ 0.008 \\
 GUNet & \textbf{69.2 $\pm$ 0.1} & \textbf{72.1 $\pm$ 0.1} & \textbf{53.5 $\pm$ 0.7} & \textbf{56.7 $\pm$ 0.8} & 37.1 $\pm$ 1.1 & 437 $\pm$ 8 & 1.032 $\pm$ 0.011 \\ \midrule
 BaseCNN & 65.3 $\pm$ 0.2 & 70.0 $\pm$ 0.1 & 49.3 $\pm$ 0.1 & 54.2 $\pm$ 0.2 & 101.4 $\pm$ 3.9 & 810 $\pm$ 22 & 1.193 $\pm$ 0.017 \\
 UNet & 66.5 $\pm$ 0.0 & 69.6 $\pm$ 0.1 & 46.3 $\pm$ 1.7 & 49.7 $\pm$ 1.6 & 41.1 $\pm$ 3.9 & 490 $\pm$ 23 & 0.994 $\pm$ 0.002 \\
 UNet++ & 65.8 $\pm$ 0.4 & 69.1 $\pm$ 0.5 & 47.9 $\pm$ 0.7 & 51.0 $\pm$ 0.7 & 43.8 $\pm$ 6.5 & 507 $\pm$ 26 & 1.032 $\pm$ 0.028 \\
 DeepLabV3 & 67.7 $\pm$ 0.1 & 70.6 $\pm$ 0.3 & 51.9 $\pm$ 0.9 & 55.3 $\pm$ 1.1 & \textbf{31.5 $\pm$ 1.5} & \textbf{419 $\pm$ 10} & 0.948 $\pm$ 0.015 \\
 Segformer & 66.1 $\pm$ 0.3 & 68.7 $\pm$ 0.2 & 47.8 $\pm$ 0.7 & 50.7 $\pm$ 0.7 & 36.2 $\pm$ 1.1 & 451 $\pm$ 11 & \textbf{0.944 $\pm$ 0.021} \\ \bottomrule
\end{tabular}
\end{subtable}
\end{table}

\begin{table}[H]
 
\ContinuedFloat
\begin{subtable}{\textwidth}
\centering
\scriptsize
\subcaption{MMU=$5px$}
\begin{tabular}{lccccccc}
\toprule
 \multirow{2}{*}{Method} & \multicolumn{2}{c}{Overall accuracy} & \multicolumn{2}{c}{F1 score} & \multirow{2}{*}{Patch density} & \multirow{2}{*}{Edge density} & \multirow{2}{*}{Entropy} \\
 & t=0 & t=1 & t=0 & t=1 &  &  &  \\ \midrule
 BaseMLP & 62.3 $\pm$ 0.2 & 66.5 $\pm$ 0.3 & 45.5 $\pm$ 0.7 & 49.3 $\pm$ 0.8 & 46.6 $\pm$ 1.9 & 621 $\pm$ 16 & 1.168 $\pm$ 0.015 \\
 BaseGNN & 66.7 $\pm$ 0.2 & 69.8 $\pm$ 0.2 & 51.6 $\pm$ 0.2 & 54.9 $\pm$ 0.2 & 30.9 $\pm$ 0.3 & 462 $\pm$ 3 & 1.054 $\pm$ 0.009 \\
 GUNet & \textbf{68.4 $\pm$ 0.1} & \textbf{70.8 $\pm$ 0.2} & \textbf{53.3 $\pm$ 0.1} & \textbf{55.8 $\pm$ 0.1} & \textbf{19.2 $\pm$ 0.4} & \textbf{331 $\pm$ 6} & 0.900 $\pm$ 0.009 \\ \midrule
 BaseCNN & 66.0 $\pm$ 0.1 & 69.0 $\pm$ 0.1 & 49.4 $\pm$ 0.1 & 52.5 $\pm$ 0.2 & 29.1 $\pm$ 0.6 & 456 $\pm$ 9 & 1.057 $\pm$ 0.012 \\
 UNet & 66.5 $\pm$ 0.2 & 69.1 $\pm$ 0.3 & 46.3 $\pm$ 1.4 & 48.8 $\pm$ 1.5 & 24.3 $\pm$ 0.3 & 378 $\pm$ 6 & 0.926 $\pm$ 0.005 \\
 UNet++ & 65.9 $\pm$ 0.6 & 68.6 $\pm$ 0.6 & 47.7 $\pm$ 0.8 & 50.3 $\pm$ 0.8 & 24.7 $\pm$ 2.0 & 391 $\pm$ 7 & 0.956 $\pm$ 0.017 \\
 DeepLabV3 & 67.6 $\pm$ 0.2 & 70.0 $\pm$ 0.3 & 51.9 $\pm$ 0.9 & 54.5 $\pm$ 1.0 & 20.2 $\pm$ 0.0 & 341 $\pm$ 1 & 0.889 $\pm$ 0.014 \\
 Segformer & 65.9 $\pm$ 0.2 & 68.1 $\pm$ 0.2 & 47.6 $\pm$ 0.7 & 50.0 $\pm$ 0.7 & 22.0 $\pm$ 0.6 & 359 $\pm$ 8 & \textbf{0.880 $\pm$ 0.018} \\ \bottomrule
\end{tabular}
\end{subtable}
\end{table}

\begin{table}[H]
 
\ContinuedFloat
\begin{subtable}{\textwidth}
\centering
\scriptsize
\subcaption{MMU=$10px$}
\begin{tabular}{lccccccc}
\toprule
 \multirow{2}{*}{Method} & \multicolumn{2}{c}{Overall accuracy} & \multicolumn{2}{c}{F1 score} & \multirow{2}{*}{Patch density} & \multirow{2}{*}{Edge density} & \multirow{2}{*}{Entropy} \\
 & t=0 & t=1 & t=0 & t=1 &  &  &  \\ \midrule
 BaseMLP & 62.9 $\pm$ 0.1 & 66.3 $\pm$ 0.1 & 46.4 $\pm$ 0.3 & 49.4 $\pm$ 0.4 & 26.5 $\pm$ 0.1 & 479 $\pm$ 2 & 1.127 $\pm$ 0.004 \\
 BaseGNN & 66.9 $\pm$ 0.0 & 69.5 $\pm$ 0.1 & 52.0 $\pm$ 0.3 & 54.7 $\pm$ 0.3 & 17.0 $\pm$ 0.2 & 342 $\pm$ 3 & 0.986 $\pm$ 0.003 \\
 GUNet & \textbf{67.6 $\pm$ 0.1} & \textbf{69.8 $\pm$ 0.1} & \textbf{52.3 $\pm$ 0.2} & \textbf{54.6 $\pm$ 0.2} & \textbf{13.0 $\pm$ 0.1} & \textbf{279 $\pm$ 3} & 0.848 $\pm$ 0.003 \\ \midrule
 BaseCNN & 65.9 $\pm$ 0.1 & 68.5 $\pm$ 0.1 & 49.1 $\pm$ 0.1 & 51.7 $\pm$ 0.2 & 17.9 $\pm$ 0.3 & 361 $\pm$ 6 & 0.997 $\pm$ 0.010 \\
 UNet & 66.3 $\pm$ 0.3 & 68.5 $\pm$ 0.2 & 45.8 $\pm$ 1.6 & 48.0 $\pm$ 1.6 & 16.4 $\pm$ 0.2 & 311 $\pm$ 1 & 0.886 $\pm$ 0.009 \\
 UNet++ & 65.7 $\pm$ 0.7 & 68.1 $\pm$ 0.7 & 47.3 $\pm$ 0.9 & 49.5 $\pm$ 0.8 & 16.5 $\pm$ 1.1 & 320 $\pm$ 4 & 0.913 $\pm$ 0.015 \\
 DeepLabV3 & 67.4 $\pm$ 0.2 & 69.5 $\pm$ 0.2 & 51.6 $\pm$ 0.8 & 53.8 $\pm$ 0.8 & 13.9 $\pm$ 0.1 & 285 $\pm$ 2 & 0.853 $\pm$ 0.014 \\
 Segformer & 65.6 $\pm$ 0.3 & 67.6 $\pm$ 0.2 & 47.2 $\pm$ 0.7 & 49.3 $\pm$ 0.7 & 14.8 $\pm$ 0.3 & 296 $\pm$ 6 & \textbf{0.841 $\pm$ 0.015} \\ \bottomrule
\end{tabular}
\end{subtable}
\end{table}

\begin{table}[H]
 
\ContinuedFloat
\begin{subtable}{\textwidth}
\centering
\scriptsize
\subcaption{MMU=$20px$}
\begin{tabular}{lccccccc}
\toprule
 \multirow{2}{*}{Method} & \multicolumn{2}{c}{Overall accuracy} & \multicolumn{2}{c}{F1 score} & \multirow{2}{*}{Patch density} & \multirow{2}{*}{Edge density} & \multirow{2}{*}{Entropy} \\
 & t=0 & t=1 & t=0 & t=1 &  &  &  \\ \midrule
 BaseMLP & 63.2 $\pm$ 0.2 & 65.7 $\pm$ 0.3 & 45.8 $\pm$ 0.9 & 48.2 $\pm$ 1.1 & 15.4 $\pm$ 0.3 & 351 $\pm$ 6 & 1.040 $\pm$ 0.016 \\
 BaseGNN & 66.3 $\pm$ 0.0 & 68.4 $\pm$ 0.0 & 51.4 $\pm$ 0.2 & 53.5 $\pm$ 0.2 & 11.2 $\pm$ 0.1 & 272 $\pm$ 2 & 0.927 $\pm$ 0.006 \\
 GUNet & 66.9 $\pm$ 0.2 & 68.7 $\pm$ 0.1 & \textbf{51.9 $\pm$ 0.4} & \textbf{53.7 $\pm$ 0.3} & \textbf{9.2 $\pm$ 0.1} & \textbf{230 $\pm$ 2} & 0.802 $\pm$ 0.004 \\ \midrule
 BaseCNN & 65.7 $\pm$ 0.1 & 67.8 $\pm$ 0.1 & 48.6 $\pm$ 0.1 & 50.6 $\pm$ 0.2 & 11.6 $\pm$ 0.2 & 282 $\pm$ 4 & 0.927 $\pm$ 0.008 \\
 UNet & 66.1 $\pm$ 0.3 & 67.9 $\pm$ 0.3 & 45.5 $\pm$ 1.7 & 47.2 $\pm$ 1.7 & 10.9 $\pm$ 0.2 & 247 $\pm$ 3 & 0.832 $\pm$ 0.011 \\
 UNet++ & 65.5 $\pm$ 0.8 & 67.4 $\pm$ 0.7 & 46.8 $\pm$ 1.0 & 48.6 $\pm$ 0.9 & 10.9 $\pm$ 0.5 & 255 $\pm$ 4 & 0.858 $\pm$ 0.013 \\
 DeepLabV3 & \textbf{67.1 $\pm$ 0.2} & \textbf{68.8 $\pm$ 0.3} & 51.2 $\pm$ 0.9 & 52.9 $\pm$ 0.9 & 9.5 $\pm$ 0.1 & 232 $\pm$ 3 & 0.805 $\pm$ 0.014 \\
 Segformer & 65.3 $\pm$ 0.3 & 66.9 $\pm$ 0.3 & 46.7 $\pm$ 0.7 & 48.5 $\pm$ 0.7 & 9.9 $\pm$ 0.2 & 237 $\pm$ 4 & \textbf{0.791 $\pm$ 0.012} \\ \bottomrule
\end{tabular}
\end{subtable}
\end{table}

\begin{table}[H]
 
\ContinuedFloat
\begin{subtable}{\textwidth}
\centering
\scriptsize
\subcaption{MMU=$40px$}
\begin{tabular}{lccccccc}
\toprule
 \multirow{2}{*}{Method} & \multicolumn{2}{c}{Overall accuracy} & \multicolumn{2}{c}{F1 score} & \multirow{2}{*}{Patch density} & \multirow{2}{*}{Edge density} & \multirow{2}{*}{Entropy} \\
 & t=0 & t=1 & t=0 & t=1 &  &  &  \\ \midrule
 BaseMLP & 63.4 $\pm$ 0.1 & 65.4 $\pm$ 0.1 & 46.6 $\pm$ 0.2 & 48.5 $\pm$ 0.3 & 9.7 $\pm$ 0.1 & 263 $\pm$ 3 & 0.977 $\pm$ 0.007 \\
 BaseGNN & 65.4 $\pm$ 0.3 & 67.0 $\pm$ 0.2 & 49.5 $\pm$ 0.8 & 51.1 $\pm$ 0.7 & 7.4 $\pm$ 0.3 & 208 $\pm$ 6 & 0.840 $\pm$ 0.022 \\
 GUNet & 65.5 $\pm$ 0.1 & 66.9 $\pm$ 0.1 & 49.4 $\pm$ 0.2 & 50.8 $\pm$ 0.2 & \textbf{6.4 $\pm$ 0.1} & \textbf{182 $\pm$ 3} & 0.735 $\pm$ 0.008 \\ \midrule
 BaseCNN & 65.3 $\pm$ 0.1 & 66.9 $\pm$ 0.1 & 47.7 $\pm$ 0.2 & 49.2 $\pm$ 0.2 & 7.6 $\pm$ 0.1 & 215 $\pm$ 2 & 0.845 $\pm$ 0.007 \\
 UNet & 65.7 $\pm$ 0.4 & 67.1 $\pm$ 0.3 & 44.9 $\pm$ 1.6 & 46.3 $\pm$ 1.6 & 7.1 $\pm$ 0.1 & 190 $\pm$ 3 & 0.763 $\pm$ 0.012 \\
 UNet++ & 65.1 $\pm$ 0.8 & 66.6 $\pm$ 0.8 & 46.2 $\pm$ 1.1 & 47.5 $\pm$ 1.1 & 7.2 $\pm$ 0.2 & 197 $\pm$ 3 & 0.788 $\pm$ 0.012 \\
 DeepLabV3 & \textbf{66.6 $\pm$ 0.2} & \textbf{68.0 $\pm$ 0.3} & \textbf{50.3 $\pm$ 0.8} & \textbf{51.8 $\pm$ 0.8} & 6.6 $\pm$ 0.1 & 183 $\pm$ 3 & 0.743 $\pm$ 0.013 \\
 Segformer & 65.0 $\pm$ 0.3 & 66.2 $\pm$ 0.3 & 46.1 $\pm$ 0.6 & 47.4 $\pm$ 0.6 & 6.7 $\pm$ 0.1 & 183 $\pm$ 2 & \textbf{0.726 $\pm$ 0.009} \\ \bottomrule
\end{tabular}
\end{subtable}
\end{table}

% 1/8 dataset

\begin{table}[H]
\caption{Metrics for the $1/8$ dataset, accompanies Figure 6(c)}
\begin{subtable}{\textwidth}
\centering
\scriptsize
\subcaption{MMU=$1px$}
\begin{tabular}{lccccccc}
\toprule
 \multirow{2}{*}{Method} & \multicolumn{2}{c}{Overall accuracy} & \multicolumn{2}{c}{F1 score} & \multirow{2}{*}{Patch density} & \multirow{2}{*}{Edge density} & \multirow{2}{*}{Entropy} \\
 & t=0 & t=1 & t=0 & t=1 &  &  &  \\ \midrule
 BaseMLP & 58.9 $\pm$ 0.3 & 63.9 $\pm$ 0.4 & 39.0 $\pm$ 0.4 & 43.5 $\pm$ 0.4 & 110.6 $\pm$ 1.7 & 844 $\pm$ 5 & 1.125 $\pm$ 0.027 \\
 BaseGNN & 65.5 $\pm$ 0.1 & 69.4 $\pm$ 0.2 & 48.7 $\pm$ 0.2 & 52.5 $\pm$ 0.3 & 64.9 $\pm$ 9.0 & 616 $\pm$ 45 & 1.118 $\pm$ 0.021 \\
 GUNet & \textbf{67.0 $\pm$ 0.1} & \textbf{69.8 $\pm$ 0.1} & 49.6 $\pm$ 0.3 & 52.7 $\pm$ 0.3 & 36.4 $\pm$ 3.4 & 438 $\pm$ 22 & 1.044 $\pm$ 0.020 \\ \midrule
 BaseCNN & 63.9 $\pm$ 0.2 & 68.0 $\pm$ 0.3 & 48.1 $\pm$ 0.2 & 52.3 $\pm$ 0.4 & 80.6 $\pm$ 17.2 & 710 $\pm$ 76 & 1.189 $\pm$ 0.017 \\
 UNet & 62.4 $\pm$ 1.0 & 65.2 $\pm$ 1.3 & 35.0 $\pm$ 2.0 & 37.5 $\pm$ 2.5 & 37.4 $\pm$ 3.8 & 459 $\pm$ 55 & 0.951 $\pm$ 0.031 \\
 UNet++ & 62.8 $\pm$ 0.3 & 65.2 $\pm$ 0.2 & 31.6 $\pm$ 0.1 & 33.2 $\pm$ 0.0 & 34.3 $\pm$ 3.7 & 430 $\pm$ 27 & \textbf{0.922 $\pm$ 0.046} \\
 DeepLabV3 & 65.6 $\pm$ 0.3 & 68.4 $\pm$ 0.3 & \textbf{49.7 $\pm$ 0.2} & \textbf{52.9 $\pm$ 0.3} & \textbf{27.5 $\pm$ 1.9} & \textbf{399 $\pm$ 20} & 1.000 $\pm$ 0.046 \\
 Segformer & 65.6 $\pm$ 0.8 & 68.3 $\pm$ 0.8 & 48.8 $\pm$ 0.8 & 51.8 $\pm$ 0.8 & 33.8 $\pm$ 1.3 & 436 $\pm$ 12 & 0.929 $\pm$ 0.014 \\ \bottomrule
\end{tabular}
\end{subtable}
\end{table}

\begin{table}[H]
 
\ContinuedFloat
\begin{subtable}{\textwidth}
\centering
\scriptsize
\subcaption{MMU=$5px$}
\begin{tabular}{lccccccc}
\toprule
 \multirow{2}{*}{Method} & \multicolumn{2}{c}{Overall accuracy} & \multicolumn{2}{c}{F1 score} & \multirow{2}{*}{Patch density} & \multirow{2}{*}{Edge density} & \multirow{2}{*}{Entropy} \\
 & t=0 & t=1 & t=0 & t=1 &  &  &  \\ \midrule
 BaseMLP & 62.0 $\pm$ 0.1 & 66.2 $\pm$ 0.1 & 44.4 $\pm$ 0.1 & 48.4 $\pm$ 0.3 & 48.0 $\pm$ 0.8 & 628 $\pm$ 4 & 1.159 $\pm$ 0.005 \\
 BaseGNN & 65.9 $\pm$ 0.1 & \textbf{69.3 $\pm$ 0.1} & 49.7 $\pm$ 0.1 & \textbf{53.4 $\pm$ 0.2} & 32.7 $\pm$ 0.6 & 475 $\pm$ 6 & 1.062 $\pm$ 0.010 \\
 GUNet & \textbf{66.7 $\pm$ 0.2} & 69.2 $\pm$ 0.3 & 49.4 $\pm$ 0.1 & 51.9 $\pm$ 0.2 & \textbf{19.7 $\pm$ 0.7} & \textbf{338 $\pm$ 9} & 0.906 $\pm$ 0.015 \\ \midrule
 BaseCNN & 64.4 $\pm$ 0.0 & 67.3 $\pm$ 0.0 & 48.2 $\pm$ 0.2 & 51.1 $\pm$ 0.2 & 28.6 $\pm$ 0.4 & 453 $\pm$ 2 & 1.076 $\pm$ 0.006 \\
 UNet & 62.3 $\pm$ 1.0 & 64.6 $\pm$ 1.2 & 34.3 $\pm$ 2.1 & 36.1 $\pm$ 2.3 & 22.9 $\pm$ 0.8 & 367 $\pm$ 22 & 0.895 $\pm$ 0.020 \\
 UNet++ & 62.6 $\pm$ 0.2 & 64.7 $\pm$ 0.2 & 31.4 $\pm$ 0.2 & 32.7 $\pm$ 0.1 & 22.6 $\pm$ 1.1 & 356 $\pm$ 18 & 0.873 $\pm$ 0.043 \\
 DeepLabV3 & 65.5 $\pm$ 0.4 & 68.0 $\pm$ 0.3 & \textbf{49.5 $\pm$ 0.1} & 52.3 $\pm$ 0.2 & 21.1 $\pm$ 1.5 & 354 $\pm$ 20 & 0.941 $\pm$ 0.045 \\
 Segformer & 65.4 $\pm$ 0.8 & 67.7 $\pm$ 0.8 & 48.7 $\pm$ 0.8 & 51.1 $\pm$ 0.9 & 22.4 $\pm$ 0.9 & 361 $\pm$ 8 & \textbf{0.871 $\pm$ 0.015} \\ \bottomrule
\end{tabular}
\end{subtable}
\end{table}

\begin{table}[H]
 
\ContinuedFloat
\begin{subtable}{\textwidth}
\centering
\scriptsize
\subcaption{MMU=$10px$}
\begin{tabular}{lccccccc}
\toprule
 \multirow{2}{*}{Method} & \multicolumn{2}{c}{Overall accuracy} & \multicolumn{2}{c}{F1 score} & \multirow{2}{*}{Patch density} & \multirow{2}{*}{Edge density} & \multirow{2}{*}{Entropy} \\
 & t=0 & t=1 & t=0 & t=1 &  &  &  \\ \midrule
 BaseMLP & 62.7 $\pm$ 0.1 & 66.0 $\pm$ 0.1 & 45.1 $\pm$ 0.2 & 48.3 $\pm$ 0.2 & 26.9 $\pm$ 0.1 & 481 $\pm$ 1 & 1.126 $\pm$ 0.002 \\
 BaseGNN & 65.8 $\pm$ 0.3 & 68.4 $\pm$ 0.2 & \textbf{49.7 $\pm$ 0.3} & \textbf{52.4 $\pm$ 0.4} & 18.3 $\pm$ 0.4 & 359 $\pm$ 6 & 1.002 $\pm$ 0.010 \\
 GUNet & \textbf{66.5 $\pm$ 0.1} & \textbf{68.5 $\pm$ 0.2} & 49.5 $\pm$ 1.1 & 51.6 $\pm$ 1.2 & \textbf{12.8 $\pm$ 0.2} & \textbf{275 $\pm$ 3} & 0.841 $\pm$ 0.012 \\ \midrule
 BaseCNN & 64.4 $\pm$ 0.1 & 66.9 $\pm$ 0.1 & 47.9 $\pm$ 0.2 & 50.3 $\pm$ 0.2 & 18.1 $\pm$ 0.2 & 364 $\pm$ 5 & 1.019 $\pm$ 0.010 \\
 UNet & 62.1 $\pm$ 0.9 & 64.1 $\pm$ 1.1 & 33.8 $\pm$ 2.0 & 35.3 $\pm$ 2.2 & 15.8 $\pm$ 0.5 & 305 $\pm$ 13 & 0.860 $\pm$ 0.020 \\
 UNet++ & 62.5 $\pm$ 0.2 & 64.3 $\pm$ 0.1 & 31.2 $\pm$ 0.1 & 32.3 $\pm$ 0.2 & 15.6 $\pm$ 0.8 & 297 $\pm$ 16 & 0.840 $\pm$ 0.043 \\
 DeepLabV3 & 65.5 $\pm$ 0.4 & 67.7 $\pm$ 0.3 & 49.5 $\pm$ 0.0 & 51.8 $\pm$ 0.1 & 14.8 $\pm$ 1.1 & 299 $\pm$ 17 & 0.901 $\pm$ 0.043 \\
 Segformer & 65.2 $\pm$ 0.8 & 67.2 $\pm$ 0.8 & 48.3 $\pm$ 0.8 & 50.4 $\pm$ 0.9 & 15.4 $\pm$ 0.6 & 301 $\pm$ 8 & \textbf{0.837 $\pm$ 0.018} \\ \bottomrule
\end{tabular}
\end{subtable}
\end{table}

\begin{table}[H]
 
\ContinuedFloat
\begin{subtable}{\textwidth}
\centering
\scriptsize
\subcaption{MMU=$20px$}
\begin{tabular}{lccccccc}
\toprule
 \multirow{2}{*}{Method} & \multicolumn{2}{c}{Overall accuracy} & \multicolumn{2}{c}{F1 score} & \multirow{2}{*}{Patch density} & \multirow{2}{*}{Edge density} & \multirow{2}{*}{Entropy} \\
 & t=0 & t=1 & t=0 & t=1 &  &  &  \\ \midrule
 BaseMLP & 62.9 $\pm$ 0.1 & 65.5 $\pm$ 0.1 & 45.2 $\pm$ 0.5 & 47.7 $\pm$ 0.5 & 16.0 $\pm$ 0.2 & 361 $\pm$ 3 & 1.062 $\pm$ 0.010 \\
 BaseGNN & 65.3 $\pm$ 0.3 & \textbf{67.4 $\pm$ 0.2} & \textbf{49.3 $\pm$ 0.2} & \textbf{51.6 $\pm$ 0.2} & 11.5 $\pm$ 0.1 & 275 $\pm$ 2 & 0.927 $\pm$ 0.006 \\
 GUNet & \textbf{65.4 $\pm$ 0.2} & 67.1 $\pm$ 0.2 & 48.3 $\pm$ 0.9 & 49.9 $\pm$ 1.0 & \textbf{8.6 $\pm$ 0.3} & \textbf{219 $\pm$ 5} & 0.771 $\pm$ 0.019 \\ \midrule
 BaseCNN & 64.1 $\pm$ 0.1 & 66.2 $\pm$ 0.1 & 47.2 $\pm$ 0.3 & 49.1 $\pm$ 0.2 & 11.8 $\pm$ 0.2 & 287 $\pm$ 5 & 0.952 $\pm$ 0.012 \\
 UNet & 61.8 $\pm$ 0.9 & 63.5 $\pm$ 1.0 & 33.1 $\pm$ 1.9 & 34.3 $\pm$ 2.0 & 10.6 $\pm$ 0.4 & 244 $\pm$ 9 & 0.813 $\pm$ 0.020 \\
 UNet++ & 62.2 $\pm$ 0.2 & 63.8 $\pm$ 0.1 & 30.9 $\pm$ 0.2 & 31.8 $\pm$ 0.2 & 10.5 $\pm$ 0.6 & 238 $\pm$ 14 & 0.796 $\pm$ 0.041 \\
 DeepLabV3 & 65.3 $\pm$ 0.4 & 67.1 $\pm$ 0.4 & 49.2 $\pm$ 0.1 & 51.1 $\pm$ 0.0 & 10.2 $\pm$ 0.7 & 244 $\pm$ 14 & 0.848 $\pm$ 0.040 \\
 Segformer & 64.9 $\pm$ 0.7 & 66.5 $\pm$ 0.7 & 47.9 $\pm$ 0.7 & 49.7 $\pm$ 0.7 & 10.3 $\pm$ 0.4 & 240 $\pm$ 7 & \textbf{0.790 $\pm$ 0.021} \\ \bottomrule
\end{tabular}
\end{subtable}
\end{table}

\begin{table}[H]
 
\ContinuedFloat
\begin{subtable}{\textwidth}
\centering
\scriptsize
\subcaption{MMU=$40px$}
\begin{tabular}{lccccccc}
\toprule
 \multirow{2}{*}{Method} & \multicolumn{2}{c}{Overall accuracy} & \multicolumn{2}{c}{F1 score} & \multirow{2}{*}{Patch density} & \multirow{2}{*}{Edge density} & \multirow{2}{*}{Entropy} \\
 & t=0 & t=1 & t=0 & t=1 &  &  &  \\ \midrule
 BaseMLP & 62.7 $\pm$ 0.1 & 64.6 $\pm$ 0.1 & 45.0 $\pm$ 0.0 & 46.8 $\pm$ 0.1 & 10.1 $\pm$ 0.1 & 270 $\pm$ 2 & 1.000 $\pm$ 0.008 \\
 BaseGNN & 64.9 $\pm$ 0.3 & \textbf{66.5 $\pm$ 0.2} & 48.6 $\pm$ 0.3 & \textbf{50.2 $\pm$ 0.3} & 7.6 $\pm$ 0.1 & 212 $\pm$ 3 & 0.852 $\pm$ 0.008 \\
 GUNet & 64.3 $\pm$ 0.4 & 65.7 $\pm$ 0.4 & 46.5 $\pm$ 1.2 & 47.8 $\pm$ 1.2 & \textbf{6.4 $\pm$ 0.2} & \textbf{181 $\pm$ 6} & 0.734 $\pm$ 0.022 \\ \midrule
 BaseCNN & 63.8 $\pm$ 0.0 & 65.4 $\pm$ 0.0 & 46.4 $\pm$ 0.4 & 47.9 $\pm$ 0.4 & 7.8 $\pm$ 0.1 & 220 $\pm$ 4 & 0.869 $\pm$ 0.012 \\
 UNet & 61.6 $\pm$ 0.9 & 62.9 $\pm$ 1.0 & 32.4 $\pm$ 1.9 & 33.3 $\pm$ 1.9 & 7.1 $\pm$ 0.2 & 188 $\pm$ 6 & 0.749 $\pm$ 0.020 \\
 UNet++ & 61.9 $\pm$ 0.2 & 63.1 $\pm$ 0.2 & 30.5 $\pm$ 0.2 & 31.2 $\pm$ 0.2 & 7.0 $\pm$ 0.4 & 185 $\pm$ 10 & 0.736 $\pm$ 0.037 \\
 DeepLabV3 & \textbf{65.1 $\pm$ 0.3} & \textbf{66.5 $\pm$ 0.3} & \textbf{48.7 $\pm$ 0.1} & 50.1 $\pm$ 0.1 & 7.0 $\pm$ 0.4 & 191 $\pm$ 10 & 0.779 $\pm$ 0.036 \\
 Segformer & 64.5 $\pm$ 0.7 & 65.8 $\pm$ 0.7 & 47.2 $\pm$ 0.7 & 48.6 $\pm$ 0.7 & 6.9 $\pm$ 0.3 & 184 $\pm$ 7 & \textbf{0.726 $\pm$ 0.023} \\ \bottomrule
\end{tabular}
\end{subtable}
\end{table}

% 1/16 dataset

\begin{table}[H]
\caption{Metrics for the $1/16$ dataset, accompanies Figure 6(d)}
\begin{subtable}{\textwidth}
\centering
\scriptsize
\subcaption{MMU=$1px$}
\begin{tabular}{lccccccc}
\toprule
 \multirow{2}{*}{Method} & \multicolumn{2}{c}{Overall accuracy} & \multicolumn{2}{c}{F1 score} & \multirow{2}{*}{Patch density} & \multirow{2}{*}{Edge density} & \multirow{2}{*}{Entropy} \\
 & t=0 & t=1 & t=0 & t=1 &  &  &  \\ \midrule
 BaseMLP & 59.5 $\pm$ 0.0 & 64.7 $\pm$ 0.1 & 42.0 $\pm$ 0.1 & 46.6 $\pm$ 0.1 & 125.4 $\pm$ 2.1 & 896 $\pm$ 3 & 1.161 $\pm$ 0.003 \\
 BaseGNN & 64.6 $\pm$ 0.1 & \textbf{68.4 $\pm$ 0.2} & 47.1 $\pm$ 0.4 & 50.7 $\pm$ 0.6 & 70.7 $\pm$ 6.6 & 640 $\pm$ 28 & 1.139 $\pm$ 0.009 \\
 GUNet & \textbf{65.4 $\pm$ 0.2} & \textbf{68.4 $\pm$ 0.2} & 47.1 $\pm$ 0.2 & 50.0 $\pm$ 0.3 & 42.3 $\pm$ 0.4 & 488 $\pm$ 3 & 1.070 $\pm$ 0.008 \\ \midrule
 BaseCNN & 62.9 $\pm$ 0.2 & 66.3 $\pm$ 0.1 & 45.1 $\pm$ 1.3 & 48.6 $\pm$ 1.1 & 55.6 $\pm$ 2.7 & 580 $\pm$ 13 & 1.142 $\pm$ 0.007 \\
 UNet & 60.1 $\pm$ 0.4 & 63.2 $\pm$ 0.5 & 30.2 $\pm$ 0.1 & 32.3 $\pm$ 0.1 & 45.7 $\pm$ 7.0 & 541 $\pm$ 39 & 1.018 $\pm$ 0.033 \\
 UNet++ & 60.4 $\pm$ 0.6 & 63.8 $\pm$ 0.3 & 30.2 $\pm$ 0.4 & 32.3 $\pm$ 0.2 & 48.9 $\pm$ 3.7 & 573 $\pm$ 26 & 1.033 $\pm$ 0.008 \\
 DeepLabV3 & 63.4 $\pm$ 0.5 & 65.6 $\pm$ 0.5 & 45.5 $\pm$ 0.6 & 47.6 $\pm$ 0.6 & \textbf{20.7 $\pm$ 0.9} & \textbf{342 $\pm$ 15} & \textbf{0.912 $\pm$ 0.008} \\
 Segformer & 63.4 $\pm$ 0.6 & 66.2 $\pm$ 0.5 & \textbf{48.5 $\pm$ 1.0} & \textbf{52.0 $\pm$ 0.9} & 33.5 $\pm$ 1.5 & 436 $\pm$ 10 & 0.965 $\pm$ 0.027 \\ \bottomrule
\end{tabular}
\end{subtable}
\end{table}

\begin{table}[H]
 
\ContinuedFloat
\begin{subtable}{\textwidth}
\centering
\scriptsize
\subcaption{MMU=$5px$}
\begin{tabular}{lccccccc}
\toprule
 \multirow{2}{*}{Method} & \multicolumn{2}{c}{Overall accuracy} & \multicolumn{2}{c}{F1 score} & \multirow{2}{*}{Patch density} & \multirow{2}{*}{Edge density} & \multirow{2}{*}{Entropy} \\
  & t=0 & t=1 & t=0 & t=1 &  &  &  \\ \midrule
 BaseMLP & 61.3 $\pm$ 0.2 & 65.5 $\pm$ 0.4 & 43.5 $\pm$ 0.3 & 47.3 $\pm$ 0.5 & 49.2 $\pm$ 3.8 & 639 $\pm$ 29 & 1.175 $\pm$ 0.027 \\
 BaseGNN & 64.2 $\pm$ 0.2 & 67.9 $\pm$ 0.4 & 47.3 $\pm$ 1.4 & 51.0 $\pm$ 1.4 & 38.1 $\pm$ 4.7 & 532 $\pm$ 41 & 1.117 $\pm$ 0.036 \\
 GUNet & \textbf{65.5 $\pm$ 0.1} & \textbf{68.1 $\pm$ 0.0} & 47.8 $\pm$ 1.1 & 50.2 $\pm$ 1.1 & 21.2 $\pm$ 0.9 & 361 $\pm$ 8 & 0.947 $\pm$ 0.011 \\ \midrule
 BaseCNN & 63.0 $\pm$ 0.1 & 65.9 $\pm$ 0.1 & 44.9 $\pm$ 1.3 & 47.6 $\pm$ 1.2 & 26.3 $\pm$ 0.5 & 433 $\pm$ 5 & 1.059 $\pm$ 0.008 \\
 UNet & 60.5 $\pm$ 0.4 & 63.1 $\pm$ 0.5 & 30.4 $\pm$ 0.2 & 32.0 $\pm$ 0.2 & 25.5 $\pm$ 2.3 & 407 $\pm$ 20 & 0.940 $\pm$ 0.020 \\
 UNet++ & 60.9 $\pm$ 0.4 & 63.6 $\pm$ 0.3 & 30.5 $\pm$ 0.3 & 32.1 $\pm$ 0.3 & 27.0 $\pm$ 1.5 & 429 $\pm$ 10 & 0.958 $\pm$ 0.008 \\
 DeepLabV3 & 63.4 $\pm$ 0.4 & 65.5 $\pm$ 0.5 & 45.5 $\pm$ 0.7 & 47.4 $\pm$ 0.6 & \textbf{18.0 $\pm$ 0.5} & \textbf{318 $\pm$ 10} & \textbf{0.868 $\pm$ 0.004} \\
 Segformer & 63.4 $\pm$ 0.5 & 65.8 $\pm$ 0.5 & \textbf{48.6 $\pm$ 1.1} & \textbf{51.3 $\pm$ 1.0} & 24.2 $\pm$ 1.5 & 373 $\pm$ 12 & 0.905 $\pm$ 0.028 \\ \bottomrule
\end{tabular}
\end{subtable}
\end{table}

\begin{table}[H]
 
\ContinuedFloat
\begin{subtable}{\textwidth}
\centering
\scriptsize
\subcaption{MMU=$10px$}
\begin{tabular}{lccccccc}
\toprule
 \multirow{2}{*}{Method} & \multicolumn{2}{c}{Overall accuracy} & \multicolumn{2}{c}{F1 score} & \multirow{2}{*}{Patch density} & \multirow{2}{*}{Edge density} & \multirow{2}{*}{Entropy} \\
 & t=0 & t=1 & t=0 & t=1 &  &  &  \\ \midrule
 BaseMLP & 61.9 $\pm$ 0.1 & 65.3 $\pm$ 0.1 & 44.0 $\pm$ 0.2 & 47.2 $\pm$ 0.2 & 28.1 $\pm$ 0.7 & 495 $\pm$ 8 & 1.141 $\pm$ 0.015 \\
 BaseGNN & \textbf{64.3 $\pm$ 0.3} & \textbf{67.0 $\pm$ 0.4} & 47.8 $\pm$ 0.1 & 50.6 $\pm$ 0.2 & 20.3 $\pm$ 1.4 & 384 $\pm$ 21 & 1.043 $\pm$ 0.021 \\
 GUNet & 64.2 $\pm$ 0.9 & 66.1 $\pm$ 0.9 & 44.9 $\pm$ 1.5 & 46.7 $\pm$ 1.5 & \textbf{12.6 $\pm$ 0.5} & \textbf{272 $\pm$ 8} & \textbf{0.834 $\pm$ 0.009} \\ \midrule
 BaseCNN & 62.9 $\pm$ 0.2 & 65.3 $\pm$ 0.1 & 44.6 $\pm$ 1.3 & 46.8 $\pm$ 1.3 & 17.5 $\pm$ 0.3 & 359 $\pm$ 5 & 1.011 $\pm$ 0.008 \\
 UNet & 60.5 $\pm$ 0.5 & 62.7 $\pm$ 0.6 & 30.3 $\pm$ 0.3 & 31.6 $\pm$ 0.3 & 16.8 $\pm$ 1.1 & 331 $\pm$ 13 & 0.896 $\pm$ 0.016 \\
 UNet++ & 61.1 $\pm$ 0.3 & 63.3 $\pm$ 0.3 & 30.5 $\pm$ 0.3 & 31.8 $\pm$ 0.2 & 17.6 $\pm$ 0.7 & 344 $\pm$ 5 & 0.914 $\pm$ 0.008 \\
 DeepLabV3 & 63.3 $\pm$ 0.4 & 65.2 $\pm$ 0.4 & 45.3 $\pm$ 0.6 & 47.0 $\pm$ 0.6 & 13.0 $\pm$ 0.2 & 273 $\pm$ 5 & 0.838 $\pm$ 0.002 \\
 Segformer & 63.2 $\pm$ 0.5 & 65.3 $\pm$ 0.5 & \textbf{48.2 $\pm$ 1.0} & \textbf{50.5 $\pm$ 1.0} & 16.7 $\pm$ 1.1 & 312 $\pm$ 11 & 0.872 $\pm$ 0.030 \\ \bottomrule
\end{tabular}
\end{subtable}
\end{table}

\begin{table}[H]
 
\ContinuedFloat
\begin{subtable}{\textwidth}
\centering
\scriptsize
\subcaption{MMU=$20px$}
\begin{tabular}{lccccccc}
\toprule
 \multirow{2}{*}{Method} & \multicolumn{2}{c}{Overall accuracy} & \multicolumn{2}{c}{F1 score} & \multirow{2}{*}{Patch density} & \multirow{2}{*}{Edge density} & \multirow{2}{*}{Entropy} \\
 & t=0 & t=1 & t=0 & t=1 &  &  &  \\ \midrule
 BaseMLP & 62.3 $\pm$ 0.2 & 64.9 $\pm$ 0.1 & 44.4 $\pm$ 0.4 & 46.8 $\pm$ 0.4 & 16.3 $\pm$ 0.2 & 368 $\pm$ 3 & 1.073 $\pm$ 0.007 \\
 BaseGNN & \textbf{64.2 $\pm$ 0.4} & \textbf{66.2 $\pm$ 0.4} & 47.3 $\pm$ 1.0 & 49.3 $\pm$ 1.1 & 11.9 $\pm$ 0.5 & 287 $\pm$ 8 & 0.949 $\pm$ 0.020 \\
 GUNet & 63.8 $\pm$ 0.3 & 65.5 $\pm$ 0.3 & 43.9 $\pm$ 2.6 & 45.6 $\pm$ 2.6 & \textbf{8.9 $\pm$ 0.3} & 226 $\pm$ 8 & \textbf{0.782 $\pm$ 0.020} \\ \midrule
 BaseCNN & 62.6 $\pm$ 0.2 & 64.6 $\pm$ 0.1 & 44.0 $\pm$ 1.3 & 45.9 $\pm$ 1.3 & 11.7 $\pm$ 0.2 & 287 $\pm$ 4 & 0.949 $\pm$ 0.008 \\
 UNet & 60.5 $\pm$ 0.6 & 62.3 $\pm$ 0.6 & 30.0 $\pm$ 0.4 & 31.1 $\pm$ 0.4 & 11.1 $\pm$ 0.5 & 261 $\pm$ 8 & 0.841 $\pm$ 0.013 \\
 UNet++ & 61.0 $\pm$ 0.3 & 62.7 $\pm$ 0.3 & 30.2 $\pm$ 0.3 & 31.3 $\pm$ 0.3 & 11.5 $\pm$ 0.3 & 269 $\pm$ 3 & 0.857 $\pm$ 0.008 \\
 DeepLabV3 & 63.2 $\pm$ 0.3 & 64.8 $\pm$ 0.3 & 45.1 $\pm$ 0.5 & 46.5 $\pm$ 0.5 & 9.2 $\pm$ 0.0 & \textbf{224 $\pm$ 2} & 0.795 $\pm$ 0.003 \\
 Segformer & 63.0 $\pm$ 0.5 & 64.7 $\pm$ 0.5 & \textbf{47.9 $\pm$ 1.0} & \textbf{49.8 $\pm$ 1.0} & 11.1 $\pm$ 0.6 & 250 $\pm$ 9 & 0.826 $\pm$ 0.030 \\ \bottomrule
\end{tabular}
\end{subtable}
\end{table}

\begin{table}[H]
 
\ContinuedFloat
\begin{subtable}{\textwidth}
\centering
\scriptsize
\subcaption{MMU=$40px$}
\begin{tabular}{lccccccc}
\toprule
 \multirow{2}{*}{Method} & \multicolumn{2}{c}{Overall accuracy} & \multicolumn{2}{c}{F1 score} & \multirow{2}{*}{Patch density} & \multirow{2}{*}{Edge density} & \multirow{2}{*}{Entropy} \\
 & t=0 & t=1 & t=0 & t=1 &  &  &  \\ \midrule
 BaseMLP & 62.1 $\pm$ 0.1 & 64.0 $\pm$ 0.1 & 43.6 $\pm$ 0.1 & 45.4 $\pm$ 0.2 & 10.2 $\pm$ 0.0 & 275 $\pm$ 0 & 1.006 $\pm$ 0.003 \\
 BaseGNN & \textbf{63.4 $\pm$ 0.4} & \textbf{64.9 $\pm$ 0.4} & 46.4 $\pm$ 1.2 & 47.9 $\pm$ 1.3 & 7.8 $\pm$ 0.3 & 218 $\pm$ 6 & 0.871 $\pm$ 0.022 \\
 GUNet & 62.4 $\pm$ 0.6 & 63.6 $\pm$ 0.7 & 42.3 $\pm$ 3.0 & 43.4 $\pm$ 3.2 & \textbf{6.0 $\pm$ 0.4} & \textbf{170 $\pm$ 13} & \textbf{0.695 $\pm$ 0.043} \\ \midrule
 BaseCNN & 62.3 $\pm$ 0.2 & 63.9 $\pm$ 0.2 & 43.2 $\pm$ 1.3 & 44.6 $\pm$ 1.3 & 7.9 $\pm$ 0.1 & 221 $\pm$ 3 & 0.870 $\pm$ 0.007 \\
 UNet & 60.3 $\pm$ 0.6 & 61.7 $\pm$ 0.6 & 29.8 $\pm$ 0.5 & 30.6 $\pm$ 0.4 & 7.3 $\pm$ 0.2 & 199 $\pm$ 6 & 0.772 $\pm$ 0.012 \\
 UNet++ & 60.8 $\pm$ 0.2 & 62.2 $\pm$ 0.2 & 30.0 $\pm$ 0.2 & 30.8 $\pm$ 0.2 & 7.5 $\pm$ 0.1 & 204 $\pm$ 3 & 0.786 $\pm$ 0.008 \\
 DeepLabV3 & 62.8 $\pm$ 0.3 & 64.1 $\pm$ 0.3 & 44.6 $\pm$ 0.5 & 45.8 $\pm$ 0.5 & 6.4 $\pm$ 0.0 & 178 $\pm$ 1 & 0.739 $\pm$ 0.004 \\
 Segformer & 62.6 $\pm$ 0.6 & 64.0 $\pm$ 0.6 & \textbf{47.4 $\pm$ 1.1} & \textbf{48.9 $\pm$ 1.1} & 7.3 $\pm$ 0.4 & 193 $\pm$ 7 & 0.762 $\pm$ 0.029 \\ \bottomrule
\end{tabular}
\end{subtable}
\end{table}

\section{Supplementary material for Section 3.3}
\label{app:supplementary_material_for_section_3_3}
\counterwithin{table}{section}

In this section, we provide the detailed results for all accuracy and fragmentation metrics regarding the experiments on the effect of the incorporation of features from a pre-trained model from Section 3.3.

\begin{table}[H]
\caption{Metrics for models with and without pre-trained features for the full dataset, accompanies Figure 7(a)}
\begin{subtable}{\textwidth}
\centering
\scriptsize
\subcaption{MMU=$1px$}
\begin{tabular}{lccccccc}
\toprule
\multirow{2}{*}{Method} & \multicolumn{2}{c}{Overall accuracy} & \multicolumn{2}{c}{F1 score} & \multirow{2}{*}{Patch density} & \multirow{2}{*}{Edge density} & \multirow{2}{*}{Entropy} \\
 & t=0 & t=1 & t=0 & t=1 &  &  &  \\ \midrule
 BaseMLP & 60.1 $\pm$ 0.0 & 65.4 $\pm$ 0.0 & 43.1 $\pm$ 0.1 & 47.7 $\pm$ 0.1 & 126.7 $\pm$ 0.5 & 907 $\pm$ 2 & 1.166 $\pm$ 0.001 \\
 BaseGNN & 67.5 $\pm$ 0.0 & 71.8 $\pm$ 0.1 & 52.1 $\pm$ 0.1 & 56.4 $\pm$ 0.1 & 74.4 $\pm$ 4.5 & 665 $\pm$ 20 & 1.124 $\pm$ 0.006 \\
 BaseCNN & 67.0 $\pm$ 0.3 & 70.7 $\pm$ 0.1 & 50.9 $\pm$ 0.1 & 54.8 $\pm$ 0.4 & 64.5 $\pm$ 9.9 & 629 $\pm$ 39 & 1.123 $\pm$ 0.011 \\ \midrule
 BaseMLP + PT & 70.7 $\pm$ 0.1 & \textbf{72.9 $\pm$ 0.1} & \textbf{55.7 $\pm$ 0.2} & \textbf{58.6 $\pm$ 0.2} & 19.4 $\pm$ 0.1 & 318 $\pm$ 1 & 0.888 $\pm$ 0.004 \\ 
 BaseGNN + PT & 68.6 $\pm$ 0.3 & 70.5 $\pm$ 0.1 & 49.8 $\pm$ 0.5 & 52.0 $\pm$ 0.6 & \textbf{16.9 $\pm$ 0.9} & \textbf{295 $\pm$ 13} & 0.866 $\pm$ 0.021 \\ 
 BaseCNN + PT & \textbf{70.9 $\pm$ 0.0} & \textbf{72.9 $\pm$ 0.0} & 53.5 $\pm$ 0.3 & 56.0 $\pm$ 0.2 & 19.3 $\pm$ 0.1 & 313 $\pm$ 0 & \textbf{0.851 $\pm$ 0.003} \\ \bottomrule
\end{tabular}
\end{subtable}
\end{table}

\begin{table}[H]
 
\ContinuedFloat
\begin{subtable}{\textwidth}
\centering
\scriptsize
\subcaption{MMU=$5px$}
\begin{tabular}{lccccccc}
\toprule
\multirow{2}{*}{Method} & \multicolumn{2}{c}{Overall accuracy} & \multicolumn{2}{c}{F1 score} & \multirow{2}{*}{Patch density} & \multirow{2}{*}{Edge density} & \multirow{2}{*}{Entropy} \\
 & t=0 & t=1 & t=0 & t=1 &  &  &  \\ \midrule
 BaseMLP & 63.1 $\pm$ 0.3 & 67.1 $\pm$ 0.3 & 46.8 $\pm$ 0.5 & 50.5 $\pm$ 0.4 & 43.9 $\pm$ 1.9 & 599 $\pm$ 15 & 1.150 $\pm$ 0.018 \\
 BaseGNN & 68.6 $\pm$ 0.1 & 71.4 $\pm$ 0.1 & 54.6 $\pm$ 0.1 & 57.6 $\pm$ 0.2 & 25.4 $\pm$ 2.2 & 403 $\pm$ 21 & 1.010 $\pm$ 0.020 \\
 BaseCNN & 67.0 $\pm$ 0.2 & 69.8 $\pm$ 0.2 & 50.5 $\pm$ 0.2 & 53.4 $\pm$ 0.3 & 25.9 $\pm$ 0.4 & 420 $\pm$ 1 & 1.028 $\pm$ 0.003 \\ \midrule
 BaseMLP + PT & 70.2 $\pm$ 0.1 & 72.4 $\pm$ 0.1 & \textbf{55.0 $\pm$ 0.3} & \textbf{57.5 $\pm$ 0.4} & 16.5 $\pm$ 0.1 & 301 $\pm$ 0 & 0.860 $\pm$ 0.003 \\ 
 BaseGNN + PT & \textbf{70.6 $\pm$ 0.0} & \textbf{72.5 $\pm$ 0.1} & 54.6 $\pm$ 0.4 & 56.7 $\pm$ 0.5 & 15.9 $\pm$ 0.3 & 293 $\pm$ 5 & 0.843 $\pm$ 0.008 \\ 
 BaseCNN + PT & 70.4 $\pm$ 0.0 & 72.4 $\pm$ 0.0 & 52.7 $\pm$ 0.3 & 55.0 $\pm$ 0.3 & \textbf{15.7 $\pm$ 0.0} & \textbf{287 $\pm$ 1} & \textbf{0.808 $\pm$ 0.003} \\ \bottomrule
\end{tabular}
\end{subtable}
\end{table}

\begin{table}[H]
 
\ContinuedFloat
\begin{subtable}{\textwidth}
\centering
\scriptsize
\subcaption{MMU=$10px$}
\begin{tabular}{lccccccc}
\toprule
\multirow{2}{*}{Method} & \multicolumn{2}{c}{Overall accuracy} & \multicolumn{2}{c}{F1 score} & \multirow{2}{*}{Patch density} & \multirow{2}{*}{Edge density} & \multirow{2}{*}{Entropy} \\
 & t=0 & t=1 & t=0 & t=1 &  &  &  \\ \midrule
 BaseMLP & 63.9 $\pm$ 0.0 & 67.2 $\pm$ 0.0 & 47.0 $\pm$ 0.0 & 50.0 $\pm$ 0.1 & 24.5 $\pm$ 0.0 & 453 $\pm$ 0 & 1.086 $\pm$ 0.001 \\
 BaseGNN & 68.2 $\pm$ 0.2 & 70.5 $\pm$ 0.2 & 54.0 $\pm$ 0.1 & \textbf{56.6 $\pm$ 0.1} & 14.9 $\pm$ 0.2 & 308 $\pm$ 3 & 0.930 $\pm$ 0.003 \\
 BaseCNN & 66.9 $\pm$ 0.2 & 69.2 $\pm$ 0.2 & 50.0 $\pm$ 0.3 & 52.4 $\pm$ 0.3 & 16.9 $\pm$ 0.1 & 342 $\pm$ 0 & 0.977 $\pm$ 0.003 \\ \midrule
 BaseMLP + PT & 69.8 $\pm$ 0.1 & 71.8 $\pm$ 0.1 & \textbf{54.1 $\pm$ 0.2} & 56.4 $\pm$ 0.3 & 12.1 $\pm$ 0.1 & 262 $\pm$ 1 & 0.828 $\pm$ 0.002 \\ 
 BaseGNN + PT & \textbf{70.1 $\pm$ 0.1} & \textbf{72.0 $\pm$ 0.0} & 53.4 $\pm$ 0.2 & 55.7 $\pm$ 0.2 & 11.5 $\pm$ 0.0 & 251 $\pm$ 1 & 0.805 $\pm$ 0.002 \\
 BaseCNN + PT & 69.9 $\pm$ 0.0 & 71.8 $\pm$ 0.0 & 52.0 $\pm$ 0.3 & 54.2 $\pm$ 0.3 & \textbf{11.4 $\pm$ 0.0} & \textbf{248 $\pm$ 1} & \textbf{0.779 $\pm$ 0.003} \\ \bottomrule
\end{tabular}
\end{subtable}
\end{table}

\begin{table}[H]
 
\ContinuedFloat
\begin{subtable}{\textwidth}
\centering
\scriptsize
\subcaption{MMU=$20px$}
\begin{tabular}{lccccccc}
\toprule
\multirow{2}{*}{Method} & \multicolumn{2}{c}{Overall accuracy} & \multicolumn{2}{c}{F1 score} & \multirow{2}{*}{Patch density} & \multirow{2}{*}{Edge density} & \multirow{2}{*}{Entropy} \\
 & t=0 & t=1 & t=0 & t=1 &  &  &  \\ \midrule
 BaseMLP & 64.6 $\pm$ 0.0 & 67.2 $\pm$ 0.0 & 48.5 $\pm$ 0.2 & 51.1 $\pm$ 0.1 & 15.7 $\pm$ 0.0 & 355 $\pm$ 0 & 1.056 $\pm$ 0.001 \\
 BaseGNN & 67.6 $\pm$ 0.2 & 69.5 $\pm$ 0.2 & 53.2 $\pm$ 0.6 & 55.2 $\pm$ 0.6 & 9.9 $\pm$ 0.4 & 244 $\pm$ 8 & 0.867 $\pm$ 0.015 \\
 BaseCNN & 66.4 $\pm$ 0.2 & 68.4 $\pm$ 0.2 & 49.0 $\pm$ 0.3 & 51.0 $\pm$ 0.3 & 11.2 $\pm$ 0.0 & 273 $\pm$ 0 & 0.913 $\pm$ 0.003 \\ \midrule
 BaseMLP + PT & 69.2 $\pm$ 0.1 & 70.9 $\pm$ 0.1 & 53.3 $\pm$ 0.1 & \textbf{55.3 $\pm$ 0.1} & 8.8 $\pm$ 0.1 & 218 $\pm$ 3 & 0.787 $\pm$ 0.005 \\
 BaseGNN + PT & \textbf{69.8 $\pm$ 0.1} & \textbf{71.4 $\pm$ 0.2} & \textbf{53.5 $\pm$ 0.8} & 55.2 $\pm$ 0.8 & 8.5 $\pm$ 0.1 & 214 $\pm$ 2 & 0.776 $\pm$ 0.007 \\
 BaseCNN + PT & 69.4 $\pm$ 0.0 & 71.0 $\pm$ 0.0 & 51.3 $\pm$ 0.3 & 53.2 $\pm$ 0.3 & \textbf{8.3 $\pm$ 0.0} & \textbf{207 $\pm$ 1} & \textbf{0.740 $\pm$ 0.002} \\ \bottomrule
\end{tabular}
\end{subtable}
\end{table}

\begin{table}[H]
 
\ContinuedFloat
\begin{subtable}{\textwidth}
\centering
\scriptsize
\subcaption{MMU=$40px$}
\begin{tabular}{lccccccc}
\toprule
\multirow{2}{*}{Method} & \multicolumn{2}{c}{Overall accuracy} & \multicolumn{2}{c}{F1 score} & \multirow{2}{*}{Patch density} & \multirow{2}{*}{Edge density} & \multirow{2}{*}{Entropy} \\
 & t=0 & t=1 & t=0 & t=1 &  &  &  \\ \midrule
 BaseMLP & 64.5 $\pm$ 0.1 & 66.4 $\pm$ 0.1 & 48.3 $\pm$ 0.2 & 50.3 $\pm$ 0.2 & 9.6 $\pm$ 0.1 & 260 $\pm$ 2 & 0.970 $\pm$ 0.005 \\
 BaseGNN & 66.5 $\pm$ 0.3 & 68.0 $\pm$ 0.3 & 51.5 $\pm$ 0.9 & 53.2 $\pm$ 1.0 & 6.9 $\pm$ 0.4 & 194 $\pm$ 9 & 0.802 $\pm$ 0.030 \\
 BaseCNN & 66.0 $\pm$ 0.2 & 67.6 $\pm$ 0.2 & 48.1 $\pm$ 0.4 & 49.6 $\pm$ 0.4 & 7.5 $\pm$ 0.0 & 210 $\pm$ 0 & 0.835 $\pm$ 0.002 \\ \midrule
 BaseMLP + PT & 68.7 $\pm$ 0.1 & 70.1 $\pm$ 0.1 & 52.5 $\pm$ 0.1 & \textbf{54.1 $\pm$ 0.1} & 6.3 $\pm$ 0.0 & 177 $\pm$ 1 & 0.736 $\pm$ 0.001 \\
 BaseGNN + PT & \textbf{69.1 $\pm$ 0.0} & \textbf{70.4 $\pm$ 0.1} & \textbf{52.4 $\pm$ 0.3} & 53.9 $\pm$ 0.3 & 6.1 $\pm$ 0.1 & 172 $\pm$ 2 & 0.719 $\pm$ 0.006 \\
 BaseCNN + PT & 68.7 $\pm$ 0.0 & 69.9 $\pm$ 0.1 & 50.2 $\pm$ 0.3 & 51.7 $\pm$ 0.3 & \textbf{6.0 $\pm$ 0.0} & \textbf{167 $\pm$ 1} & \textbf{0.688 $\pm$ 0.002} \\ \bottomrule
\end{tabular}
\end{subtable}
\end{table}

\begin{table}[H]
\caption{Metrics for models with and without pre-trained features for the $1/2$ dataset, accompanies Figure 7(b)}
\begin{subtable}{\textwidth}
\centering
\scriptsize
\subcaption{MMU=$1px$}
\begin{tabular}{lccccccc}
\toprule
\multirow{2}{*}{Method} & \multicolumn{2}{c}{Overall accuracy} & \multicolumn{2}{c}{F1 score} & \multirow{2}{*}{Patch density} & \multirow{2}{*}{Edge density} & \multirow{2}{*}{Entropy} \\
 & t=0 & t=1 & t=0 & t=1 &  &  &  \\ \midrule
 BaseMLP & 59.8 $\pm$ 0.1 & 65.0 $\pm$ 0.1 & 42.8 $\pm$ 0.1 & 47.4 $\pm$ 0.1 & 118.9 $\pm$ 1.7 & 882 $\pm$ 7 & 1.157 $\pm$ 0.005 \\
 BaseGNN & 66.9 $\pm$ 0.0 & 70.7 $\pm$ 0.1 & 50.4 $\pm$ 0.1 & 54.3 $\pm$ 0.2 & 63.5 $\pm$ 2.2 & 611 $\pm$ 12 & 1.107 $\pm$ 0.006 \\
 BaseCNN & 66.1 $\pm$ 0.4 & 70.6 $\pm$ 0.6 & 50.2 $\pm$ 0.4 & 55.0 $\pm$ 1.1 & 104.3 $\pm$ 29.7 & 782 $\pm$ 133 & 1.167 $\pm$ 0.055 \\ \midrule
 BaseMLP + PT & \textbf{70.6 $\pm$ 0.0} & 72.7 $\pm$ 0.2 & \textbf{54.5 $\pm$ 0.4} & \textbf{57.1 $\pm$ 0.6} & 19.4 $\pm$ 2.8 & \textbf{314 $\pm$ 25} & \textbf{0.873 $\pm$ 0.027} \\
 BaseGNN + PT & 68.5 $\pm$ 0.1 & 70.6 $\pm$ 0.1 & 50.1 $\pm$ 0.2 & 52.6 $\pm$ 0.2 & \textbf{19.3 $\pm$ 0.1} & 319 $\pm$ 2 & 0.899 $\pm$ 0.004 \\
 BaseCNN + PT & \textbf{70.6 $\pm$ 0.1} & \textbf{72.9 $\pm$ 0.1} & 52.9 $\pm$ 0.2 & 56.0 $\pm$ 0.1 & 23.5 $\pm$ 0.3 & 362 $\pm$ 3 & 0.927 $\pm$ 0.004 \\ \bottomrule
\end{tabular}%
\end{subtable}
\end{table}

\begin{table}[H]
\caption{Metrics for models with and without pre-trained features for the $1/4$ dataset, accompanies Figure 7(c)}
\begin{subtable}{\textwidth}
\centering
\scriptsize
\subcaption{MMU=$1px$}
%
\end{subtable}
\end{table}

\begin{table}[H]
\caption{Metrics for models with and without pre-trained features for the $1/8$ dataset, accompanies Figure 7(d)}
\begin{subtable}{\textwidth}
\centering
\scriptsize
\subcaption{MMU=$1px$}
%
\end{subtable}
\end{table}

\begin{table}[H]
\caption{Metrics for models with and without pre-trained features for the $1/16$ dataset, accompanies Figure 7(e)}
\begin{subtable}{\textwidth}
\centering
\scriptsize
\subcaption{MMU=$1px$}
%
\end{subtable}
\end{table}

\section{Supplementary material for Section 3.4}
\label{app:supplementary_material_for_section_3_4}
\counterwithin{table}{section}

In this section, we provide the detailed results for all accuracy and fragmentation metrics regarding the comparison of LC-SLab to third-party land cover products from Section 3.4.

\begin{table}[H]
\caption{Metrics of GUNet and DeepLabV3 compared to ESA-WC and ESRI-LC. Accompanies Figure 8}
\begin{subtable}{\textwidth}
\centering
\scriptsize
%
\end{subtable}
\end{table}

\section{Supplementary material for Section 3.5.1}
\label{app:supplementary_material_for_section_3_5_1}
\counterwithin{table}{section}

In this section, we provide the detailed results for all accuracy and fragmentation metrics regarding the ablation studies on the object definition module from Section 3.5.1.

\begin{table}[H]
\caption{Metrics for different models and MMUs when using the FH and SLIC algorithm for the object definition, respectively. Accompanies Figure 9.}
\begin{subtable}{\textwidth}
\centering
\scriptsize
\subcaption{MMU=$5px$}
\begin{tabular}{lccccccc}
\toprule
 \multirow{2}{*}{Method} & \multicolumn{2}{c}{Overall accuracy} & \multicolumn{2}{c}{F1 score} & \multirow{2}{*}{Patch density} & \multirow{2}{*}{Edge density} & \multirow{2}{*}{Entropy} \\
 & t=0 & t=1 & t=0 & t=1 &  &  &  \\ \midrule
 BaseMLP, FH & 63.1 $\pm$ 0.3 & 67.1 $\pm$ 0.3 & 46.8 $\pm$ 0.5 & 50.5 $\pm$ 0.4 & 43.9 $\pm$ 1.9 & 599 $\pm$ 15 & 1.150 $\pm$ 0.018 \\
 BaseGNN, FH & 68.6 $\pm$ 0.1 & 71.4 $\pm$ 0.1 & 54.6 $\pm$ 0.1 & 57.6 $\pm$ 0.2 & 25.4 $\pm$ 2.2 & 403 $\pm$ 21 & 1.010 $\pm$ 0.020 \\
 GUNet, FH & 70.7 $\pm$ 0.1 & \textbf{73.1 $\pm$ 0.1} & 56.9 $\pm$ 0.3 & \textbf{59.6 $\pm$ 0.4} & 18.6 $\pm$ 0.1 & 324 $\pm$ 2 & 0.874 $\pm$ 0.006 \\ 
 BaseCNN, FH & 67.0 $\pm$ 0.2 & 69.8 $\pm$ 0.2 & 50.5 $\pm$ 0.2 & 53.4 $\pm$ 0.3 & 25.9 $\pm$ 0.4 & 420 $\pm$ 1 & 1.028 $\pm$ 0.003 \\
 UNet, FH & 69.9 $\pm$ 0.1 & 72.2 $\pm$ 0.2 & 54.1 $\pm$ 0.2 & 56.7 $\pm$ 0.3 & 23.1 $\pm$ 1.8 & 340 $\pm$ 12 & 0.881 $\pm$ 0.015 \\
 UNet++, FH & 69.6 $\pm$ 0.3 & 72.4 $\pm$ 0.3 & 54.7 $\pm$ 0.5 & 58.0 $\pm$ 0.5 & 29.0 $\pm$ 1.0 & 398 $\pm$ 1 & 0.975 $\pm$ 0.004 \\
 DeepLabV3, FH & \textbf{70.9 $\pm$ 0.4} & \textbf{73.1 $\pm$ 0.4} & \textbf{57.0 $\pm$ 0.3} & 59.5 $\pm$ 0.4 & 18.2 $\pm$ 0.2 & 319 $\pm$ 2 & \textbf{0.861 $\pm$ 0.002} \\
 Segformer, FH & 69.3 $\pm$ 0.0 & 71.6 $\pm$ 0.0 & 55.3 $\pm$ 0.3 & 57.9 $\pm$ 0.3 & 20.3 $\pm$ 0.1 & 336 $\pm$ 1 & \textbf{0.861 $\pm$ 0.002} \\ \midrule
 BaseMLP, SLIC & 62.6 $\pm$ 0.1 & 66.8 $\pm$ 0.2 & 45.6 $\pm$ 0.3 & 49.7 $\pm$ 0.5 & 44.4 $\pm$ 1.0 & 613 $\pm$ 6 & 1.176 $\pm$ 0.006 \\
 BaseGNN, SLIC & 67.2 $\pm$ 0.3 & 70.2 $\pm$ 0.4 & 51.8 $\pm$ 0.8 & 55.2 $\pm$ 1.1 & 20.3 $\pm$ 0.4 & 387 $\pm$ 5 & 1.041 $\pm$ 0.011 \\
 GUNet, SLIC & 70.0 $\pm$ 0.2 & 72.2 $\pm$ 0.2 & 55.4 $\pm$ 0.7 & 58.1 $\pm$ 0.8 & \textbf{14.2 $\pm$ 0.4} & \textbf{295 $\pm$ 7} & 0.898 $\pm$ 0.019 \\
 BaseCNN, SLIC & 66.6 $\pm$ 0.2 & 69.5 $\pm$ 0.2 & 49.7 $\pm$ 0.2 & 53.1 $\pm$ 0.3 & 22.7 $\pm$ 0.7 & 416 $\pm$ 4 & 1.055 $\pm$ 0.004 \\
 UNet, SLIC & 69.7 $\pm$ 0.1 & 72.0 $\pm$ 0.2 & 53.3 $\pm$ 0.2 & 56.5 $\pm$ 0.4 & 16.3 $\pm$ 0.8 & 311 $\pm$ 9 & 0.873 $\pm$ 0.011 \\
 UNet++, SLIC & 69.3 $\pm$ 0.2 & 72.1 $\pm$ 0.2 & 54.4 $\pm$ 0.1 & 57.9 $\pm$ 0.4 & 21.0 $\pm$ 0.6 & 373 $\pm$ 2 & 0.972 $\pm$ 0.005 \\
 DeepLabV3, SLIC & 70.7 $\pm$ 0.3 & 72.9 $\pm$ 0.4 & 56.1 $\pm$ 0.1 & 59.0 $\pm$ 0.3 & 14.7 $\pm$ 0.1 & 307 $\pm$ 2 & 0.871 $\pm$ 0.002 \\
 Segformer, SLIC & 69.1 $\pm$ 0.0 & 71.5 $\pm$ 0.0 & 54.5 $\pm$ 0.3 & 57.6 $\pm$ 0.3 & 16.3 $\pm$ 0.1 & 320 $\pm$ 1 & 0.871 $\pm$ 0.003 \\ \bottomrule
\end{tabular}
\end{subtable}
\end{table}

\begin{table}[H]
\ContinuedFloat
\begin{subtable}{\textwidth}
\centering
\scriptsize
\subcaption{MMU=$10px$}
\begin{tabular}{lccccccc}
\toprule
 \multirow{2}{*}{Method} & \multicolumn{2}{c}{Overall accuracy} & \multicolumn{2}{c}{F1 score} & \multirow{2}{*}{Patch density} & \multirow{2}{*}{Edge density} & \multirow{2}{*}{Entropy} \\
 & t=0 & t=1 & t=0 & t=1 &  &  &  \\ \midrule
 BaseMLP, FH & 63.9 $\pm$ 0.0 & 67.2 $\pm$ 0.0 & 47.0 $\pm$ 0.0 & 50.0 $\pm$ 0.1 & 24.5 $\pm$ 0.0 & 453 $\pm$ 0 & 1.086 $\pm$ 0.001 \\
 BaseGNN, FH & 68.2 $\pm$ 0.2 & 70.5 $\pm$ 0.2 & 54.0 $\pm$ 0.1 & 56.6 $\pm$ 0.1 & 14.9 $\pm$ 0.2 & 308 $\pm$ 3 & 0.930 $\pm$ 0.003 \\
 GUNet, FH & 69.4 $\pm$ 0.2 & 71.4 $\pm$ 0.2 & 56.0 $\pm$ 0.3 & 58.3 $\pm$ 0.3 & 13.0 $\pm$ 0.2 & 277 $\pm$ 3 & 0.852 $\pm$ 0.005 \\ 
 BaseCNN, FH & 66.9 $\pm$ 0.2 & 69.2 $\pm$ 0.2 & 50.0 $\pm$ 0.3 & 52.4 $\pm$ 0.3 & 16.9 $\pm$ 0.1 & 342 $\pm$ 0 & 0.977 $\pm$ 0.003 \\
 UNet, FH & 69.4 $\pm$ 0.1 & 71.5 $\pm$ 0.2 & 53.3 $\pm$ 0.3 & 55.8 $\pm$ 0.3 & 16.7 $\pm$ 1.5 & 291 $\pm$ 10 & 0.859 $\pm$ 0.018 \\
 UNet++, FH &  69.2 $\pm$ 0.3 & 71.7 $\pm$ 0.3 & 53.6 $\pm$ 0.9 & 56.7 $\pm$ 0.7 & 20.4 $\pm$ 0.8 & 333 $\pm$ 3 & 0.946 $\pm$ 0.006 \\
 DeepLabV3, FH & \textbf{70.5 $\pm$ 0.4} & \textbf{72.5 $\pm$ 0.4} & \textbf{56.4 $\pm$ 0.2} & \textbf{58.7 $\pm$ 0.3} & 12.6 $\pm$ 0.1 & 269 $\pm$ 1 & 0.828 $\pm$ 0.001 \\
 Segformer, FH & 69.0 $\pm$ 0.0 & 71.1 $\pm$ 0.1 & 54.9 $\pm$ 0.4 & 57.2 $\pm$ 0.4 & 13.9 $\pm$ 0.1 & 280 $\pm$ 0 & 0.825 $\pm$ 0.001 \\ \midrule
 BaseMLP, SLIC & 63.3 $\pm$ 0.1 & 66.5 $\pm$ 0.1 & 46.3 $\pm$ 0.1 & 49.7 $\pm$ 0.2 & 21.4 $\pm$ 0.1 & 465 $\pm$ 2 & 1.124 $\pm$ 0.003 \\
 BaseGNN, SLIC & 67.1 $\pm$ 0.3 & 69.3 $\pm$ 0.3 & 51.2 $\pm$ 0.7 & 53.9 $\pm$ 0.9 & 11.5 $\pm$ 0.1 & 303 $\pm$ 2 & 0.956 $\pm$ 0.006 \\
 GUNet, SLIC & 69.0 $\pm$ 0.1 & 71.0 $\pm$ 0.1 & 54.1 $\pm$ 0.5 & 56.6 $\pm$ 0.6 & \textbf{9.1 $\pm$ 0.1} & \textbf{252 $\pm$ 2} & 0.839 $\pm$ 0.012 \\
 BaseCNN, SLIC & 66.1 $\pm$ 0.2 & 68.5 $\pm$ 0.2 & 48.6 $\pm$ 0.3 & 51.4 $\pm$ 0.3 & 13.1 $\pm$ 0.1 & 334 $\pm$ 1 & 0.987 $\pm$ 0.003 \\
 UNet, SLIC & 68.9 $\pm$ 0.1 & 70.8 $\pm$ 0.2 & 52.1 $\pm$ 0.1 & 54.8 $\pm$ 0.2 & 11.8 $\pm$ 0.9 & 267 $\pm$ 8 & 0.839 $\pm$ 0.014 \\
 UNet++, SLIC & 68.7 $\pm$ 0.4 & 71.0 $\pm$ 0.3 & 52.9 $\pm$ 0.4 & 56.2 $\pm$ 0.4 & 14.4 $\pm$ 0.4 & 309 $\pm$ 2 & 0.926 $\pm$ 0.006 \\
 DeepLabV3, SLIC & 69.9 $\pm$ 0.4 & 71.8 $\pm$ 0.4 & 55.1 $\pm$ 0.2 & 57.7 $\pm$ 0.3 & 9.5 $\pm$ 0.1 & 257 $\pm$ 1 & 0.825 $\pm$ 0.001 \\
 Segformer, SLIC & 68.4 $\pm$ 0.1 & 70.3 $\pm$ 0.1 & 53.5 $\pm$ 0.4 & 56.1 $\pm$ 0.4 & 10.6 $\pm$ 0.1 & 267 $\pm$ 1 & \textbf{0.823 $\pm$ 0.003} \\ \bottomrule
\end{tabular}
\end{subtable}
\end{table}

\begin{table}[H]
\ContinuedFloat
\begin{subtable}{\textwidth}
\centering
\scriptsize
\subcaption{MMU=$20px$}
\begin{tabular}{lccccccc}
\toprule
 \multirow{2}{*}{Method} & \multicolumn{2}{c}{Overall accuracy} & \multicolumn{2}{c}{F1 score} & \multirow{2}{*}{Patch density} & \multirow{2}{*}{Edge density} & \multirow{2}{*}{Entropy} \\
 & t=0 & t=1 & t=0 & t=1 &  &  &  \\ \midrule
 BaseMLP, FH & 64.6 $\pm$ 0.0 & 67.2 $\pm$ 0.0 & 48.5 $\pm$ 0.2 & 51.1 $\pm$ 0.1 & 15.7 $\pm$ 0.0 & 355 $\pm$ 0 & 1.056 $\pm$ 0.001 \\
 BaseGNN, FH & 67.6 $\pm$ 0.2 & 69.5 $\pm$ 0.2 & 53.2 $\pm$ 0.6 & 55.2 $\pm$ 0.6 & 9.9 $\pm$ 0.4 & 244 $\pm$ 8 & 0.867 $\pm$ 0.015 \\
 GUNet, FH & 68.4 $\pm$ 0.1 & 70.1 $\pm$ 0.1 & 54.7 $\pm$ 0.5 & 56.6 $\pm$ 0.6 & 8.9 $\pm$ 0.0 & 225 $\pm$ 1 & 0.799 $\pm$ 0.005 \\
 BaseCNN, FH & 66.4 $\pm$ 0.2 & 68.4 $\pm$ 0.2 & 49.0 $\pm$ 0.3 & 51.0 $\pm$ 0.3 & 11.2 $\pm$ 0.0 & 273 $\pm$ 0 & 0.913 $\pm$ 0.003 \\
 UNet, FH & 69.0 $\pm$ 0.2 & 70.9 $\pm$ 0.2 & 52.8 $\pm$ 0.3 & 54.8 $\pm$ 0.3 & 11.3 $\pm$ 0.9 & 239 $\pm$ 8 & 0.819 $\pm$ 0.017 \\
 UNet++, FH & 68.7 $\pm$ 0.4 & 70.8 $\pm$ 0.4 & 52.7 $\pm$ 0.9 & 55.3 $\pm$ 0.8 & 13.3 $\pm$ 0.5 & 267 $\pm$ 2 & 0.894 $\pm$ 0.006 \\
 DeepLabV3, FH & \textbf{70.0 $\pm$ 0.4} & \textbf{71.7 $\pm$ 0.4} & \textbf{55.7 $\pm$ 0.3} & \textbf{57.5 $\pm$ 0.3} & 8.9 $\pm$ 0.0 & 222 $\pm$ 1 & 0.785 $\pm$ 0.002 \\
 Segformer, FH & 68.7 $\pm$ 0.1 & 70.4 $\pm$ 0.1 & 54.3 $\pm$ 0.4 & 56.2 $\pm$ 0.4 & 9.4 $\pm$ 0.1 & 226 $\pm$ 0 & 0.778 $\pm$ 0.001 \\ \midrule
 BaseMLP, SLIC & 63.4 $\pm$ 0.0 & 66.0 $\pm$ 0.0 & 46.3 $\pm$ 0.2 & 49.3 $\pm$ 0.2 & 13.3 $\pm$ 0.0 & 370 $\pm$ 1 & 1.072 $\pm$ 0.002 \\
BaseGNN, SLIC & 66.5 $\pm$ 0.1 & 68.5 $\pm$ 0.1 & 50.4 $\pm$ 0.1 & 53.0 $\pm$ 0.1 & 8.5 $\pm$ 0.1 & 263 $\pm$ 3 & 0.920 $\pm$ 0.005 \\
GUNet, SLIC & 67.9 $\pm$ 0.1 & 69.6 $\pm$ 0.1 & 53.0 $\pm$ 0.2 & 55.2 $\pm$ 0.3 & \textbf{6.8 $\pm$ 0.1} & 217 $\pm$ 2 & 0.797 $\pm$ 0.004 \\
BaseCNN, SLIC & 65.6 $\pm$ 0.2 & 67.7 $\pm$ 0.2 & 47.4 $\pm$ 0.2 & 49.9 $\pm$ 0.3 & 9.0 $\pm$ 0.0 & 275 $\pm$ 0 & 0.924 $\pm$ 0.003 \\
UNet, SLIC & 68.2 $\pm$ 0.1 & 70.0 $\pm$ 0.2 & 50.8 $\pm$ 0.2 & 53.3 $\pm$ 0.2 & 9.3 $\pm$ 0.9 & 230 $\pm$ 8 & 0.807 $\pm$ 0.017 \\
UNet++, SLIC & 67.9 $\pm$ 0.3 & 70.0 $\pm$ 0.3 & 51.3 $\pm$ 0.7 & 54.2 $\pm$ 0.6 & 11.2 $\pm$ 0.6 & 262 $\pm$ 3 & 0.888 $\pm$ 0.008 \\
DeepLabV3, SLIC & 69.2 $\pm$ 0.4 & 70.9 $\pm$ 0.4 & 54.0 $\pm$ 0.3 & 56.3 $\pm$ 0.3 & 6.9 $\pm$ 0.0 & \textbf{216 $\pm$ 1} & 0.779 $\pm$ 0.002 \\
Segformer, SLIC & 67.8 $\pm$ 0.0 & 69.5 $\pm$ 0.0 & 52.5 $\pm$ 0.3 & 54.8 $\pm$ 0.3 & 7.6 $\pm$ 0.1 & 223 $\pm$ 1 & \textbf{0.776 $\pm$ 0.002} \\ \bottomrule
\end{tabular}
\end{subtable}
\end{table}

\begin{table}[H]
\ContinuedFloat
\begin{subtable}{\textwidth}
\centering
\scriptsize
\subcaption{MMU=$40px$}
\begin{tabular}{lccccccc}
\toprule
\multirow{2}{*}{Method} & \multicolumn{2}{c}{Overall accuracy} & \multicolumn{2}{c}{F1 score} & \multirow{2}{*}{Patch density} & \multirow{2}{*}{Edge density} & \multirow{2}{*}{Entropy} \\
& t=0 & t=1 & t=0 & t=1 &  &  &  \\ \midrule
BaseMLP, FH & 64.5 $\pm$ 0.1 & 66.4 $\pm$ 0.1 & 48.3 $\pm$ 0.2 & 50.3 $\pm$ 0.2 & 9.6 $\pm$ 0.1 & 260 $\pm$ 2 & 0.970 $\pm$ 0.005 \\
BaseGNN, FH & 66.5 $\pm$ 0.3 & 68.0 $\pm$ 0.3 & 51.5 $\pm$ 0.9 & 53.2 $\pm$ 1.0 & 6.9 $\pm$ 0.4 & 194 $\pm$ 9 & 0.802 $\pm$ 0.030 \\
GUNet, FH & 67.0 $\pm$ 0.1 & 68.4 $\pm$ 0.0 & 52.5 $\pm$ 0.1 & 54.0 $\pm$ 0.1 & 6.4 $\pm$ 0.1 & 181 $\pm$ 4 & 0.745 $\pm$ 0.011 \\
BaseCNN, FH & 66.0 $\pm$ 0.2 & 67.6 $\pm$ 0.2 & 48.1 $\pm$ 0.4 & 49.6 $\pm$ 0.4 & 7.5 $\pm$ 0.0 & 210 $\pm$ 0 & 0.835 $\pm$ 0.002 \\
UNet, FH & 68.5 $\pm$ 0.1 & 70.0 $\pm$ 0.2 & 52.0 $\pm$ 0.3 & 53.5 $\pm$ 0.4 & 7.3 $\pm$ 0.4 & 187 $\pm$ 5 & 0.757 $\pm$ 0.014 \\
UNet++, FH & 68.2 $\pm$ 0.4 & 69.8 $\pm$ 0.3 & 52.1 $\pm$ 0.7 & 54.0 $\pm$ 0.6 & 8.2 $\pm$ 0.2 & 205 $\pm$ 1 & 0.815 $\pm$ 0.005 \\
DeepLabV3, FH & \textbf{69.3 $\pm$ 0.4} & \textbf{70.7 $\pm$ 0.4} & \textbf{54.5 $\pm$ 0.3} & \textbf{56.1 $\pm$ 0.3} & 6.3 $\pm$ 0.0 & 177 $\pm$ 0 & 0.728 $\pm$ 0.002 \\
Segformer, FH & 68.1 $\pm$ 0.0 & 69.5 $\pm$ 0.0 & 53.4 $\pm$ 0.4 & 54.9 $\pm$ 0.4 & 6.4 $\pm$ 0.1 & 177 $\pm$ 0 & 0.718 $\pm$ 0.001 \\ \midrule
BaseMLP, SLIC & 63.0 $\pm$ 0.0 & 65.0 $\pm$ 0.1 & 45.7 $\pm$ 0.2 & 48.0 $\pm$ 0.2 & 8.3 $\pm$ 0.2 & 282 $\pm$ 5 & 0.987 $\pm$ 0.016 \\
BaseGNN, SLIC & 65.8 $\pm$ 0.2 & 67.3 $\pm$ 0.1 & 49.8 $\pm$ 0.2 & 51.7 $\pm$ 0.3 & 5.5 $\pm$ 0.3 & 197 $\pm$ 9 & 0.817 $\pm$ 0.023 \\
GUNet, SLIC & 66.6 $\pm$ 0.1 & 67.9 $\pm$ 0.1 & 51.1 $\pm$ 0.3 & 52.9 $\pm$ 0.3 & \textbf{4.9 $\pm$ 0.0} & \textbf{175 $\pm$ 1} & 0.723 $\pm$ 0.004 \\
BaseCNN, SLIC & 64.8 $\pm$ 0.2 & 66.4 $\pm$ 0.2 & 45.9 $\pm$ 0.3 & 47.7 $\pm$ 0.3 & 6.1 $\pm$ 0.0 & 218 $\pm$ 0 & 0.840 $\pm$ 0.002 \\
UNet, SLIC & 67.4 $\pm$ 0.1 & 68.8 $\pm$ 0.2 & 49.5 $\pm$ 0.2 & 51.5 $\pm$ 0.3 & 6.8 $\pm$ 0.7 & 192 $\pm$ 7 & 0.754 $\pm$ 0.019 \\
UNet++, SLIC & 67.1 $\pm$ 0.4 & 68.7 $\pm$ 0.4 & 49.9 $\pm$ 0.7 & 52.2 $\pm$ 0.7 & 7.9 $\pm$ 0.5 & 214 $\pm$ 3 & 0.824 $\pm$ 0.008 \\
DeepLabV3, SLIC & 68.4 $\pm$ 0.4 & 69.8 $\pm$ 0.4 & 52.7 $\pm$ 0.2 & 54.5 $\pm$ 0.3 & 5.0 $\pm$ 0.0 & 176 $\pm$ 1 & 0.716 $\pm$ 0.003 \\
Segformer, SLIC & 67.1 $\pm$ 0.1 & 68.4 $\pm$ 0.1 & 51.4 $\pm$ 0.3 & 53.3 $\pm$ 0.3 & 5.4 $\pm$ 0.1 & 181 $\pm$ 0 & \textbf{0.715 $\pm$ 0.000} \\ \bottomrule
\end{tabular}
\end{subtable}
\end{table}

\section{Supplementary material for Section 3.5.2}
\label{app:supplementary_material_for_section_3_5_2}
\counterwithin{table}{section}

In this section, we provide the detailed results for all accuracy and fragmentation metrics regarding the ablation studies on graph convolution operators in input-level aggregation methods from Section 3.5.2.

\begin{table}[H]
\caption{Metrics for different graph convolution operators used in BaseGNN and GUNet. Accompanies Figure 10}
\begin{subtable}{\textwidth}
\centering
\scriptsize
\subcaption{MMU=$1px$}
\begin{tabular}{lccccccc}
\toprule
\multirow{2}{*}{Method} & \multicolumn{2}{c}{Overall accuracy} & \multicolumn{2}{c}{F1 score} & \multirow{2}{*}{Patch density} & \multirow{2}{*}{Edge density} & \multirow{2}{*}{Entropy} \\
& t=0 & t=1 & t=0 & t=1 &  &  &  \\ \midrule
BaseGNN, GCN & 64.5 $\pm$ 0.1 & 67.2 $\pm$ 0.2 & 47.7 $\pm$ 0.3 & 50.5 $\pm$ 0.3 & 23.7 $\pm$ 0.7 & 388 $\pm$ 8 & 1.063 $\pm$ 0.013 \\
BaseGNN, GraphSAGE & 66.6 $\pm$ 0.0 & 70.3 $\pm$ 0.1 & 50.7 $\pm$ 0.0 & 54.7 $\pm$ 0.1 & 64.5 $\pm$ 0.4 & 613 $\pm$ 1 & 1.116 $\pm$ 0.001 \\
BaseGNN, GAT & 67.6 $\pm$ 0.2 & 70.8 $\pm$ 0.4 & 52.0 $\pm$ 0.6 & 55.3 $\pm$ 0.8 & 33.5 $\pm$ 4.1 & 470 $\pm$ 35 & 1.087 $\pm$ 0.017 \\
BaseGNN, GT & 67.5 $\pm$ 0.0 & 71.8 $\pm$ 0.1 & 52.1 $\pm$ 0.1 & 56.4 $\pm$ 0.1 & 74.4 $\pm$ 4.5 & 665 $\pm$ 20 & 1.124 $\pm$ 0.006 \\ \midrule
GUNet, GCN & 66.0 $\pm$ 0.1 & 68.1 $\pm$ 0.1 & 49.8 $\pm$ 0.1 & 52.1 $\pm$ 0.2 & 14.9 $\pm$ 0.3 & 295 $\pm$ 4 & 1.014 $\pm$ 0.008 \\
GUNet, GraphSAGE & 69.9 $\pm$ 0.5 & 72.5 $\pm$ 0.6 & 55.6 $\pm$ 0.9 & 58.6 $\pm$ 0.8 & 31.3 $\pm$ 1.7 & 402 $\pm$ 11 & 1.015 $\pm$ 0.002 \\
GUNet, GAT & 69.1 $\pm$ 0.2 & 71.0 $\pm$ 0.1 & 54.3 $\pm$ 0.3 & 56.2 $\pm$ 0.3 & \textbf{12.7 $\pm$ 0.0} & \textbf{259 $\pm$ 0} & \textbf{0.944 $\pm$ 0.002} \\
GUNet, GT & \textbf{71.0 $\pm$ 0.1} & \textbf{73.5 $\pm$ 0.1} & \textbf{56.9 $\pm$ 0.1} & \textbf{59.9 $\pm$ 0.1} & 32.3 $\pm$ 0.5 & 400 $\pm$ 1 & 0.997 $\pm$ 0.003 \\ \bottomrule
\end{tabular}
\end{subtable}
\end{table}

\begin{table}[H]
\ContinuedFloat
\begin{subtable}{\textwidth}
\centering
\scriptsize
\subcaption{MMU=$5px$}
\begin{tabular}{lccccccc}
\toprule
 \multirow{2}{*}{Method} & \multicolumn{2}{c}{Overall accuracy} & \multicolumn{2}{c}{F1 score} & \multirow{2}{*}{Patch density} & \multirow{2}{*}{Edge density} & \multirow{2}{*}{Entropy} \\
 & t=0 & t=1 & t=0 & t=1 &  &  &  \\ \midrule
 BaseGNN, GCN & 66.2 $\pm$ 0.0 & 68.2 $\pm$ 0.1 & 49.9 $\pm$ 0.1 & 52.1 $\pm$ 0.1 & 13.5 $\pm$ 0.1 & 251 $\pm$ 1 & 0.871 $\pm$ 0.002 \\
 BaseGNN, GraphSAGE & 68.2 $\pm$ 0.1 & 71.4 $\pm$ 0.1 & 53.7 $\pm$ 0.1 & 57.1 $\pm$ 0.2 & 29.8 $\pm$ 0.6 & 444 $\pm$ 4 & 1.023 $\pm$ 0.003 \\
 BaseGNN, GAT & 67.8 $\pm$ 0.2 & 69.8 $\pm$ 0.2 & 52.5 $\pm$ 0.0 & 54.8 $\pm$ 0.1 & 14.9 $\pm$ 0.4 & 278 $\pm$ 5 & 0.918 $\pm$ 0.003 \\
 BaseGNN, GT & 68.6 $\pm$ 0.1 & 71.4 $\pm$ 0.1 & 54.6 $\pm$ 0.1 & 57.6 $\pm$ 0.2 & 25.4 $\pm$ 2.2 & 403 $\pm$ 21 & 1.010 $\pm$ 0.020 \\ \midrule
 GUNet, GCN & 65.8 $\pm$ 0.3 & 67.5 $\pm$ 0.4 & 50.3 $\pm$ 0.9 & 52.2 $\pm$ 1.0 & 11.2 $\pm$ 0.5 & 215 $\pm$ 8 & 0.809 $\pm$ 0.017 \\
 GUNet, GraphSAGE & 69.2 $\pm$ 0.1 & 71.6 $\pm$ 0.2 & 55.2 $\pm$ 0.2 & 57.7 $\pm$ 0.3 & 18.8 $\pm$ 0.3 & 326 $\pm$ 3 & 0.890 $\pm$ 0.005 \\
 GUNet, GAT & 67.1 $\pm$ 0.2 & 68.7 $\pm$ 0.3 & 52.4 $\pm$ 0.4 & 54.1 $\pm$ 0.5 & \textbf{10.1 $\pm$ 0.0} & \textbf{206 $\pm$ 1} & \textbf{0.801 $\pm$ 0.000} \\
 GUNet, GT & \textbf{70.7 $\pm$ 0.1} & \textbf{73.1 $\pm$ 0.1} & \textbf{56.9 $\pm$ 0.3} & \textbf{59.6 $\pm$ 0.4} & 18.6 $\pm$ 0.1 & 324 $\pm$ 2 & 0.874 $\pm$ 0.006 \\ \bottomrule
\end{tabular}
\end{subtable}
\end{table}

\begin{table}[H]
\ContinuedFloat
\begin{subtable}{\textwidth}
\centering
\scriptsize
\subcaption{MMU=$10px$}
\begin{tabular}{lccccccc}
\toprule
 \multirow{2}{*}{Method} & \multicolumn{2}{c}{Overall accuracy} & \multicolumn{2}{c}{F1 score} & \multirow{2}{*}{Patch density} & \multirow{2}{*}{Edge density} & \multirow{2}{*}{Entropy} \\
 & t=0 & t=1 & t=0 & t=1 &  &  &  \\ \midrule
 BaseGNN, GCN & 65.4 $\pm$ 0.1 & 66.8 $\pm$ 0.2 & 48.7 $\pm$ 0.1 & 50.3 $\pm$ 0.2 & 8.3 $\pm$ 0.2 & 183 $\pm$ 4 & 0.756 $\pm$ 0.007 \\
 BaseGNN, GraphSAGE & 67.8 $\pm$ 0.0 & 70.5 $\pm$ 0.0 & 53.2 $\pm$ 0.1 & 56.0 $\pm$ 0.1 & 17.8 $\pm$ 0.3 & 346 $\pm$ 4 & 0.967 $\pm$ 0.003 \\
 BaseGNN, GAT & 66.7 $\pm$ 0.2 & 68.3 $\pm$ 0.2 & 50.6 $\pm$ 0.3 & 52.5 $\pm$ 0.3 & 9.8 $\pm$ 0.0 & 218 $\pm$ 1 & 0.834 $\pm$ 0.005 \\
 BaseGNN, GT & 68.2 $\pm$ 0.2 & 70.5 $\pm$ 0.2 & 54.0 $\pm$ 0.1 & 56.6 $\pm$ 0.1 & 14.9 $\pm$ 0.2 & 308 $\pm$ 3 & 0.930 $\pm$ 0.003 \\ \midrule
 GUNet, GCN & 64.6 $\pm$ 0.2 & 65.9 $\pm$ 0.2 & 48.9 $\pm$ 0.1 & 50.3 $\pm$ 0.2 & 7.7 $\pm$ 0.1 & 171 $\pm$ 2 & 0.719 $\pm$ 0.005 \\
 GUNet, GraphSAGE & 68.8 $\pm$ 0.1 & 70.9 $\pm$ 0.1 & 55.0 $\pm$ 0.2 & 57.3 $\pm$ 0.2 & 12.8 $\pm$ 0.1 & 275 $\pm$ 2 & 0.848 $\pm$ 0.005 \\
 GUNet, GAT & 65.9 $\pm$ 0.2 & 67.2 $\pm$ 0.2 & 50.5 $\pm$ 0.5 & 51.9 $\pm$ 0.5 & \textbf{6.9 $\pm$ 0.0} & \textbf{160 $\pm$ 2} & \textbf{0.715 $\pm$ 0.004} \\
 GUNet, GT & \textbf{69.4 $\pm$ 0.2} & \textbf{71.4 $\pm$ 0.2} & \textbf{56.0 $\pm$ 0.3} & \textbf{58.3 $\pm$ 0.3} & 13.0 $\pm$ 0.2 & 277 $\pm$ 3 & 0.852 $\pm$ 0.005 \\ \bottomrule
\end{tabular}
\end{subtable}
\end{table}

\begin{table}[H]
 
\ContinuedFloat
\begin{subtable}{\textwidth}
\centering
\scriptsize
\subcaption{MMU=$20px$}
\begin{tabular}{lccccccc}
\toprule
 \multirow{2}{*}{Method} & \multicolumn{2}{c}{Overall accuracy} & \multicolumn{2}{c}{F1 score} & \multirow{2}{*}{Patch density} & \multirow{2}{*}{Edge density} & \multirow{2}{*}{Entropy} \\
 & t=0 & t=1 & t=0 & t=1 &  &  &  \\ \midrule
 BaseGNN, GCN & 64.0 $\pm$ 0.1 & 64.9 $\pm$ 0.1 & 46.7 $\pm$ 0.2 & 47.8 $\pm$ 0.1 & 5.2 $\pm$ 0.0 & 126 $\pm$ 1 & 0.618 $\pm$ 0.003 \\
 BaseGNN, GraphSAGE & 67.8 $\pm$ 0.1 & 69.8 $\pm$ 0.1 & 52.9 $\pm$ 0.2 & 55.1 $\pm$ 0.3 & 10.7 $\pm$ 0.2 & 260 $\pm$ 4 & 0.887 $\pm$ 0.007 \\
 BaseGNN, GAT & 65.4 $\pm$ 0.3 & 66.7 $\pm$ 0.4 & 48.3 $\pm$ 0.4 & 49.8 $\pm$ 0.5 & 6.3 $\pm$ 0.2 & 160 $\pm$ 5 & 0.721 $\pm$ 0.014 \\
 BaseGNN, GT & 67.6 $\pm$ 0.2 & 69.5 $\pm$ 0.2 & 53.2 $\pm$ 0.6 & 55.2 $\pm$ 0.6 & 9.9 $\pm$ 0.4 & 244 $\pm$ 8 & 0.867 $\pm$ 0.015 \\ \midrule
 GUNet, GCN & 62.8 $\pm$ 0.1 & 63.7 $\pm$ 0.2 & 46.4 $\pm$ 0.3 & 47.5 $\pm$ 0.4 & 5.0 $\pm$ 0.1 & 122 $\pm$ 1 & \textbf{0.604 $\pm$ 0.004} \\
 GUNet, GraphSAGE & 67.4 $\pm$ 0.1 & 69.1 $\pm$ 0.1 & 52.6 $\pm$ 0.3 & 54.4 $\pm$ 0.2 & 8.8 $\pm$ 0.2 & 222 $\pm$ 5 & 0.784 $\pm$ 0.013 \\
 GUNet, GAT & 64.2 $\pm$ 0.3 & 65.2 $\pm$ 0.3 & 48.3 $\pm$ 0.7 & 49.4 $\pm$ 0.7 & \textbf{4.8 $\pm$ 0.1} & \textbf{122 $\pm$ 4} & 0.628 $\pm$ 0.011 \\
 GUNet, GT & \textbf{68.4 $\pm$ 0.1} & \textbf{70.1 $\pm$ 0.1} & \textbf{54.7 $\pm$ 0.5} & \textbf{56.6 $\pm$ 0.6} & 8.9 $\pm$ 0.0 & 225 $\pm$ 1 & 0.799 $\pm$ 0.005 \\ \bottomrule
\end{tabular}
\end{subtable}
\end{table}

\begin{table}[H]
\ContinuedFloat
\begin{subtable}{\textwidth}
\centering
\scriptsize
\subcaption{MMU=$40px$}
\begin{tabular}{lccccccc}
\toprule
 \multirow{2}{*}{Method} & \multicolumn{2}{c}{Overall accuracy} & \multicolumn{2}{c}{F1 score} & \multirow{2}{*}{Patch density} & \multirow{2}{*}{Edge density} & \multirow{2}{*}{Entropy} \\
 & t=0 & t=1 & t=0 & t=1 &  &  &  \\ \midrule
 BaseGNN, GCN & 62.4 $\pm$ 0.5 & 63.1 $\pm$ 0.5 & 44.5 $\pm$ 1.0 & 45.4 $\pm$ 1.1 & 3.5 $\pm$ 0.2 & 87 $\pm$ 6 & \textbf{0.480 $\pm$ 0.026} \\
 BaseGNN, GraphSAGE & 66.6 $\pm$ 0.2 & 68.1 $\pm$ 0.2 & 51.7 $\pm$ 0.5 & 53.4 $\pm$ 0.6 & 7.3 $\pm$ 0.2 & 203 $\pm$ 4 & 0.824 $\pm$ 0.009 \\
 BaseGNN, GAT & 63.9 $\pm$ 0.1 & 65.0 $\pm$ 0.0 & 47.1 $\pm$ 0.1 & 48.3 $\pm$ 0.1 & 4.8 $\pm$ 0.0 & 131 $\pm$ 2 & 0.656 $\pm$ 0.005 \\
 BaseGNN, GT & 66.5 $\pm$ 0.3 & 68.0 $\pm$ 0.3 & 51.5 $\pm$ 0.9 & 53.2 $\pm$ 1.0 & 6.9 $\pm$ 0.4 & 194 $\pm$ 9 & 0.802 $\pm$ 0.030 \\ \midrule
 GUNet, GCN & 60.5 $\pm$ 0.3 & 61.2 $\pm$ 0.3 & 42.6 $\pm$ 0.6 & 43.3 $\pm$ 0.7 & 3.5 $\pm$ 0.1 & \textbf{86 $\pm$ 2} & 0.489 $\pm$ 0.005 \\
 GUNet, GraphSAGE & 66.3 $\pm$ 0.6 & 67.7 $\pm$ 0.6 & 51.2 $\pm$ 1.2 & 52.6 $\pm$ 1.3 & 6.3 $\pm$ 0.1 & 179 $\pm$ 2 & 0.730 $\pm$ 0.009 \\
 GUNet, GAT & 62.7 $\pm$ 0.1 & 63.4 $\pm$ 0.1 & 46.4 $\pm$ 0.1 & 47.3 $\pm$ 0.1 & \textbf{3.4 $\pm$ 0.0} & 89 $\pm$ 0 & 0.523 $\pm$ 0.003 \\
 GUNet, GT & \textbf{67.0 $\pm$ 0.1} & \textbf{68.4 $\pm$ 0.0} & \textbf{52.5 $\pm$ 0.1} & \textbf{54.0 $\pm$ 0.1} & 6.4 $\pm$ 0.1 & 181 $\pm$ 4 & 0.745 $\pm$ 0.011 \\ \bottomrule
\end{tabular}
\end{subtable}
\end{table}

\section{Supplementary material for Section 3.5.3}
\label{app:supplementary_material_for_section_3_5_3}
\counterwithin{table}{section}

In this section, we provide the detailed results for all accuracy and fragmentation metrics regarding the ablation studies on the features used in input-level aggregation methods from Section 3.5.3.

\begin{table}[H]
\caption{Metrics for different input-level object aggregation approaches when using different sets of input features. Accompanies Figure 11.}
\begin{subtable}{\textwidth}
\centering
\scriptsize
\subcaption{MMU=$5px$}

\end{subtable}
\end{table}

\begin{table}[H]
\ContinuedFloat
\begin{subtable}{\textwidth}
\centering
\scriptsize
\subcaption{MMU=$10px$}
%
\end{subtable}
\end{table}

\begin{table}[H]
 
\ContinuedFloat
\begin{subtable}{\textwidth}
\centering
\scriptsize
\subcaption{MMU=$20px$}
%
\end{subtable}
\end{table}

\begin{table}[H]
\ContinuedFloat
\begin{subtable}{\textwidth}
\centering
\scriptsize
\subcaption{MMU=$40px$}
%
\end{subtable}
\end{table}

\section{Supplementary material for Section 3.5.4}
\label{app:supplementary_material_for_section_3_5_4}
\counterwithin{table}{section}

In this section, we provide the detailed results for all accuracy and fragmentation metrics regarding the ablation studies on the backbones of the segmentation models used in output-level aggregation methods from Section 3.5.4.

\begin{table}[H]
\caption{Output-level object aggregation, accompanies Figure 7}
\begin{subtable}{\textwidth}
\centering
\scriptsize
\subcaption{MMU=$1px$}
%
\end{subtable}
\end{table}

\section{Supplementary material for Section 3.5.5}
\label{app:supplementary_material_for_section_3_5_5}
\counterwithin{table}{section}

In this section, we provide the detailed results for all accuracy and fragmentation metrics regarding the ablation studies on rescaling the input images from $64 \times 64$ to $224 \times 224$ pixels in output-level aggregation methods from Section 3.5.5.

\begin{table}[H]
\caption{Output-level object aggregation, accompanies Figure 7}
\begin{subtable}{\textwidth}
\centering
\scriptsize
\subcaption{MMU=$1px$}
\begin{tabular}{lccccccc}
\toprule
 \multirow{2}{*}{Method} & \multicolumn{2}{c}{Overall accuracy} & \multicolumn{2}{c}{F1 score} & \multirow{2}{*}{Patch density} & \multirow{2}{*}{Edge density} & \multirow{2}{*}{Entropy} \\
 & t=0 & t=1 & t=0 & t=1 &  &  &  \\ \midrule
 UNet & 70.5 $\pm$ 0.1 & 72.9 $\pm$ 0.2 & 54.8 $\pm$ 0.2 & 57.9 $\pm$ 0.3 & 34.1 $\pm$ 2.7 & 411 $\pm$ 17 & 0.919 $\pm$ 0.011 \\ 
 UNet++ & 69.8 $\pm$ 0.1 & 73.1 $\pm$ 0.2 & 55.6 $\pm$ 0.1 & 59.7 $\pm$ 0.3 & 48.2 $\pm$ 4.0 & 523 $\pm$ 15 & 1.039 $\pm$ 0.003 \\ 
 DeepLabV3 & \textbf{71.3 $\pm$ 0.4} & \textbf{73.8 $\pm$ 0.4} & \textbf{57.4 $\pm$ 0.2} & \textbf{60.5 $\pm$ 0.3} & 29.2 $\pm$ 0.7 & 396 $\pm$ 4 & 0.917 $\pm$ 0.003 \\ 
 Segformer & 69.7 $\pm$ 0.0 & 72.3 $\pm$ 0.0 & 55.7 $\pm$ 0.3 & 58.8 $\pm$ 0.4 & 33.4 $\pm$ 0.3 & 419 $\pm$ 2 & 0.920 $\pm$ 0.004 \\ \midrule
 UNet, no rescaling & 67.2 $\pm$ 0.2 & 69.6 $\pm$ 0.5 & 50.3 $\pm$ 2.3 & 53.0 $\pm$ 2.6 & 25.4 $\pm$ 3.5 & 364 $\pm$ 31 & 0.913 $\pm$ 0.044 \\
 UNet++, no rescaling & 66.9 $\pm$ 0.1 & 69.5 $\pm$ 0.3 & 50.3 $\pm$ 1.2 & 53.0 $\pm$ 1.5 & 26.5 $\pm$ 4.8 & 383 $\pm$ 29 & 0.963 $\pm$ 0.016 \\ 
 DeepLabV3, no rescaling & 66.2 $\pm$ 0.9 & 68.0 $\pm$ 1.0 & 50.4 $\pm$ 1.4 & 52.8 $\pm$ 1.6 & \textbf{11.7 $\pm$ 1.3} & \textbf{233 $\pm$ 19} & \textbf{0.820 $\pm$ 0.041} \\
 Segformer, no rescaling & 63.2 $\pm$ 0.2 & 65.2 $\pm$ 0.2 & 46.6 $\pm$ 1.0 & 49.0 $\pm$ 1.0 & 13.1 $\pm$ 0.2 & 246 $\pm$ 1 & 0.902 $\pm$ 0.011 \\ \bottomrule
\end{tabular}
\end{subtable}
\end{table}

\begin{table}[H]
 
\ContinuedFloat
\begin{subtable}{\textwidth}
\centering
\scriptsize
\subcaption{MMU=$5px$}
\begin{tabular}{lccccccc}
\toprule
 \multirow{2}{*}{Method} & \multicolumn{2}{c}{Overall accuracy} & \multicolumn{2}{c}{F1 score} & \multirow{2}{*}{Patch density} & \multirow{2}{*}{Edge density} & \multirow{2}{*}{Entropy} \\
 & t=0 & t=1 & t=0 & t=1 &  &  &  \\ \midrule
 UNet & 69.9 $\pm$ 0.1 & 72.2 $\pm$ 0.2 & 54.1 $\pm$ 0.2 & 56.7 $\pm$ 0.3 & 23.1 $\pm$ 1.8 & 340 $\pm$ 12 & 0.881 $\pm$ 0.015 \\
 UNet++ &  69.6 $\pm$ 0.3 & 72.4 $\pm$ 0.3 & 54.7 $\pm$ 0.5 & 58.0 $\pm$ 0.5 & 29.0 $\pm$ 1.0 & 398 $\pm$ 1 & 0.975 $\pm$ 0.004 \\
 DeepLabV3 & \textbf{70.9 $\pm$ 0.4} & \textbf{73.1 $\pm$ 0.4} & \textbf{57.0 $\pm$ 0.3} & \textbf{59.5 $\pm$ 0.4} & 18.2 $\pm$ 0.2 & 319 $\pm$ 2 & 0.861 $\pm$ 0.002 \\
 Segformer & 69.3 $\pm$ 0.0 & 71.6 $\pm$ 0.0 & 55.3 $\pm$ 0.3 & 57.9 $\pm$ 0.3 & 20.3 $\pm$ 0.1 & 336 $\pm$ 1 & 0.861 $\pm$ 0.002 \\ \midrule
 UNet, no rescaling & 67.0 $\pm$ 0.3 & 69.2 $\pm$ 0.5 & 50.2 $\pm$ 2.5 & 52.6 $\pm$ 2.9 & 19.2 $\pm$ 1.8 & 318 $\pm$ 21 & 0.863 $\pm$ 0.037 \\
 UNet++, no rescaling & 66.8 $\pm$ 0.2 & 69.2 $\pm$ 0.2 & 50.1 $\pm$ 1.1 & 52.6 $\pm$ 1.2 & 21.0 $\pm$ 2.0 & 339 $\pm$ 11 & 0.912 $\pm$ 0.009 \\ 
 DeepLabV3, no rescaling & 66.3 $\pm$ 1.0 & 68.1 $\pm$ 1.1 & 50.5 $\pm$ 1.5 & 52.8 $\pm$ 1.7 & \textbf{14.4 $\pm$ 1.1} & \textbf{255 $\pm$ 17} & \textbf{0.785 $\pm$ 0.039} \\
 Segformer, no rescaling & 63.8 $\pm$ 0.2 & 65.7 $\pm$ 0.2 & 47.2 $\pm$ 1.0 & 49.2 $\pm$ 1.1 & 17.4 $\pm$ 0.6 & 293 $\pm$ 3 & 0.832 $\pm$ 0.013 \\\bottomrule
\end{tabular}
\end{subtable}
\end{table}

\begin{table}[H]
 
\ContinuedFloat
\begin{subtable}{\textwidth}
\centering
\scriptsize
\subcaption{MMU=$10px$}
\begin{tabular}{lccccccc}
\toprule
 \multirow{2}{*}{Method} & \multicolumn{2}{c}{Overall accuracy} & \multicolumn{2}{c}{F1 score} & \multirow{2}{*}{Patch density} & \multirow{2}{*}{Edge density} & \multirow{2}{*}{Entropy} \\
 & t=0 & t=1 & t=0 & t=1 &  &  &  \\ \midrule
 UNet & 69.4 $\pm$ 0.1 & 71.5 $\pm$ 0.2 & 53.3 $\pm$ 0.3 & 55.8 $\pm$ 0.3 & 16.7 $\pm$ 1.5 & 291 $\pm$ 10 & 0.859 $\pm$ 0.018 \\ 
 UNet++ & 69.2 $\pm$ 0.3 & 71.7 $\pm$ 0.3 & 53.6 $\pm$ 0.9 & 56.7 $\pm$ 0.7 & 20.4 $\pm$ 0.8 & 333 $\pm$ 3 & 0.946 $\pm$ 0.006 \\
 DeepLabV3 & \textbf{70.5 $\pm$ 0.4} & \textbf{72.5 $\pm$ 0.4} & \textbf{56.4 $\pm$ 0.2} & \textbf{58.7 $\pm$ 0.3} & 12.6 $\pm$ 0.1 & 269 $\pm$ 1 & 0.828 $\pm$ 0.001 \\ 
 Segformer & 69.0 $\pm$ 0.0 & 71.1 $\pm$ 0.1 & 54.9 $\pm$ 0.4 & 57.2 $\pm$ 0.4 & 13.9 $\pm$ 0.1 & 280 $\pm$ 0 & 0.825 $\pm$ 0.001 \\ \midrule
 UNet, no rescaling & 66.8 $\pm$ 0.3 & 68.7 $\pm$ 0.5 & 49.8 $\pm$ 2.5 & 51.9 $\pm$ 2.8 & 13.6 $\pm$ 1.1 & 270 $\pm$ 15 & 0.830 $\pm$ 0.033 \\
 UNet++, no rescaling & 66.6 $\pm$ 0.3 & 68.6 $\pm$ 0.2 & 49.7 $\pm$ 1.1 & 51.9 $\pm$ 1.1 & 15.0 $\pm$ 1.1 & 289 $\pm$ 4 & 0.880 $\pm$ 0.007 \\ 
 DeepLabV3, no rescaling & 66.2 $\pm$ 1.0 & 67.9 $\pm$ 1.1 & 50.2 $\pm$ 1.4 & 52.2 $\pm$ 1.6 & \textbf{10.8 $\pm$ 0.7} & \textbf{224 $\pm$ 14} & \textbf{0.759 $\pm$ 0.038} \\
 Segformer, no rescaling & 63.7 $\pm$ 0.2 & 65.4 $\pm$ 0.2 & 47.1 $\pm$ 0.9 & 48.9 $\pm$ 0.9 & 12.4 $\pm$ 0.5 & 251 $\pm$ 5 & 0.803 $\pm$ 0.015 \\ \bottomrule
\end{tabular}
\end{subtable}
\end{table}

\begin{table}[H]
 
\ContinuedFloat
\begin{subtable}{\textwidth}
\centering
\scriptsize
\subcaption{MMU=$20px$}
\begin{tabular}{lccccccc}
\toprule
 \multirow{2}{*}{Method} & \multicolumn{2}{c}{Overall accuracy} & \multicolumn{2}{c}{F1 score} & \multirow{2}{*}{Patch density} & \multirow{2}{*}{Edge density} & \multirow{2}{*}{Entropy} \\
 & t=0 & t=1 & t=0 & t=1 &  &  &  \\ \midrule
 UNet & 69.0 $\pm$ 0.2 & 70.9 $\pm$ 0.2 & 52.8 $\pm$ 0.3 & 54.8 $\pm$ 0.3 & 11.3 $\pm$ 0.9 & 239 $\pm$ 8 & 0.819 $\pm$ 0.017 \\
 UNet++ & 68.7 $\pm$ 0.4 & 70.8 $\pm$ 0.4 & 52.7 $\pm$ 0.9 & 55.3 $\pm$ 0.8 & 13.3 $\pm$ 0.5 & 267 $\pm$ 2 & 0.894 $\pm$ 0.006 \\
 DeepLabV3 & \textbf{70.0 $\pm$ 0.4} & \textbf{71.7 $\pm$ 0.4} & \textbf{55.7 $\pm$ 0.3} & \textbf{57.5 $\pm$ 0.3} & 8.9 $\pm$ 0.0 & 222 $\pm$ 1 & 0.785 $\pm$ 0.002 \\ 
 Segformer & 68.7 $\pm$ 0.1 & 70.4 $\pm$ 0.1 & 54.3 $\pm$ 0.4 & 56.2 $\pm$ 0.4 & 9.4 $\pm$ 0.1 & 226 $\pm$ 0 & 0.778 $\pm$ 0.001 \\ \midrule
 UNet, no rescaling & 66.6 $\pm$ 0.4 & 68.2 $\pm$ 0.6 & 49.4 $\pm$ 2.5 & 51.2 $\pm$ 2.7 & 9.4 $\pm$ 0.6 & 222 $\pm$ 11 & 0.788 $\pm$ 0.029 \\
 UNet++, no rescaling & 66.3 $\pm$ 0.3 & 68.1 $\pm$ 0.3 & 49.3 $\pm$ 1.0 & 51.1 $\pm$ 1.1 & 10.3 $\pm$ 0.5 & 238 $\pm$ 3 & 0.835 $\pm$ 0.007 \\ 
 DeepLabV3, no rescaling & 66.0 $\pm$ 1.0 & 67.5 $\pm$ 1.1 & 49.9 $\pm$ 1.6 & 51.6 $\pm$ 1.7 & \textbf{7.8 $\pm$ 0.5} & \textbf{191 $\pm$ 12} & \textbf{0.725 $\pm$ 0.037} \\
 Segformer, no rescaling & 63.5 $\pm$ 0.2 & 65.0 $\pm$ 0.2 & 46.8 $\pm$ 0.9 & 48.3 $\pm$ 0.9 & 8.9 $\pm$ 0.3 & 210 $\pm$ 5 & 0.766 $\pm$ 0.018 \\ \bottomrule
\end{tabular}
\end{subtable}
\end{table}

\begin{table}[H]
 
\ContinuedFloat
\begin{subtable}{\textwidth}
\centering
\scriptsize
\subcaption{MMU=$40px$}
\begin{tabular}{lccccccc}
\toprule
 \multirow{2}{*}{Method} & \multicolumn{2}{c}{Overall accuracy} & \multicolumn{2}{c}{F1 score} & \multirow{2}{*}{Patch density} & \multirow{2}{*}{Edge density} & \multirow{2}{*}{Entropy} \\
 & t=0 & t=1 & t=0 & t=1 &  &  &  \\ \midrule
 UNet & 68.5 $\pm$ 0.1 & 70.0 $\pm$ 0.2 & 52.0 $\pm$ 0.3 & 53.5 $\pm$ 0.4 & 7.3 $\pm$ 0.4 & 187 $\pm$ 5 & 0.757 $\pm$ 0.014 \\
 UNet++ & 68.2 $\pm$ 0.4 & 69.8 $\pm$ 0.3 & 52.1 $\pm$ 0.7 & 54.0 $\pm$ 0.6 & 8.2 $\pm$ 0.2 & 205 $\pm$ 1 & 0.815 $\pm$ 0.005 \\
 DeepLabV3 & \textbf{69.3 $\pm$ 0.4} & \textbf{70.7 $\pm$ 0.4} & \textbf{54.5 $\pm$ 0.3} & \textbf{56.1 $\pm$ 0.3} & 6.3 $\pm$ 0.0 & 177 $\pm$ 0 & 0.728 $\pm$ 0.002 \\
 Segformer & 68.1 $\pm$ 0.0 & 69.5 $\pm$ 0.0 & 53.4 $\pm$ 0.4 & 54.9 $\pm$ 0.4 & 6.4 $\pm$ 0.1 & 177 $\pm$ 0 & 0.718 $\pm$ 0.001 \\ \midrule
 UNet, no rescaling & 66.2 $\pm$ 0.4 & 67.5 $\pm$ 0.5 & 48.7 $\pm$ 2.4 & 50.1 $\pm$ 2.5 & 6.5 $\pm$ 0.3 & 176 $\pm$ 7 & 0.730 $\pm$ 0.024 \\
 UNet++, no rescaling & 66.0 $\pm$ 0.3 & 67.4 $\pm$ 0.3 & 48.6 $\pm$ 0.9 & 50.0 $\pm$ 1.0 & 7.0 $\pm$ 0.1 & 188 $\pm$ 3 & 0.773 $\pm$ 0.008 \\ 
 DeepLabV3, no rescaling & 65.7 $\pm$ 1.0 & 66.9 $\pm$ 1.1 & 49.2 $\pm$ 1.6 & 50.7 $\pm$ 1.7 & \textbf{5.7 $\pm$ 0.3} & \textbf{158 $\pm$ 9} & \textbf{0.680 $\pm$ 0.035} \\
 Segformer, no rescaling & 63.3 $\pm$ 0.2 & 64.4 $\pm$ 0.2 & 46.4 $\pm$ 0.9 & 47.6 $\pm$ 0.9 & 6.3 $\pm$ 0.2 & 170 $\pm$ 5 & 0.716 $\pm$ 0.019 \\ \bottomrule
\end{tabular}
\end{subtable}
\end{table}

\end{document}